\documentclass{article}

\usepackage[nonatbib,final]{neurips_2020}

\usepackage[utf8]{inputenc} 
\usepackage[T1]{fontenc}    
\usepackage{hyperref}       
\usepackage{url}            
\usepackage{booktabs}       
\usepackage{amsfonts}       
\usepackage{nicefrac}       
\usepackage{microtype}      
\usepackage{enumitem}
\usepackage{graphicx}
\usepackage{subcaption}
\usepackage{tabularx}
\usepackage{multirow}
\usepackage{xr}
\usepackage{wrapfig}

\usepackage{hyperref}

\usepackage{amsmath,amsfonts,bm,amsthm}
\usepackage{xcolor}

\newcommand{\norm}[1]{\left\Vert#1\right\Vert}

\newcommand{\abs}[1]{\left\vert#1\right\vert}

\newcommand{\set}[1]{\left\{#1\right\}}
\newcommand{\parr}[1]{\left (#1\right )}
\newcommand{\brac}[1]{\left [#1\right ]}

\newcommand{\Real}{\mathbb R}

\newcommand{\too}{\rightarrow}

\newcommand{\diag}{\textrm{diag}} 
\newcommand{\din}{d_{\scriptscriptstyle{\textrm{in}}}} 
\newcommand{\dout}{d_{\scriptscriptstyle{\textrm{out}}}} 

\newcommand{\one}{\mathbf{1}}

\newcommand{\eg}{{e.g.}}
\newcommand{\ie}{{i.e.}}

\makeatletter
\newtheorem*{rep@theorem}{\rep@title}
\newcommand{\newreptheorem}[2]{%
\newenvironment{rep#1}[1]{%
 \def\rep@title{#2 \ref{##1}}%
 \begin{rep@theorem}}%
 {\end{rep@theorem}}}
\makeatother

\newtheorem{theorem}{Theorem}
\newreptheorem{theorem}{Theorem}
\newtheorem{lemma}{Lemma}
\newreptheorem{lemma}{Lemma}
\newtheorem{proposition}{Proposition}

\usepackage{amsmath,amsfonts,bm}

\def\eqref#1{equation~\ref{#1}}

\def\1{\bm{1}}

\def\valpha{{\bm{\alpha}}}
\def\vphi{{\bm{\phi}}}
\def\vpsi{{\bm{\psi}}}
\def\vbeta{{\bm{\beta}}}
\def\va{{\bm{a}}}

\def\vi{{\bm{i}}}

\def\vm{{\bm{m}}}

\def\vp{{\bm{p}}}
\def\vq{{\bm{q}}}

\def\vu{{\bm{u}}}
\def\vv{{\bm{v}}}

\def\vx{{\bm{x}}}
\def\vy{{\bm{y}}}

\def\mA{{\bm{A}}}
\def\mB{{\bm{B}}}

\def\mG{{\bm{G}}}
\def\mH{{\bm{H}}}
\def\mI{{\bm{I}}}

\def\mU{{\bm{U}}}
\def\mV{{\bm{V}}}

\def\mX{{\bm{X}}}
\def\mY{{\bm{Y}}}

\def\mSigma{{\bm{\Sigma}}}

\DeclareMathAlphabet{\mathsfit}{\encodingdefault}{\sfdefault}{m}{sl}
\SetMathAlphabet{\mathsfit}{bold}{\encodingdefault}{\sfdefault}{bx}{n}
\newcommand{\tens}[1]{\bm{\mathsfit{#1}}}
\def\tA{{\tens{A}}}
\def\tB{{\tens{B}}}

\def\tF{{\tens{F}}}
\def\tG{{\tens{G}}}

\def\tP{{\tens{P}}}
\def\tQ{{\tens{Q}}}

\def\tX{{\tens{X}}}
\def\tY{{\tens{Y}}}

\def\gN{{\mathcal{N}}}

\def\gX{{\mathcal{X}}}

\newcommand{\R}{\mathbb{R}}

\newcommand{\softmax}{\mathrm{softmax}}

\title{Set2Graph: Learning Graphs From Sets}

\author{%
  Hadar Serviansky$^1$ 
  \And
  Nimrod Segol$^1$ 
  \And
  Jonathan Shlomi$^1$ 
  \And
  Kyle Cranmer$^2$ 
  \And
  Eilam Gross$^1$ 
  \And
  Haggai Maron$^3$ 
  \And
  Yaron Lipman$^1$ \AND
 $^1$\textnormal{Weizmann Institute of Science} \ $^2$\textnormal{New York University} \ $^3$\textnormal{NVIDIA Research}
}

\begin{document}

\maketitle

\begin{abstract}
Many problems in machine learning can be cast as learning functions from sets to graphs, or more generally to hypergraphs; in short, Set2Graph functions. Examples include clustering, learning vertex and edge features on graphs, and learning features on triplets in a collection. \\
A natural approach for building Set2Graph models is to characterize all linear equivariant set-to-hypergraph layers and stack them with non-linear activations. This poses two challenges: (i) the expressive power of these networks is not well understood; and (ii) these models would suffer from high, often intractable computational and memory complexity, as their dimension grows exponentially.

This paper advocates a family of neural network models for learning Set2Graph functions that is both practical and of maximal expressive power (universal), that is, can approximate arbitrary continuous Set2Graph functions over compact sets. Testing these models on different machine learning tasks, mainly an application to particle physics, we find them favorable to existing baselines. \vspace{-5pt}


\end{abstract}

\section{Introduction}
\begin{wrapfigure}{R}{0.45\columnwidth}
	\centering
	\vspace{-10pt}
	\includegraphics[width=0.45\columnwidth]{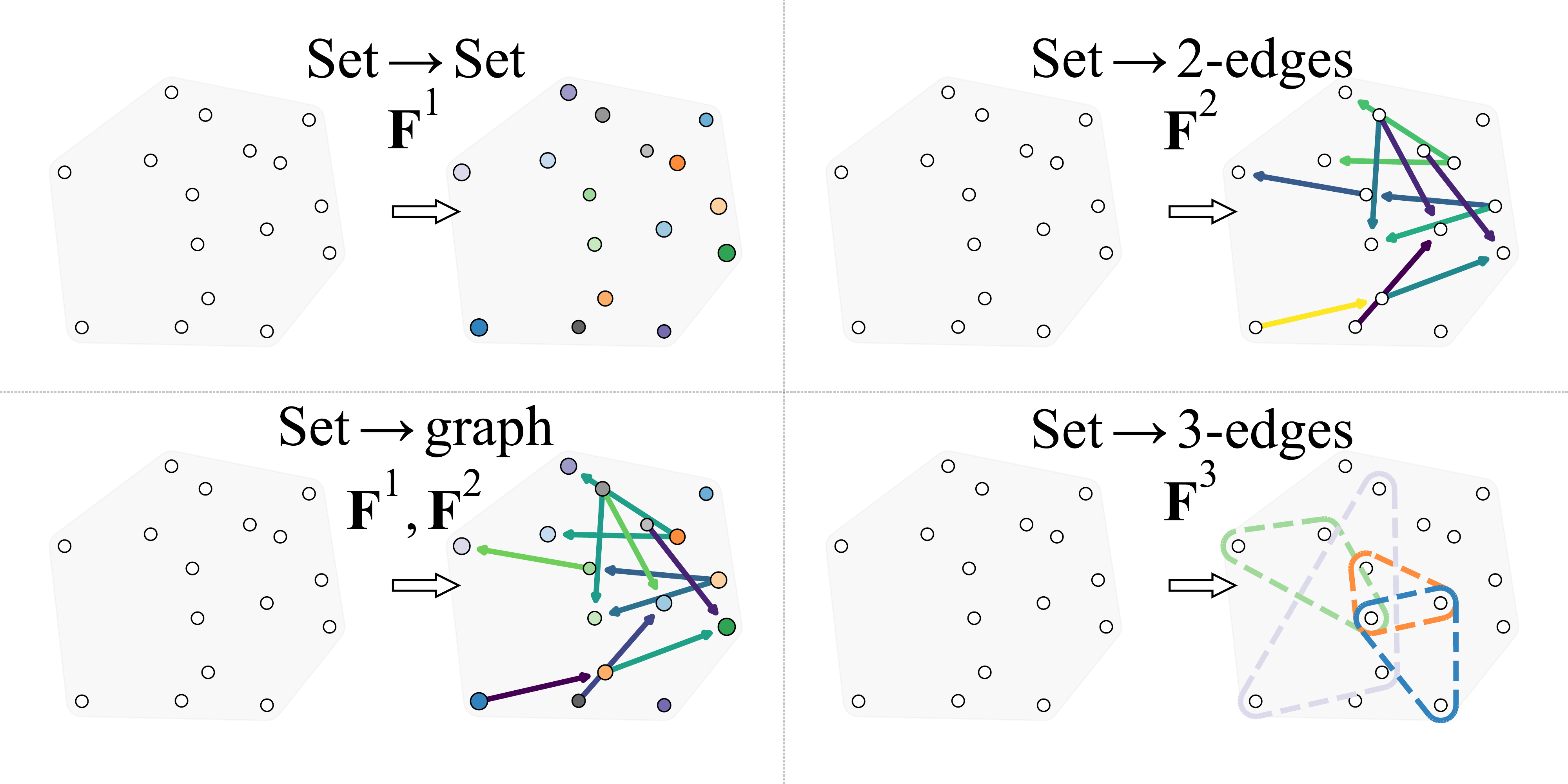}
	\caption{Set-to-graph functions are represented as collections of set-to-k-edge functions.}
	\label{fig:explaintask}
	\vspace{-15pt}
\end{wrapfigure}
We consider the problem of learning functions taking sets of vectors in $\Real^{\din}$ to graphs, or more generally hypergraphs; we name this problem Set2Graph, or set-to-graph. Set-to-graph functions appear in machine-learning applications such as clustering, predicting features on edges and nodes in graphs, and learning $k$-edge information in sets. 

Mathematically, we represent each set-to-graph function as a collection of set-to-$k$-edge functions, where each set-to-$k$-edge function learns features on $k$-edges. That is, given an input set $\gX=\set{\vx_1,\ldots,\vx_n}\subset \Real^{\din}$ we consider functions $\tF^k$ attaching feature vectors to $k$-edges: each $k$-tuple $(\vx_{i_1},\ldots,\vx_{i_k})$ is assigned with an output vector $\tF^k(\gX)_{i_1,i_2,\ldots,i_k,:}\in\Real^{\dout}$. Now, functions mapping sets to hypergraphs with hyper-edges of size up-to $k$ are modeled by $(\tF^1,\tF^2,\ldots,\tF^k)$. For example, functions mapping sets to standard graphs are represented by $(\tF^1,\tF^2)$, see Figure~\ref{fig:explaintask}.

Set-to-graph functions are well-defined if they satisfy a property called \emph{equivariance} (defined later), and therefore the set-to-graph problem is an instance of the bigger class of equivariant learning \cite{cohen2016group,Ravanbakhsh2017,Kondor2018a}. A natural approach for learning equivariant set-to-graph model is using out-of-the-box full equivariant model as in \cite{maron2018invariant}. 

A central question is: Are equivariant models \emph{universal} for set-to-graph functions?  That is, can equivariant models approximate any continuous equivariant function? In equivariant learning literature set-to-set models \cite{zaheer2017deep,qi2017pointnet} are proven equivariant universal \cite{keriven2019universal,segol2020on,sannai2019universal}. In contrast, the situation for graph-to-graph equivariant models is more intricate: some models, such as message passing (a.k.a.~graph convolutional networks), are known to be non-universal \cite{xu2018how,morris2018weisfeiler,maron2019provably,chen2019equivalence}, while high-order equivariant models are known to be universal \cite{maron2019universality} but require using high order tensors and therefore not practical. Universality of equivariant set-to-graph models is not known, as far as we are aware. In particular, are high order tensors required for universality (as the graph-to-graph case), or low order tensors (as in the set-to-set case) are sufficient?

In this paper we: (i) show that low order tensors are sufficient for set-to-graph universality, and (ii) build an equivariant model for the set-to-graph problem that is both \emph{practical} (\ie, small number of parameters and no-need to build high-order tensors in memory) and \emph{provably universal}. We achieve that with a composition of three networks: $\tF^k = \vpsi\circ \vbeta \circ \vphi$, where $\vphi$ is a set-to-set model, $\vbeta$ is a \emph{non-learnable} broadcasting set-to-graph layer, and $\vpsi$ is a simple graph-to-graph network using only a single Multi-Layer Perceptron (MLP) acting on each $k$-edge independently.  

Our main motivation for this work comes from an important set-to-$2$-edges learning problem in particle physics: partitioning (clustering) of simulated particles generated in the Large Hadron Collider (LHC). We demonstrate our model produces state of the art results on this task compared to relevant baselines. We also experimented with another set-to-$2$-edges problem of Delaunay triangulation, and a set-to-$3$-edges problem of 3D convex hull, in which we also achieve superior performances to the baselines.\vspace{-5pt}


\section{Previous work}\vspace{-7pt}
\textbf{Equivariant learning.} In many learning setups the task is invariant or equivariant to certain transformations of the input. The Canonical example is image recognition tasks \cite{lecun1998gradient,krizhevsky2012imagenet} and set classification tasks \cite{zaheer2017deep,qi2017pointnet}. Earlier methods such as \cite{vinyals2015order} used non-equivariant methods to learn set functions. Restricting models to be invariant or equivariant to these transformation was shown to be an excellent approach for reducing the number of parameters of models while improving generalization \cite{zaheer2017deep,qi2017pointnet,kipf,Gilmer2017,Velickovic2017,xu2018how,Kondor2018,maron2018invariant,maron2019provably,cohen2016group,Cohen2016,dieleman2016exploiting,worrall2017harmonic,Welling2018,esteves20173d,Weiler2018,worrall2018cubenet,Weiler2018}.  There has been a keen interest in the analysis of equivariant models \cite{Ravanbakhsh2017,Kondor2018}, especially the analysis of their approximation power \cite{zaheer2017deep,qi2017pointnet,maron2019universality,keriven2019universal,segol2020on,maron2020learning}. As far we know, the set-to-graph case was not treated before. 


\textbf{Similarity learning.} Our work is related to the field of similarity learning, in which the goal is to learn a similarity function on pairs of inputs. In most cases, a siamese architecture is used in order to extract features for each input and then a similarity score is calculated based on this pair of features \cite{bromley1994signature, simo2015discriminative, zagoruyko2015learning}. The difference from our setup is that similarity learning is aimed at extracting pairwise relations between two inputs, independently from the other members of the set, while we learn these pairwise relations globally from the entire input set. In the experimental section we show that the independence assumption taken in similarity learning might cause a significant degradation in performance compared to our global approach.\vspace{-0pt}

\textbf{Other related methods.} \cite{jiang2019meta} suggest a method for meta-clustering that can be seen as an instance of the set2graph setup. Their method is based on LSTMs and therefore depends on the order of the set elements. In contrast, our method is blind (equivariant) to the chosen order of the input sets. The Neural relational Inference model \cite{kipf2018neural} is another related work that targets learning relations and dynamics of a set of objects in an unsupervised manner. \cite{vinyals2015pointer} had previously tackled planar Delaunay triangulation and convex hull prediction problems using a non-equivariant network.\vspace{-7pt}

\section{Learning hypergraphs from sets}\vspace{-7pt}
We would like to learn functions of sets of $n$ vectors in $\Real^{\din}$ to hypergraphs with $n$ nodes (think of the nodes as corresponding to the set elements), and arbitrary $k$-edge feature vectors in $\Real^{\dout}$, where a $k$-edge is defined as a $k$-tuple of set elements. %
A function mapping sets of vectors to $k$-edges is called set-to-$k$-edge function and denoted $\tF^k:\Real^{n\times \din} \too \Real^{n^k\times \dout}$. Consequently, a set-to-hypergraph function would be modeled as a sequence $(\tF^1,\tF^2,\ldots,\tF^K)$, for target hypergraphs with hyperedges of maximal size $K$. For example, $\tF^2$ learns pairwise relations in a set; and $(\tF^1,\tF^2)$ is a function from sets to graphs (outputs both node features and pairwise relations); see Figure \ref{fig:explaintask}.

\textbf{Our goal} is to design permutation equivariant neural network models for $\tF^k$ that are as-efficient-as-possible in terms of number of parameters and memory usage, but on the same time with maximal expressive power, \ie, universal.

\paragraph{Representing sets and $k$-edges.} A matrix $\mX=(\vx_1,\vx_2,\ldots,\vx_n)^T\in\Real^{n\times \din}$ represents a set of $n$ vectors $\vx_i\in\Real^{\din}$ and therefore should be considered up to re-ordering of its rows. 
We denote by $S_n=\set{\sigma}$ the symmetric group, that is the group of bijections (permutations) $\sigma:[n]\too [n]$, where $[n]=\set{1,\ldots,n}$. We denote by $\sigma \cdot \mX$ the matrix resulting in reordering the rows of $\mX$ by the permutation $\sigma$, \ie, $(\sigma \cdot \mX)_{i,j}=\mX_{\sigma^{-1}(i),j}$. In this notation, $\mX$ and $\sigma\cdot \mX$ represent the same set, for all permutations $\sigma$. 

$k$-edges are represented as a tensor $\tY\in\Real^{n^k \times \dout}$, where $\tY_{\vi,:}\in\Real^{\dout}$ denotes the feature vector attached to the $k$-edge defined by the $k$-tuple  $(\vx_{i_1},\vx_{i_2},\ldots,\vx_{i_k})$, where $\vi=(i_1,i_2,\ldots,i_k)\in[n]^k$ is a multi-index with non-repeating indices. Similarly to the set case, $k$-edges are considered up-to renumbering of the nodes by some permutation $\sigma\in S_n$. That is, if we define the action $\sigma\cdot \tY$ by $(\sigma\cdot \tY)_{\vi,j} = \tY_{\sigma^{-1}(\vi),j}$, where $\sigma^{-1}(\vi)=(\sigma^{-1}(i_1),\sigma^{-1}(i_2),\ldots,\sigma^{-1}(i_k))$, then $\tY$ and $\sigma\cdot \tY$ represent the same $k$-edge data, for all $\sigma\in S_n$. 

\paragraph{Equivariance.} A sufficient condition for $\tF^k$ to represent a well-defined map between sets $\mX\in\Real^{n\times \din}$ and $k$-edge data $\tY\in\Real^{n^k\times\dout}$ is \emph{equivariance} to permutations, namely  
\begin{equation}\label{e:equivariant}
\tF^k(\sigma\cdot \mX) = \sigma \cdot \tF^k(\mX),
\end{equation}
for all sets $\mX\in\Real^{n\times \din}$ and permutations $\sigma\in S_n$. Equivariance guarantees, in particular, that the two equivalent sets $\mX$ and $\sigma \cdot \mX$ are mapped to equivalent $k$-edge data tensors $\tF^k(\mX)$ and $\sigma\cdot \tF^k(\mX)$.  

\paragraph{Set-to-$k$-edge models.} In this paper we explore the following equivariant neural network model family for approximating $\tF^k$:
\begin{equation}\label{e:model}
    \tF^k(\mX;\theta) =  \vpsi \circ \vbeta \circ \vphi(\mX),
\end{equation}
where $\vphi,\vbeta$, and $\vpsi$ will be defined soon. For $\tF^k$ to be equivariant (as in \eqref{e:equivariant}) it is sufficient that its constituents, namely $\vphi,\vbeta,\vpsi$, are equivariant. That is, $\vphi,\vbeta,\vpsi$ all satisfy \eqref{e:equivariant}. 

\paragraph{Set-to-graphs models.} Given the model of set-to-$k$-edge functions, a model for a set-to-graph function can now be constructed from a pair of set-to-$k$-edge networks $(\tF^1,\tF^2)$. Similarly, set-to-hypergraph function would require $(\tF^1,\ldots,\tF^K)$, where $K$ is the maximal hyperedge size. Figure~\ref{fig:explaintask} shows an illustration of set-to-$k$-edge and set-to-graph functions

\paragraph{$\vphi$ component.} 
$\vphi:\Real^{n\times \din} \too \Real^{n\times d_1}$ is a set-to-set equivariant model, that is $\vphi$ is mapping sets of vectors in $\Real^{\din}$ to sets of vectors in $\Real^{d_1}$. To achieve the universality goal we will need $\vphi$ to be universal as set-to-set model; that is, $\vphi$ can approximate arbitrary continuous set-to-set functions. Several options exists \cite{keriven2019universal,sannai2019universal} although probably the simplest option is either DeepSets \cite{zaheer2017deep} or one of its variations; all were proven to be universal recently in \cite{segol2020on}.

In practice, as will be clear later from the proof of the universality of the model, when building set-to-graph or set-to-hypergraph model, the $\vphi$ (set-to-set) part of the $k$-edge networks can be shared between different set-to-$k$-edge models, $\tF^k$, without compromising universality.


\paragraph{$\vbeta$ component.}
$\vbeta:\Real^{n\times d_1} \too \Real^{n^k\times d_2}$ is a non-learnable linear \emph{broadcasting layer} mapping sets to $k$-edges. In theory, as shown in \cite{maron2018invariant} the space of equivariant linear mappings $\Real^{n\times d_1} \too \Real^{n^k\times d_2}$ is of dimension $d_1 d_2\mathrm{bell}(k+1)$ which can be very high since $\mathrm{bell}$ numbers have exponential growth. Interestingly, in the set-to-$k$-edge case one can achieve universality with only $k$ linear operators. We define the broadcasting operator to be
\begin{equation}\label{e:beta}
  \vbeta(\mX)_{\vi,:}=\brac{\vx_{i_1},\vx_{i_2},\ldots,\vx_{i_k}},
\end{equation}
where $\vi=(i_1,\ldots,i_k)$ and brackets denote concatenation in the feature dimension, that is, for $\mA\in\Real^{n^k\times d_a}$, $\mB\in\Real^{n^k \times d_b}$ their concatenation is $\brac{\mA,\mB}\in\Real^{n^k \times (d_a+d_b)}$. Therefore, the feature output dimension of $\vbeta$ is $d_2=k d_1$. 
 
\begin{wrapfigure}{R}{0.5\columnwidth}
	\centering
	\vspace{-7pt}
	\includegraphics[width=0.5\columnwidth]{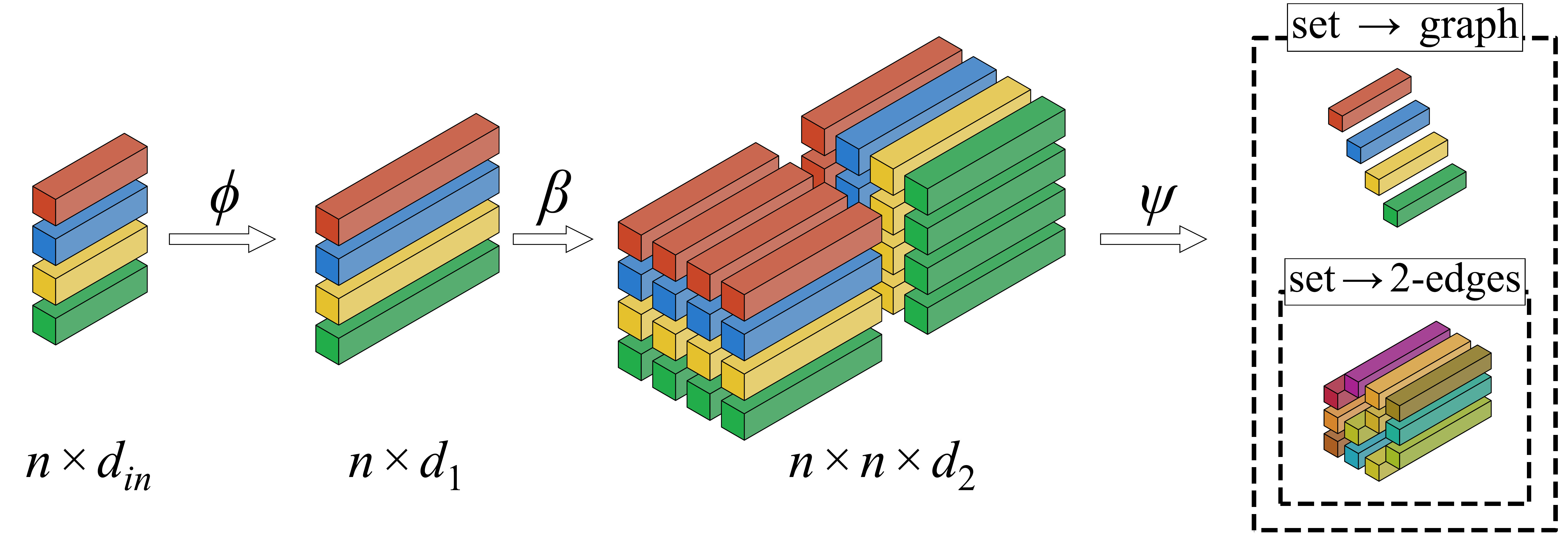}
	\caption{The model architecture for the Set-to-graph and set-to-2-edge functions.}
	\label{fig:modelarch}
	\vspace{-10pt}
\end{wrapfigure}

As an example, consider the graph case, where $k=2$. In this case $\vbeta(\mX)_{i_1,i_2,:}=[\vx_{i_1},\vx_{i_2}]$. This function is illustrated in Figure \ref{fig:modelarch} broadcasting data in $\Real^{n\times d_1}$ to tensor $\Real^{n\times n\times d_2}$. 

To see that the broadcasting layer is equivariant, it is enough to consider a single feature $\vbeta(\mX)_{\vi}=\vx_{i_1}$. Permuting the rows of $\mX$ by a permutation $\sigma$ we get $\vbeta(\sigma \cdot \mX)_{\vi,j}=\vx_{\sigma^{-1}(i_1),j}=\vbeta(\mX)_{\sigma^{-1}(\vi),j}=(\sigma\cdot \vbeta(\mX))_{\vi,j}$.

\paragraph{$\vpsi$ component.} $\vpsi:\Real^{n^k\times d_2}\too \Real^{n^k \times \dout}$ is a mapping of $k$-tensors to $k$-tensors. Here the theory of equivariant operators indicates that the space of linear equivariant maps is of dimension $d_2 \dout \mathrm{bell}(2k)$ that suggests a huge number of model parameters even for a single linear layer. Surprisingly, universality can be achieved with much less, in fact a single linear operator (\ie, scaled identity) in each layer. In the multi-feature multi-layer case this boils to applying a Multi-Layer Perceptron $\vm:\Real^{d_2}\too\Real^{\dout}$ independently to each feature vector in the input tensor $\tX\in\Real^{n^k\times d_2}$. That is, we use
\begin{equation}\label{e:psi}
    \vpsi(\tX)_{\vi,:}=\vm(\tX_{\vi,:}).
\end{equation} 

Figure \ref{fig:modelarch} illustrates set-to-$2$-edges and set-to-graph models incorporating the three components $\vphi,\vbeta,\vpsi$ discussed above.  We note that, Indeed, $\vphi,\vbeta,\vpsi$ are equivariant.

\section{Universality of set-to-graph models}\label{s:universality}\vspace{-5pt}
In this section we prove that the model $\tF^k$ introduced above, is universal, in the sense it can approximate arbitrary continuous equivariant set-to-$k$-edge functions $\tG^k:\Real^{n\times \din}\too \Real^{n^k\times \dout}$ over compact domains $K\subset \Real^{n\times \din}$.
\begin{theorem}\label{thm:set_to_k_edge_universal}
The model $\tF^k$ is set-to-$k$-edge universal. 
\end{theorem}
A corollary of Theorem \ref{thm:set_to_k_edge_universal} establishes set-to-hypergraph universal models:
\begin{theorem}\label{thm:set_to_graph}
The model $(\tF^1,\ldots,\tF^k)$ is set-to-hypergraph universal. 
\end{theorem}
Our main tool for proving Theorem \ref{thm:set_to_k_edge_universal} is a characterization of the equivariant set-to-$k$-edge \emph{polynomials} $\tP^k$. This characterization can be seen as a generalization of the characterization of set-to-set equivariant polynomial recently appeared in \cite{segol2020on}.

We consider an arbitrary set-to-$k$-edge continuous mapping $\tG^k(\mX)$ over a compact set $K\subset\Real^{n\times \din}$. Since $\tG^k$ is equivariant we can assume $K$ is symmetric, \ie, $\sigma\cdot K = K$ for all $\sigma\in S_n$. The proof consists of three parts: (i) Characterization of the equivariant set-to-$k$-edge polynomials $\tP^k$. (ii) Showing that every equivariant continuous set-to-$k$-edge function $\tG^k$ can be approximated by some $\tP^k$. (iii) Every $\tP^k$ can be approximated by our model $\tF^k$.

Before providing the full proof which contains some technical derivations let us provide a simpler universality proof (under some mild extra conditions) for the set-to-$2$-edge model, $\tF^2$, based on the Singular Value Decomposition (SVD). 

\subsection{A simple proof for universality of second-order tensors}
It is enough to consider the $\dout=1$ case; the general case is implied by applying the argument for each output feature dimension independently. Let $\tG^2$ be an arbitrary continuous equivariant set-to-$2$-edge function $\tG^2:K\subset \Real^{n\times \din}\too\Real^{n\times n}$. We want to approximate $\tG^2$ with our model $\tF^2$. First, note that without losing generality we can assume $\tG^2(\mX)$ has a simple spectrum (\ie, eigenvalues are all different) for all $\mX\in K$. Indeed, if this is not the case we can always choose $\lambda>0$ sufficiently large and consider $\tG^2+\lambda \diag(1,2,\ldots,n)$. This diagonal addition does not change the $2$-edge values assigned by $\tG^2$, and it guarantees a simple specturm using standard hermitian matrix eigenvalue perturbation theory (see \eg, \cite{stewart1990matrix}, Section IV:4).

Now let $\tG^2(\mX)=\mU(\mX)\mSigma(\mX)\mV(\mX)^T$ be the SVD of $\tG^2(\mX)$, where $\mU=[\vu_1,\ldots,\vu_n]$ and $\mV=[\vv_1,\ldots,\vv_n]$. Since $\tG^2(\mX)$ has a simple spectrum, $\mU,\mV,\mSigma$ are all continuous in $\mX$; $\mSigma$ is unique, and $\mU,\mV$ are unique up to a sign flip of the singular vectors (\ie, columns of $\mU,\mV$) \cite{o2005critical}. Let us first assume that the singular vectors can be chosen uniquely also up to a sign, later we show how we achieve this with some additional mild assumption.

Now, uniqueness of the SVD together with the equivariance of $\tG^2$ imply that $\mU,\mV$ are continuous \emph{set-to-set} equivariant and $\mSigma$ is a continuous \emph{set invariant} function:
\begin{align}\nonumber
&(\sigma\cdot\mU(\mX)) \mSigma(\mX) (\sigma\cdot \mV(\mX))^T\\ &= \sigma\cdot \mG(\mX)=\mG(\sigma\cdot \mX) \label{e:svd} \\ \nonumber &= \mU(\sigma\cdot\mX) \mSigma(\sigma\cdot\mX) \mV(\sigma\cdot\mX)^T.
\end{align}
Lastly, since $\vphi$ is set-to-set universal there is a choice of its parameters so that it approximates arbitrarily well the equivariant set-to-set function $\mY=[\mU,\mV,\one\one^T\mSigma]$. The $\vpsi$ component can be chosen by noting that $\tG^2(\mX)_{i_1,i_2}=\sum_{j=1}^n \sigma_j \mU_{i_1,j}\mV_{i_2,j}=\vp(\vbeta(\mY)_{i_1,i_2,:})$, where $\sigma_j$ are the singular values, and $\vp:\Real^{6n}\too\Real$ is a cubic polynomial. To conclude pick $\vm$ to approximate $\vp$ sufficiently well so that $\vpsi\circ\vbeta\circ\vphi$ approximates $\tG^2$ to the desired accuracy. 
 
To achieve uniqueness of the singular vectors up-to a sign we can add, \eg, the following assumption: $\one^T \vu_i(\mX)\ne 0 \ne \one^T \vv_i(\mX)$ for all singular vectors and $\mX\in K$. Using this assumption we can always pick $\vu_i(\mX)$, $\vv_i(\mX)$ in the SVD so that $\one^T \vu_i(\mX)> 0$, $\one^T \vv_i(\mX)>0$, for all $i\in[n]$. 
Lastly, note that \eqref{e:svd} suggests that also outer-product can be used as a broadcasting layer. 
We now move to the general proof. \vspace{-5pt}

\subsection{Equivariant set-to-$k$-edge polynomials}\vspace{-3pt} We start with a characterization of the set-to-$k$-edge equivariant polynomials $\tP^k:\Real^{n\times \din}\too \Real^{n^k\times \dout}$. 
We need some more notation. Given a vector $\vx\in\Real^d$, and a multi-index $\alpha\in [n]^d$, we set $\vx^\alpha=\prod_{i=1}^d x_i^{\alpha_i}$; $\abs{\alpha}=\sum_{i=1}^d \alpha_i$; and define accordingly $\mX^\alpha=(\vx_1^\alpha,\ldots,\vx_n^\alpha)^T$. 
Given two tensors $\tA\in\Real^{n^{k_1}}$, $\tB\in\Real^{n^{k_2}}$ we use the notation $\tA\otimes \tB\in\Real^{n^{k_1+k_2}}$ to denote the tensor-product, defined by $(\tA\otimes \tB)_{\vi_1,\vi_2}=\tA_{\vi_1}\tB_{\vi_2}$, where $\vi_1,\vi_2$ are suitable multi-indices.
Lastly, we denote by $\valpha=(\alpha^1,\ldots,\alpha^k)$ a vector of multi-indices $\alpha^i\in[n]^d$, and $\mX^{\valpha}=\mX^{\alpha^1}\otimes \cdots \otimes \mX^{\alpha^k}$. 


\begin{figure}[ht!]
	\begin{center}	
		\includegraphics[width=1.0\textwidth]{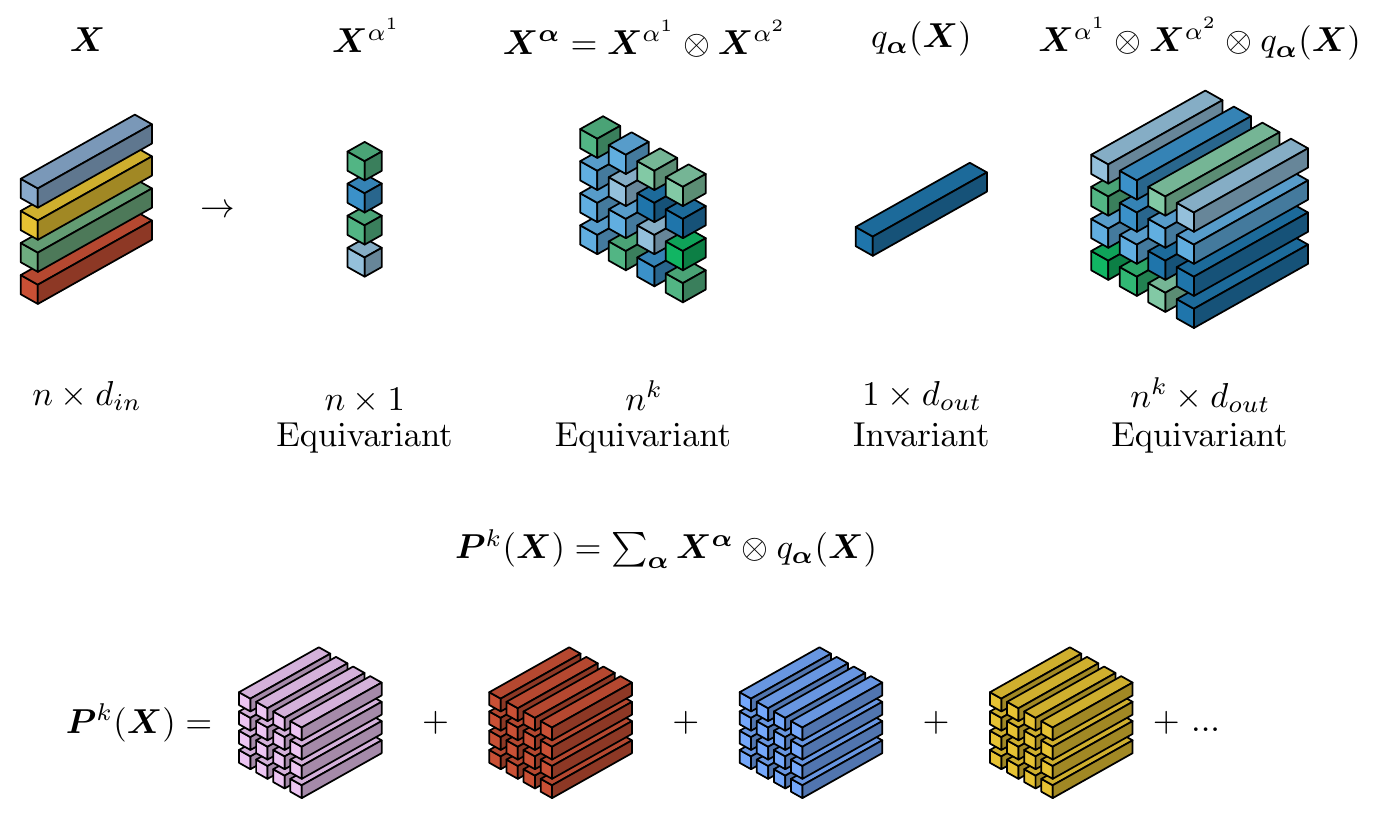}
		\caption{Illustration of the structure of an equivariant set-to-$k$-edge polynomial $\tP^k$, for $k=2$.}
		\label{fig:explain_pk}
	\end{center}
\end{figure}

\begin{theorem}\label{thm:set_to_k_edge_poly}
An equivariant set-to-$k$-edge polynomial $\tP^k:\Real^{n\times \din}\too \Real^{n^k\times \dout}$ can be written as
\begin{equation}\label{e:thm_P^k}
    \tP^k(\mX) = \sum_\valpha \mX^{\valpha}\otimes  \vq_{\valpha}(\mX) 
\end{equation}
where $\valpha=(\alpha^1,\ldots,\alpha^k)$, $\alpha^i\in [n]^{\din}$, and $\vq_{\valpha}:\Real^{n\times \din}\too\dout$ are $S_n$ invariant polynomials. 
\end{theorem}

As an example, consider the graph case, where $k=2$. Equivariant set-to-$2$-edge polynomials take the form: 
\begin{equation}
    \tP^k(\mX) = \sum_{\alpha^1,\alpha^2} \mX^{\alpha^1}\otimes \mX^{\alpha^2}\otimes  \vq_{\alpha^1,\alpha^2}(\mX), 
\end{equation}
and coordinate-wise
\begin{equation}
    \tP^k_{ijl}(\mX) = \sum_{\alpha^1,\alpha^2} \vx_i^{\alpha_1} \vx_j^{\alpha_2}q_{\alpha_1,\alpha_2,l}(\mX).
\end{equation}
The general proof idea and proof itself is given in the supplementary. Figure \ref{fig:explain_pk} provides an illustration of these polynomials.

\paragraph{Approximating $\tG^k$ with a polynomial $\tP^k$.}
We denote for an arbitrary tensor $\tA\in\Real^{a\times b\times \cdots \times c}$ its infinity norm by $\norm{\tA}_{\infty}=\max_{\vi}\abs{\tA_\vi}$.
\begin{lemma}\label{lem:approx_poly}
Let $\tG^k:K\subset \Real^{n\times \din}\too\Real^{n^k\times \dout}$ be a continuous equivariant function over a symmetric domain $K\subset \Real^{n\times \dout}$. For an arbitrary $\epsilon>0$, there exists an equivariant polynomial $\tP^k:\Real^{n\times \din}\too\Real^{n^k\times \dout}$ so that 
$$\max_{\mX\in K} \norm{\tG^k(\mX)-\tP^k(\mX)}_\infty<\epsilon.$$
\end{lemma}

This is a standard lemma, similar to \cite{yarotsky2018universal,maron2019universality,segol2020on};  we provide a proof in the supplementary. 

\paragraph{Approximating $\tP^k$ with a network $\tF^k$.} The final component of the proof of Theorem \ref{thm:set_to_k_edge_universal} is showing that an equivariant polynomial $\tP^k$ can be approximated over $K$ using a network of the form in \eqref{e:model}. 
The key is to use the characterization of Theorem \ref{thm:set_to_k_edge_poly} and write $\tP^k$ in a similar form to our model in \eqref{e:model}:
\begin{equation}\label{e:tP^k_as_our_model}
  \tP^k_{\vi,:}(\mX) = \vp(\vbeta(\mH(\mX))_{\vi,:}),  
\end{equation}
where $\mH:K\too\Real^{n\times d_1}$ defined by $\mH(\mX)_{i,:}=\brac{\vx_i,\vq(\mX)}$, where $\vq(\mX)=\brac{\vq_{\valpha_1}(\mX),\ldots,\vq_{\valpha_m}(\mX)}$, and $\valpha_1,\valpha_2,\ldots,\valpha_m$ are all the multi-indices participating in the sum in \eqref{e:thm_P^k}. Note that 
$$\vbeta(\mH(\mX))_{\vi,:}= \brac{\vx_{i_1},\vq(\mX),\vx_{i_2},\vq(\mX),\ldots,\vx_{i_k},\vq(\mX)}.$$
Therefore, $\vp:\Real^{d_2}\too \Real^{\dout}$ is chosen as the polynomial
$$\vp:\brac{\vx_1,\vy,\vx_2,\vy,\ldots,\vx_k,\vy}\mapsto \sum_{\valpha}\vx_{1}^{\alpha^1}\cdots \vx_{k}^{\alpha^k}\vy_{\valpha},$$
where $\vy=\brac{\vy_{\valpha_1},\ldots,\vy_{\valpha_m}}\in\Real^{m\dout}$, and $\vy_{\valpha_i}\in \Real^{\dout}$. 

In view of \eqref{e:tP^k_as_our_model} all we have left is to choose $\vphi$ and $\vpsi$ (\ie, $\vm$) to approximate $\mH,\vp$ (resp.) to a desired accuracy. We detail the rest of the proof in the supplementary.

\textbf{Universality of the set-to-hypergraph model.}
Theorem \ref{thm:set_to_graph} follows from Theorem \ref{thm:set_to_k_edge_universal} by considering a set-to-hypergraph continuous function $\tG$ as a collection $\tG^k$ of set-to-$k$-edge functions and approximating each one using our model $\tF^k$. Note that universality still holds if $\tF^1,\ldots,\tF^K$ all share the $\vphi$ part of the network (assuming sufficient width $d_1$). 

Note that a set-to-$k$-edge model (in \eqref{e:model}) is not universal when approximating set-to-hypergraph functions:
\begin{proposition}\label{prop:not_universal} 
The set-to-$2$-edge model, $\tF^2$, cannot approximate general set-to-graph functions. 
\end{proposition} 
The proof is in the supplementary; it shows that even the constant function that outputs $1$ for 1-edges (nodes), and $0$ for 2-edges cannot be approximated by a set-to-$2$-edge model $\tF^2$.\vspace{-5pt}

\section{Applications}\vspace{-5pt}
\subsection{Model variants and baselines}\vspace{-5pt}
We tested our model on three learning tasks from two categories: set-to-2-edge and set-to-3-edge.\vspace{-5pt}
\paragraph{Variants of our model.} We consider two variations of our model:\vspace{-5pt}
\begin{itemize}[leftmargin=*]
\item \textbf{S2G}: This is Our basic model. We used the $\tF^2$ and $\tF^3$ (resp.) models for these learning tasks.  for $\tF^2$, $\vphi$ is implemented using DeepSets \cite{zaheer2017deep} with $5$ layers and output dimension $d_1\in\set{5,80}$; $\vpsi$ is implemented with an MLP, $\vm$, with $\set{2,3}$ layers with input dimension $d_2$ defined by $d_1$ and $\vbeta$. $\vbeta$ is implemented according to \eqref{e:beta}: for $k=2$ it uses $d_2=2*d_1$ output features. For $\tF^3$, S2G is described in section \ref{s:convexhull}.
\item  \textbf{S2G+}: For the $k=2$ case we have also tested a more general (but not more expressive) broadcasting $\vbeta$ defined using the full equivariant basis $\Real^n\too\Real^{n^2}$ from \cite{maron2018invariant} that contains $\mathrm{bell}(3)=5$ basis operations. This broadcasting layer gives $d_2=5*d_1$. \vspace{-5pt}
\end{itemize}
 
\vspace{-5pt}

\paragraph{Baselines.}  We compare our results to the following baselines: \vspace{-5pt}
\begin{itemize}[leftmargin=*]
	\item \textbf{MLP}: A standard multilayer perceptron applied to the flattened set features. 
    \item \textbf{SIAM}: A popular similarity learning model (see \eg,  \cite{zagoruyko2015learning}) based on Siamese networks. This model has the same structure as in \eqref{e:model} where $\vphi$ is a Siamese MLP (a non-universal set-to-set function) that is applied independently to each element in the set. We use the same loss we use with our model (according to the task at hand).
    \item \textbf{SIAM-3}: The same architecture as \textbf{SIAM} but with a triplet loss \cite{weinberger2006distance} on the learned representations based on $l2$ distance, see \eg, \cite{schroff2015facenet}. Edge predictions are obtained by thresholding distances of pairs of learned representations. 
     \item \textbf{GNN}: A Graph Neural Network \cite{morris2018weisfeiler} applied to the $k$-NN ($k\in\set{0,5,10}$) induced graph. Edge prediction is done via outer-product \cite{kipf2016variational}.  
    \item \textbf{AVR}: A non-learnable geometric-based baseline called Adaptive Vertex Reconstruction \cite{Waltenberger:2011zz} typically used for the particle physics problem we tackle. More information can be found in the supplementary material.
\end{itemize}

More architecture, implementation, hyper-parameter details and number of parameters can be found in the supplementary material.

\subsection{Partitioning for particle physics}
The first learning setup we tackle is learning set-to-$2$-edge functions. Here, each training example is a pair $(\mX,\mY)$ where $\mX$ is a set $\mX=(\vx_1, \vx_2, \ldots, \vx_n)^T\in\Real^{n\times \din}$ and $\mY \in \{0,1\}^{n\times n}$ is an adjacency matrix (the diagonal of $\mY$ is ignored). Our main experiment tackles an important particle partitioning problem.

\begin{wrapfigure}{R}{0.31\columnwidth}
    \vspace{-10pt}
	\centering
	\includegraphics[width=0.9\linewidth]{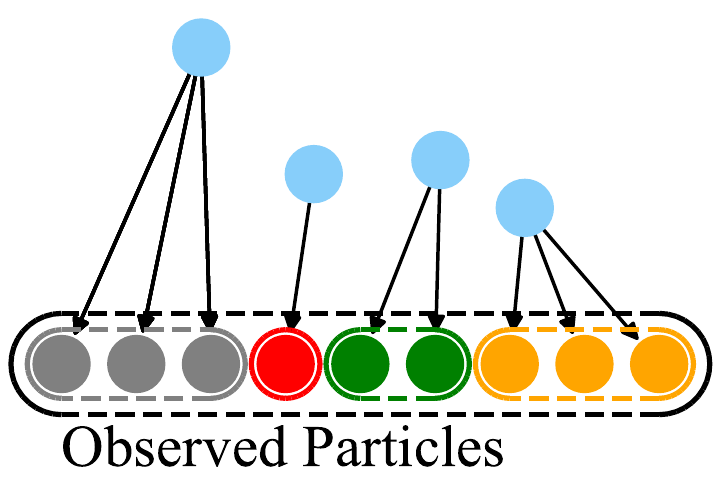} 
	\caption{Illustration of a particle physics experiment. The task is to partition the set of observed particles based on their point of origin (in blue).}
	\label{fig:examplejet}
\end{wrapfigure}

\textbf{Problem statement.} In particle physics experiments, such as the Large Hadron Collider (LHC), beams of incoming particles are collided at high energies. The results of the collision are outgoing particles, whose properties (such as the trajectory) are measured by detectors surrounding the collision point. 
A critical low-level task for analyzing this data is to associate the particle trajectories to their progenitor, which can be formalized as partitioning sets of particle trajectories into subsets according to their unobserved point of origin in space. This task is referred to as vertex reconstruction in particle physics and is illustrated in Figure~\ref{fig:examplejet}. We cast this problem as a set-to-$2$-edge problem by treating the measured particle trajectories as elements in the input set and nodes in the output graph, where the parameters that characterize them serve as the node features. An edge between two nodes indicates that the two particles come from a common progenitor or vertex.

\textbf{Data.} We consider three different types (or \textit{flavors}) of particle sets (called {\textit jets}) corresponding to three different fundamental data generating processes labeled bottom-jets, charm-jets, and light-jets (B/C/L). The important distinction between the flavors is the typical number of partitions in each set.  Since it is impossible to label real data collected in the detectors at the LHC, algorithms for particle physics are typically designed with high-fidelity simulators, which can provide labeled training data. These algorithms are then applied to and calibrated with real data collected by the LHC experiments. The generated sets are small, ranging from 2 to 14 elements each, with around 0.9M sets divided to train/val/test using the ratios 0.6/0.2/0.2. Each set element has 10 features ($d_{in}$). More information can be found in the supplementary material. 

\textbf{Evaluation metrics, loss and post processing.} We consider multiple quantities to quantify the performance of the partitioning: the F1 score, the Rand Index (RI), and the Adjusted Rand Index ($\mathrm{ARI}=({\mathrm{RI}-\mathbb{E}[{\mathrm{RI}}]})/({1-\mathbb{E}[\mathrm{RI}]})$).
All models are trained to minimize the F1 score. We make sure the adjacency matrix of the output graph encodes a valid partitioning of nodes to clusters by considering any connected components as a clique.

\begin{wraptable}[12]{R}{0.45\textwidth} \vspace*{-10pt}
\centering
   \resizebox{0.45\textwidth}{!}{
	\begin{tabular}{c|c|c}
        Model & Epochs & training-time (minutes) \\
        \hline
        S2G   & 193 & 62 \\
        S2G+  & 139 & 47 \\
        GNN   & 91  & 21 \\
        SIAM  & 77  & 24 \\
        SIAM3 & 22  & 322 \\
        MLP   & 132 & 22 \\
    \end{tabular}
    }
    \caption{Training times of different models. Middle column: number of epochs with early stopping. Right column: total training time in minutes.}
    \label{tab:runtime_comparison}
    \vspace{0pt}
\end{wraptable}

\textbf{Results.}  We compare the results of all learning based methods and a typical baseline algorithm used in particle physics (AVR). We also add the results of a trivial baseline that predicts that all nodes have the same progenitor. All models have roughly the same number of parameters. We performed each experiment 11 times with different random initializations, and evaluated the model F1 score, RI and ARI on the test set. The results are shown in Table~\ref{tab:results}-Jets. For bottom and charm jets, which have secondary vertices, both of our models significantly outperform the baselines by 5\%-10\% in all performance metrics. In light-jets, without secondary decays, our models yield similar scores. We also performed an extensive ablation study, see Table A1 in the supplementary material.  Note that S2G+ has the same expressive power as S2G, and produces equivalent results in practice. Table~\ref{tab:runtime_comparison} compares the training times of the different methods.

\subsection{Learning Delaunay triangulations}\label{s:delaunay}
In a second set-to-$2$-edge task we test our model's ability to learn Delaunay triangulations, namely given a set of planar points we want to predict the Delaunay edges between pairs of points, see \eg, \cite{de1997computational} Chapter 9. 
We generated $50k$ planar point sets as training data and $5k$ planar point sets as test data; the point sets, $\mX\in\mathbb{R}^{n\times 2}$, were uniformly sampled in the unit square, and a ground truth matrix in $\{0,1\}^{n\times n}$ was computed per point set using a Delaunay triangulation algorithm. The number of points in a set, $n$, is either $50$ or varies and is randomly chosen from $\{20, \ldots, 80\} $. Training was stopped after 100 epochs. As in the previous experiment, all models have roughly the number of parameters. See more implementation details in the supplementary material. In Table \ref{tab:results}-Delaunay we report accuracy of prediction as well as precision recall and F1 score.  Evidently, both of our models (S2G and S2G+) outperform the baselines. We also tried the MLP baseline that yielded very low F1 scores (\ie, $\leq0.1$). See also Figure 1 in the supplementary material for visualizations of several qualitative examples.

\subsection{Set to 3-edges: learning the convex-hull of a point cloud}\label{s:convexhull}
In the last experiment, we demonstrate learning of set-to-$3$-edge function. The learning task we target is finding supporting triangles in the convex hull of a set of points in $\R^3$. In this scenario, the input is a point set $\mX\in\Real^{n \times 3}$, and the function we wanted to learn is $\tF^3:\Real^{n\times 3} \too \Real^{n^3}$ where the output is a probability for each triplet of nodes (triangle) $\set{\vx_{i_1}, \vx_{i_2}, \vx_{i_3}}$ to belong to the triangular mesh that describes the convex hull of $\mX$. \cite{vinyals2015pointer} had previously tackled a 2-dimensional version of this problem, but since their network predicts the order of nodes in the 2D convex hull, it is not easily adapted to the 3D settings.

Note that storing $3$-rd order tensors in memory is not feasible, hence we concentrate on a \emph{local} version of the problem: Given a point set $\mX\subset\Real^3$, identify the triangles within the $K$-Nearest-Neighbors of each point that belong to the convex hull of the entire point set $\mX$. We used $K=10$.
%
Therefore, for broadcasting ($\vbeta$) from point data to 3-edge data, instead of holding a $3$-rd order tensor in memory we broadcast only the subset of $K$-NN neighborhoods. This allows working with high-order information with relatively low memory footprint. Furthermore, since we want to consider $3$-edges (triangles) with no order we used invariant universal set model (DeepSets again) as $\vm$.
For $k=3$, \textbf{S2G} is implemented as follows: $\vphi$ is implemented using DeepSets with $3$ layers and output dimension $d_1=512$; $\vbeta$ triplets of points to sets. $\vpsi$ is implemented with a DeepSets with $3$ layers of $64$ features, followed by an MLP, $\vm$, with $3$ layers. More details are in the supplementary.

\begin{table}[]
    \centering
    \setlength{\tabcolsep}{1pt}
    \begin{tabularx}{\textwidth}{c|c|c|c}
\tiny
    \setlength{\tabcolsep}{2pt}
		\begin{tabular}[t]{ll|ccc}
         & Model & F1 & RI & ARI \\
\hline

   & S2G      & 0.646$\pm$0.003         & 0.736$\pm$0.004         & 0.491$\pm$0.006         \\
   & S2G+     & \textbf{0.655$\pm$0.004}& \textbf{0.747$\pm$0.006}& \textbf{0.508$\pm$0.007}\\
   & GNN      & 0.586$\pm$0.003         & 0.661$\pm$0.004         & 0.381$\pm$0.005         \\
 B & SIAM     & 0.606$\pm$0.002         & 0.675$\pm$0.005         & 0.411$\pm$0.004         \\
   & SIAM-3   & 0.597$\pm$0.002         & 0.673$\pm$0.005         & 0.396$\pm$0.005         \\
   & MLP      & 0.533$\pm$0.000         & 0.643$\pm$0.000         & 0.315$\pm$0.000         \\
   & AVR      & 0.565         & 0.612           & 0.318 \\
   & trivial  & 0.438         & 0.303         & 0.026         \\
\hline
   & S2G      & 0.747$\pm$0.001         & 0.727$\pm$0.003         & 0.457$\pm$0.004         \\
   & S2G+     & \textbf{0.751$\pm$0.002}& \textbf{0.733$\pm$0.003}& \textbf{0.467$\pm$0.005}\\
   & GNN      & 0.720$\pm$0.002         & 0.689$\pm$0.003         & 0.390$\pm$0.005         \\
 C & SIAM     & 0.729$\pm$0.001         & 0.695$\pm$0.002         & 0.406$\pm$0.004         \\
   & SIAM-3   & 0.719$\pm$0.001         & 0.710$\pm$0.003         & 0.421$\pm$0.005         \\
   & MLP      & 0.686$\pm$0.000         & 0.658$\pm$0.000         & 0.319$\pm$0.000         \\
   & trivial  & 0.610        & 0.472         & 0.078 \\
   & AVR      & 0.695        & 0.650         & 0.326 \\
\hline
   & S2G      & 0.972$\pm$0.001         & \textbf{0.970$\pm$0.001}& \textbf{0.931$\pm$0.003}\\
   & S2G+     & 0.971$\pm$0.002         & 0.969$\pm$0.002         & 0.929$\pm$0.003         \\
   & GNN      & 0.972$\pm$0.001         & \textbf{0.970$\pm$0.001}& 0.929$\pm$0.003         \\
 L & SIAM     & \textbf{0.973$\pm$0.001}& \textbf{0.970$\pm$0.001}& 0.925$\pm$0.003         \\
   & SIAM-3   & 0.895$\pm$0.006         & 0.876$\pm$0.008         & 0.729$\pm$0.015         \\
   & MLP      & 0.960$\pm$0.000         & 0.957$\pm$0.000         & 0.894$\pm$0.000         \\
   & trivial  & 0.910         & 0.867         & 0.675 \\
   & AVR      & 0.970         & 0.965         & 0.922 \\
		\end{tabular}%
	&  
    \tiny
    \setlength\tabcolsep{2pt}
     \begin{tabular}[t]{l|rrrr}
          & \multicolumn{1}{l}{Acc} & \multicolumn{1}{l}{Prec} & \multicolumn{1}{l}{Rec} & \multicolumn{1}{l}{F1} \\
        \hline
        \multicolumn{5}{c}{$n=50$} \\
        \hline
        S2G  & \textbf{0.984} & \textbf{0.927} & 0.926 & \textbf{0.926} \\
        S2G+ & 0.983 & \textbf{0.927} & 0.925 & \textbf{0.926} \\
        GNN0 & 0.826 & 0.384 & 0.966 & 0.549 \\
        GNN5 & 0.809 & 0.363 & \textbf{0.985} & 0.530 \\
        GNN10 & 0.759 & 0.311 & 0.978 & 0.471 \\
        SIAM & 0.939 & 0.766 & 0.653 & 0.704 \\
        SIAM-3 & 0.911 & 0.608 & 0.538 & 0.570 \\
        \multicolumn{5}{l}{} \\
        \multicolumn{5}{l}{} \\
        \hline
        \multicolumn{5}{c}{$n\in\{20, \ldots, 80\}$} \\
        \hline
        S2G  & \textbf{0.947} & \textbf{0.736} & 0.934 & \textbf{0.799} \\
        S2G+ & \textbf{0.947} & 0.735 & 0.934 & 0.798 \\
        GNN0 & 0.810 & 0.387 & 0.946 & 0.536 \\
        GNN5 & 0.777 & 0.352 & \textbf{0.975} & 0.506 \\
        GNN10 & 0.746 & 0.322 & 0.970 & 0.474 \\
        SIAM & 0.919 & 0.667 & 0.764 & 0.687 \\
        SIAM-3 & 0.895 & 0.578 & 0.622 & 0.587 \\
    \end{tabular}
     & \tiny
     \setlength\tabcolsep{2pt}
		\begin{tabular}[t]{l|ccc}
             & \# points & F1 & {\tiny AUC-ROC} \\
             \hline
             \multicolumn{4}{c}{Spherical} \\
             \hline
			 S2G & 30      & \textbf{0.780} & \textbf{0.988} \\ 
			 GNN5 & 30     & 0.693 & 0.974 \\
             SIAM & 30     & 0.425 & 0.885 \\ 
             \hline
             S2G & 50      & 0.686 & \textbf{0.975} \\ 
             GNN5 & 50     & \textbf{0.688} & 0.973 \\
             SIAM & 50     & 0.424 & 0.890 \\ 
             \hline
             S2G & 20-100  & 0.535 & 0.953 \\ 
             GNN5 & 20-100 & \textbf{0.667} & \textbf{0.970} \\
             SIAM & 20-100 & 0.354 & 0.885 \\ 
             \hline
             \multicolumn{4}{c}{Gaussian} \\
             \hline
             S2G & 30      & \textbf{0.707} & \textbf{0.996} \\ 
             GNN5 & 30 & 0.5826 & 0.9865 \\
             SIAM & 30     & 0.275 & 0.946 \\ 
             \hline
             S2G & 50      & \textbf{0.661} & \textbf{0.997} \\ 
             GNN5 & 50 & 0.4834 & 0.9917 \\
             SIAM & 50     & 0.254 & 0.974 \\ 
             \hline
             S2G & 20-100  & \textbf{0.552} & \textbf{0.994} \\ 
             GNN5 & 20-100 & 0.41 & 0.9866 \\
             SIAM & 20-100 & 0.187 & 0.969 \\
		\end{tabular}%
     & \tiny
     \setlength\tabcolsep{1pt} 
	\begin{tabular}[t]{cc} 
        \scriptsize GT & \scriptsize predicted \\
        \hline
        \multicolumn{2}{c}{\scriptsize $n=30$} \\
        \hline \\
		\includegraphics[width=0.06\columnwidth,keepaspectratio]{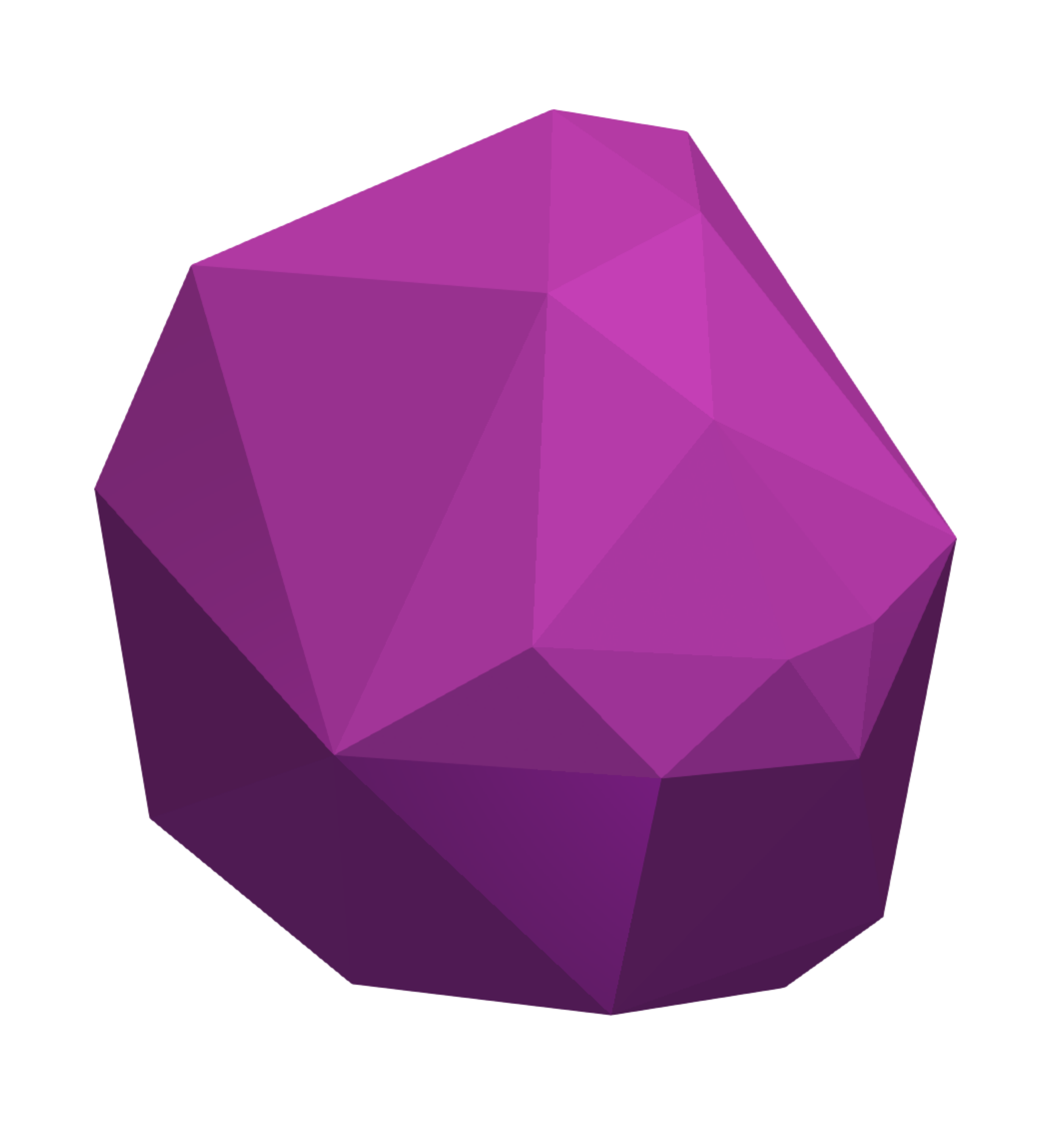} &
		\includegraphics[width=0.06\columnwidth,keepaspectratio]{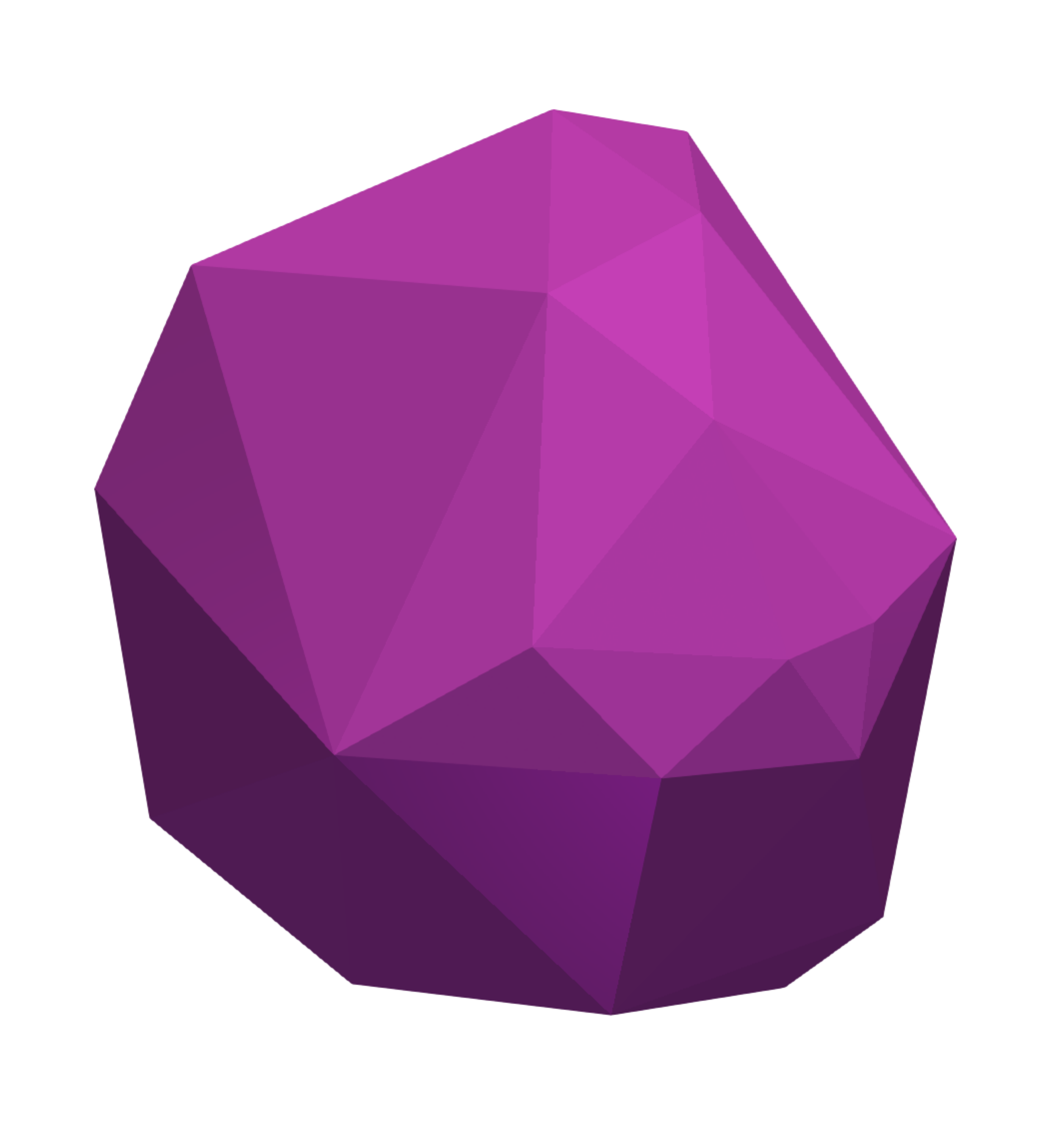} \\
		\includegraphics[width=0.06\columnwidth , keepaspectratio]{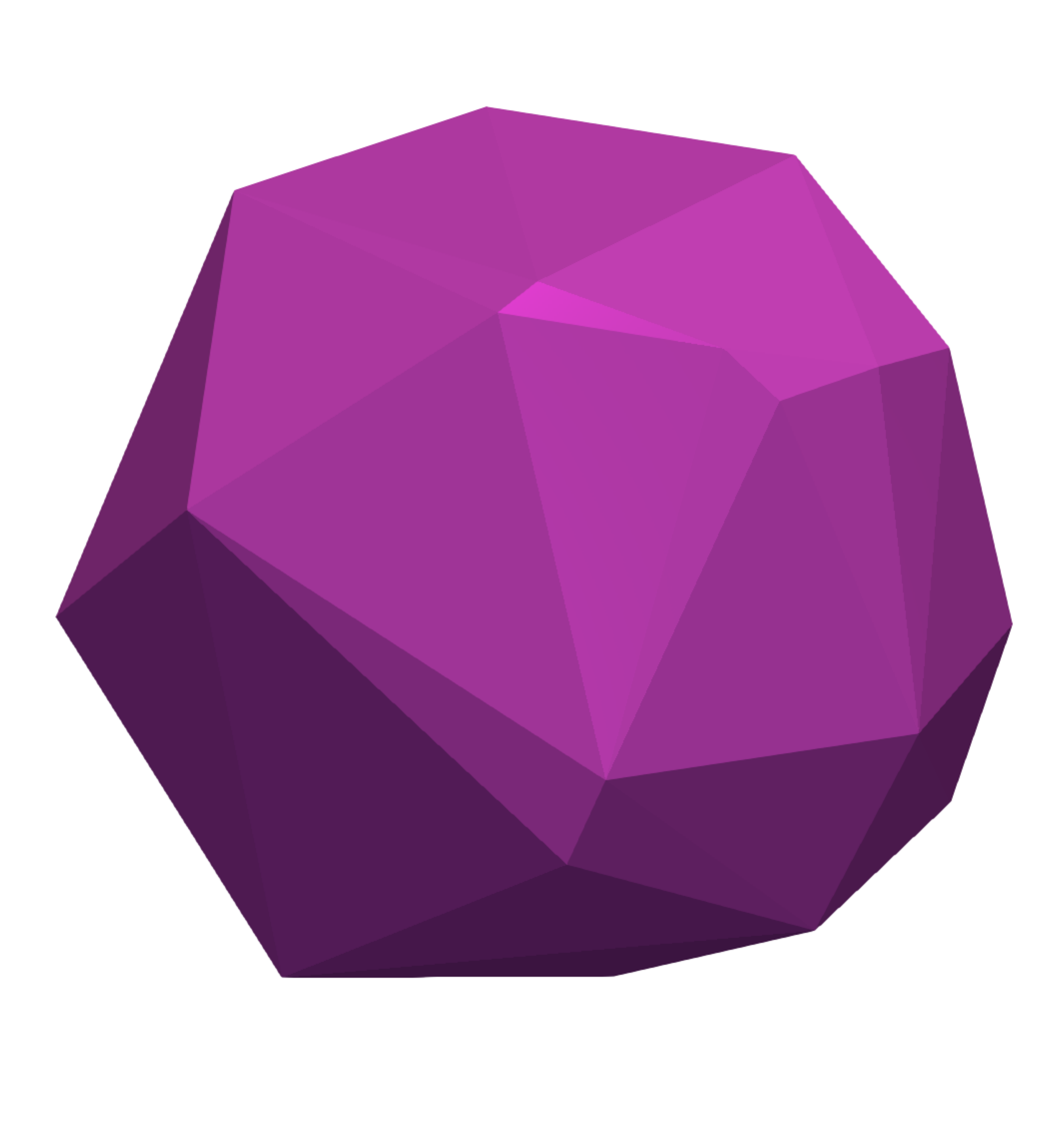} & 
		\includegraphics[width=0.06\columnwidth , keepaspectratio]{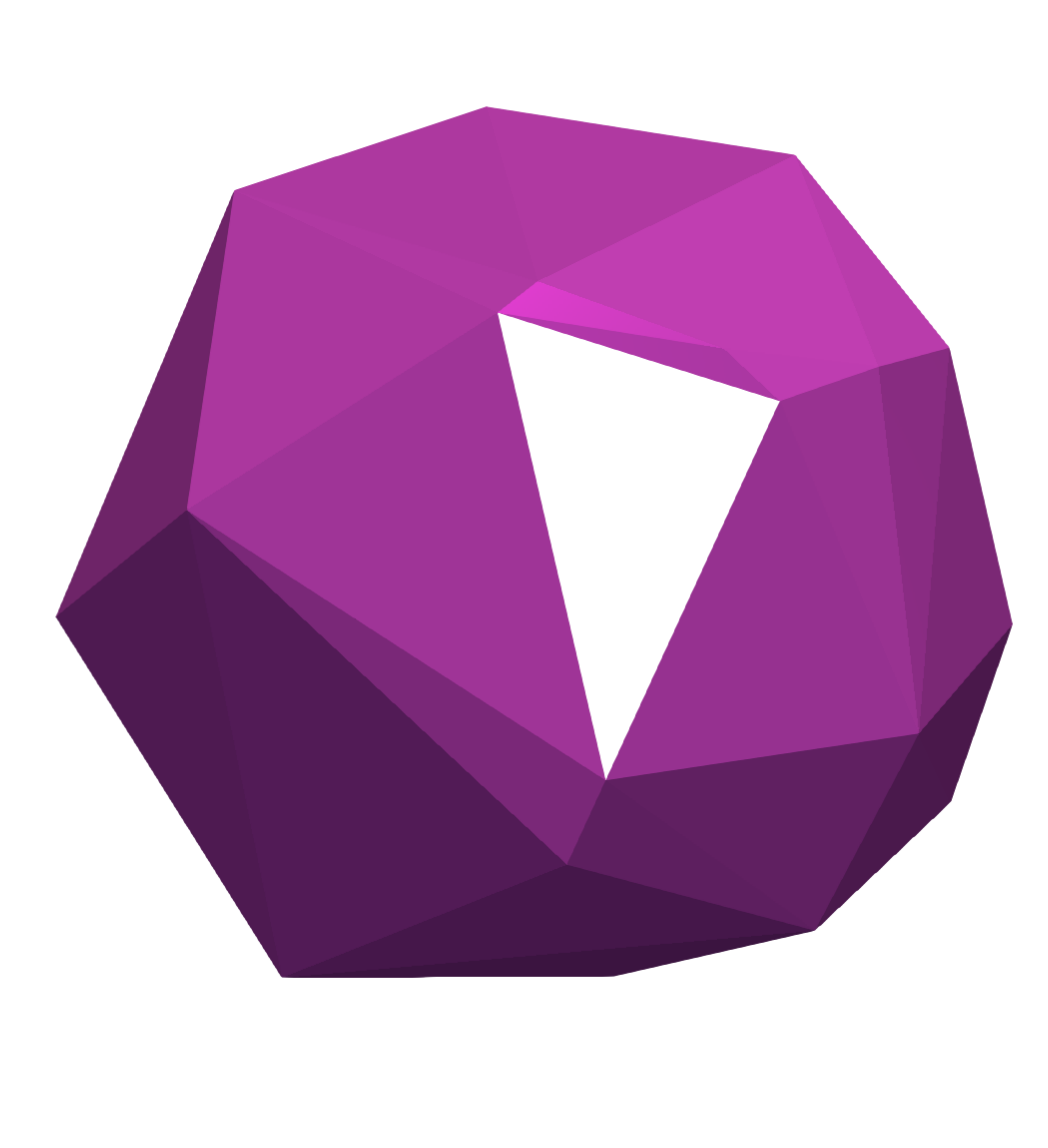} \\
		\hline
		\multicolumn{2}{c}{\scriptsize $n=50$} \\
		\hline \\
		\includegraphics[width=0.06\columnwidth, keepaspectratio]{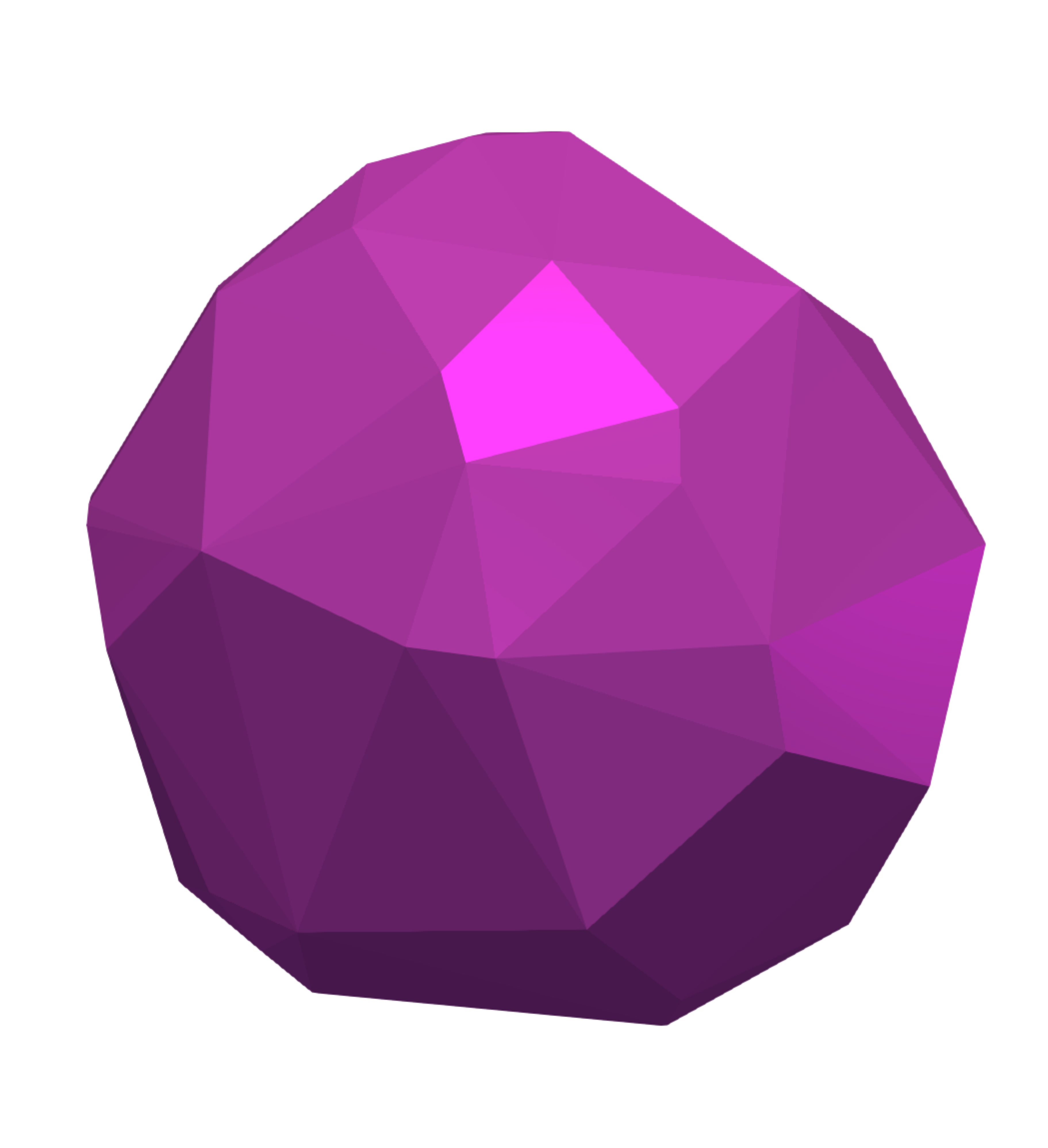}&
		\includegraphics[width=0.06\columnwidth, keepaspectratio]{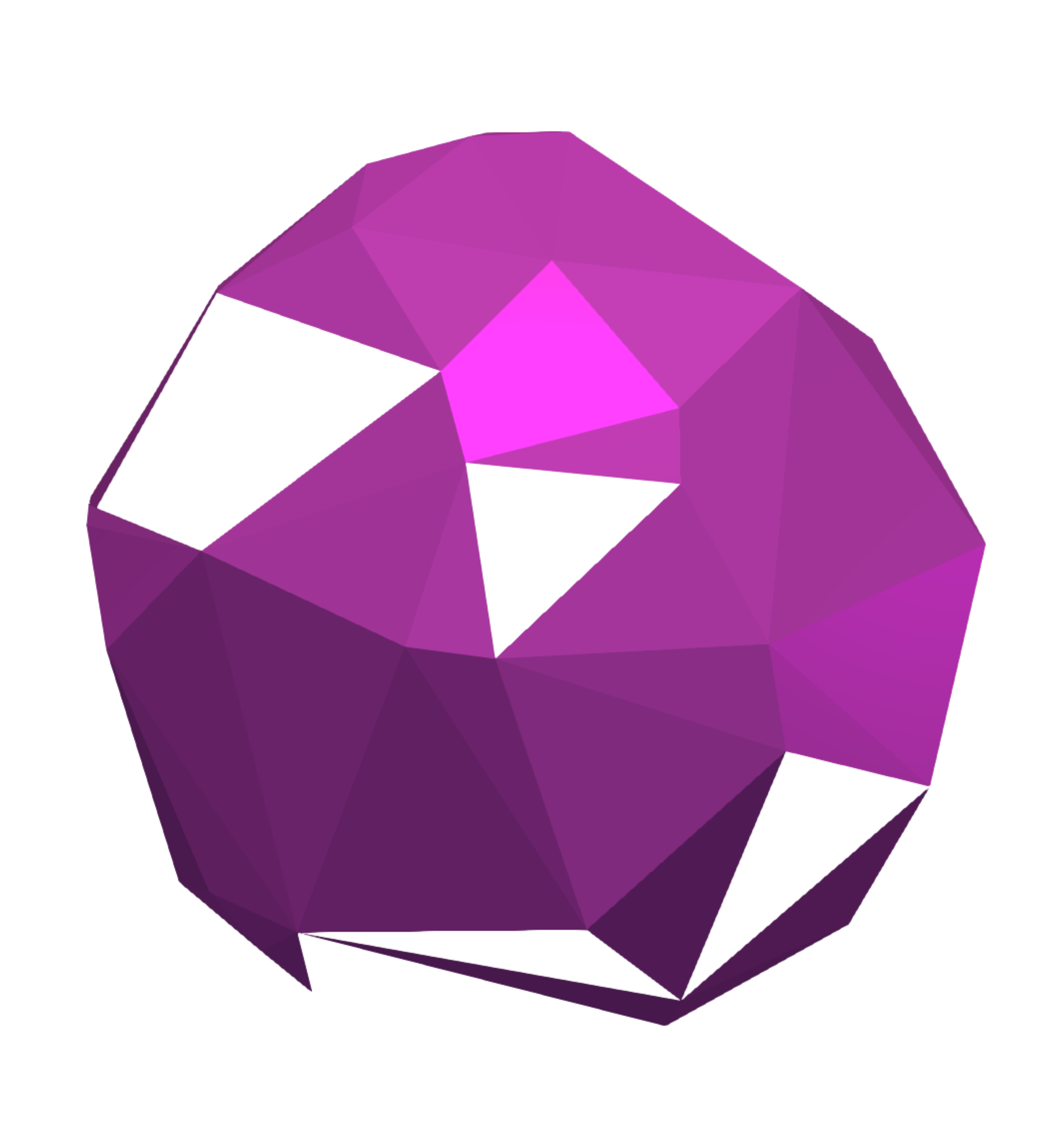} \\
		\includegraphics[width=0.06\columnwidth, keepaspectratio]{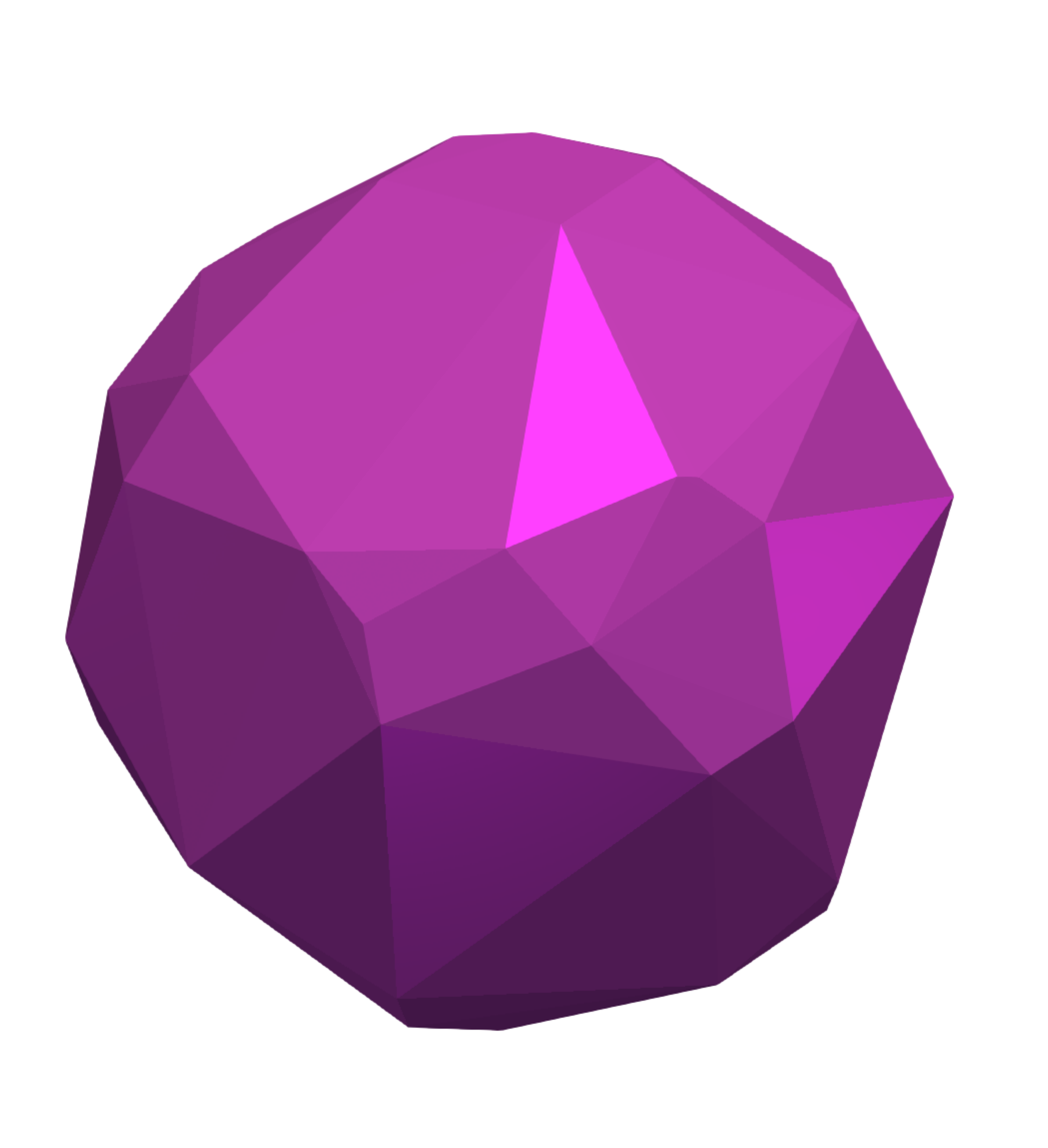}	&
		\includegraphics[width=0.06\columnwidth, keepaspectratio]{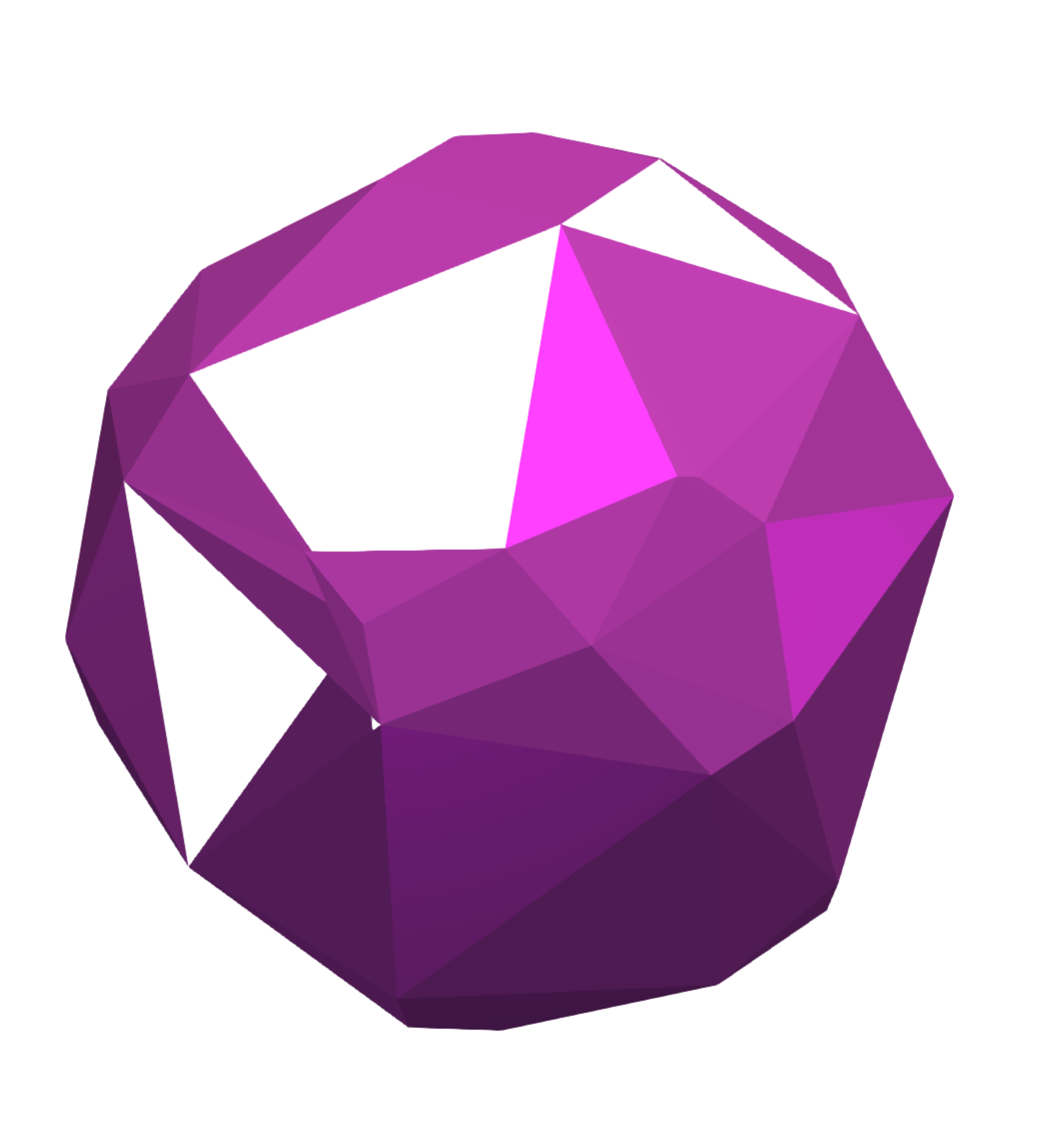}  \\
		\end{tabular}
     \\
    {\tiny Jets} & {\tiny Delaunay} & {\tiny Convex Hull (a)} & {\tiny Convex Hull (b)} 
\end{tabularx}
    \caption{Jets - Performance of partitioning for three types of jets. Delaunay - Results on the Delaunay triangulation task. Convex Hull (a) and (b) - Convex hull learning quantitative and qualitative results. }
    \label{tab:results}
    \vspace{-10pt}
\end{table}


We tested our S2G model on two types of data: Gaussian and spherical. For both types we draw point sets in $\Real^3$ i.i.d.~from standard normal distribution, $\gN(0,1)$, where for the spherical data we normalize each point to unit length.
%
%
We generated $20k$ point set samples as a training set, $2k$ for validation and another $2k$ for test set. Point sets are in $\Real^{n\times 3}$, where $n=30$, $n=50$, and $n\in\brac{20,100}$. 
We compare our method, S2G, to the SIAM, GNN and MLP baselines. The F1 scores and AUC-ROC of the predicted convex hull triangles are shown in Table \ref{tab:results}-Convex Hull, where our model outperform the baselines in most cases (as in the previous experiment we exclude MLP from the table since it yields very low results). See Figure (b) for several examples of triangles predicted using our trained model compared to the ground truth.

\subsection{Discussion}
In the experiments above we compared our model S2G (and its variant S2G+) with several, broadly used, state-of-the-art architectures. As noted, our models compare favorably with these baselines in all tasks. 
We attribute this to the increased expressive power (universality) of these models. Our architectures are able to learn relations between set instances while reasoning about the whole set, where Siamese networks rely only on the relation between pairs or triplets of points. The GNN models use a prescribed connectivity that hinder efficient set learning.

\section{Conclusion}
In this paper, we presented a novel neural network family that can model equivariant set-to-$k$-edge functions and consequently set-to-graph, or more generally set-to-hypergraph functions. The family uses a relatively small number of parameters, compared to other equivariant models, and is shown to be a universal approximator of such continuous equivariant functions. We show the efficacy of these networks on several tasks, including a real-life particle physics problem.  There are many directions for future work. One is adapting the model to learn valid clustering of sets (\ie, learn graphs that are built of disjoint cliques). Another direction is incorporating our network architecture in set models, with the goal of improving performance of general set tasks by constructing an intermediate latent graph representation, for example for constructing scene graphs as in  \cite{raboh2020differentiable}.

\section*{Broader Impact}
Our contribution describes a class of neural network models for functions from sets to graphs and includes theoretical results that the model family is universal. The potential uses of these models is very broad and includes the physical sciences, computer graphics, and social networks. The paper includes experiments that show the positive impact of these models in the context of particle physics, and similar tasks appear repeatedly in the physical sciences. The models could also be used for social networks and areas with  more complex ethical and societal consequences. Because the models treat the input as a set and are permutation equivariant, they have the potential to mitigate potential bias in data due to sorting and other pre-processing that could impact methods that treat the input as a sequence. Otherwise, the considerations of bias in the data and impact of failure are no different for our model than the generic considerations of the use of supervised learning and neural networks. Finally we note that the models we describe are already being used in real-world particle physics research.

\iftrue
\begin{ack}
HS, NS and YL were supported in part by the European Research Council (ERC Consolidator Grant, "LiftMatch" 771136), the Israel Science Foundation (Grant No. 1830/17) and by a research grant from the Carolito Stiftung (WAIC). JS and EG were supported by the NSF-BSF Grant 2017600 and the ISF Grant 2871/19. KC was supported by the National Science Foundation under the awards ACI-1450310, OAC-1836650, and OAC-1841471 and by the Moore-Sloan data science environment at NYU.
\end{ack}
\fi

\bibliography{set2graph_arxiv}
\bibliographystyle{abbrv} 

\renewcommand{\theequation}{A\arabic{equation}}
\setcounter{equation}{0}
\renewcommand{\thefigure}{A\arabic{figure}}
\setcounter{figure}{0}
\renewcommand{\thelemma}{A\arabic{lemma}}
\setcounter{lemma}{0}
\renewcommand{\thetable}{A\arabic{table}}
\setcounter{table}{0} 
\renewcommand{\thesection}{A\arabic{section}}
\setcounter{section}{0}
\renewcommand{\thesubsection}{A\arabic{subsection}}
\setcounter{subsection}{0}

\newpage
\title{Set2Graph: Learning Graphs From Sets: Supplementary Material}
\maketitle

\section{Architectures and hyper-parameters} \label{imp_details}

Our \textbf{S2G} model (as well as \textbf{S2G+} for the main task) follows the formula $\tF^k = \vpsi\circ \vbeta \circ \vphi$, where $\vphi$ is a set-to-set model, $\vbeta$ is a \emph{non-learnable} broadcasting set-to-graph layer, and $\vpsi$ is a simple graph-to-graph network using only a single Multi-Layer Perceptron (MLP) acting on each $k$-edge feature vector independently. We note that all the hyper-parameters were chosen using the validation scores.
All of the models used in the experiments are explained in section 5 in the main paper. Here, we add more implementation details, hyper-parameters and number of parameters. 

\paragraph{Notation.}
"DeepSets / MLP of widths $\brac{256,256,5}$" means that we use a DeepSets/MLP network with 3 layers, and each layer's output feature size is its corresponding argument in the array (\eg, the first and second layers have output feature size of $256$, while the third layer output feature size is $5$). Between the layers we use ReLU as a non linearity.

\paragraph{Partitioning for particle physics applications.}
For our model \textbf{S2G}, $\vphi$ is implemented using DeepSets \cite{zaheer2017deep} with 5 layers of width $\brac{256, 256, 256, 256, 5}$. $\vpsi$ is implemented with an MLP $\brac{256, 1}$, with output considered as edge probabilities.
Instead of using a max or sum pooling in DeepSets layers, we used a self-attention mechanism based on \cite{ilse2018attention} and \cite{vaswani2017attention}:
\begin{equation} \label{eq:attention}
    Attention\parr{\mX} = \softmax\parr{{\frac{\tanh{f_1(\mX)} \cdot f_2(\mX)^T}{\sqrt{d_{\text{small}}}}}} \cdot \mX,
\end{equation}
where $f_1, f_2$ are implemented by two single layer MLPs of width $\brac{d_\text{small}=\frac{d_{X}}{10}}$. The model has 457449 learnable parameters. 
\textbf{S2G+} is identical to \textbf{S2G}, except that $\vbeta$ is defined using the full equivariant basis $\Real^n\too\Real^{n^2}$ from \cite{maron2018invariant} that contains $\mathrm{bell}(3)=5$ basis operations. It has 461289 learnable parameters.

We used a grid search for the following hyper-parameters: learning rate in $$\set{1e-5, 3e-5, 1e-4, 3e-4, 1e-3, 3e-3},$$ DeepSets layers' width in $\set{64, 128, 256, 512}$, number of layers in $\set{3,4,5}$, $\vpsi$ (MLP) with widths in $\set{\brac{128,1}, \brac{256,1}, \brac{512,1}, \brac{128, 256, 128, 1}}$, and with or without attention mechanism in DeepSets. 

The following hyper-parameters choice is true for all models, unless stated otherwise. 
As a loss, we used a combination of soft F1 score loss and an edge-wise binary cross-entropy loss. We use early stopping based on validation score, batch size of 2048, adam optimizer \cite{kingma2014adam} with learning rate of $1e-3$. Training takes place in less than 2 hours on a single Tesla V100 GPU.

The deep learning baselines are implemented as follows: \textbf{SIAM} is implemented similarly to S2G, with the exception that instead of using DeepSets as $\vphi$, we use MLP $\brac{384, 384, 384, 384, 5}$. The learning rate is $3e-3$ and SIAM has 452742 learnable parameters. \textbf{SIAM-3} uses a Siamese MLP of widths $\brac{384, 384, 384, 384, 20}$ to extract node features, and the edge logits are the $l_2$ distances between the nodes. SIAM-3 uses a triplets loss \cite{weinberger2006distance} - we draw random triplets \textit{anchor, neg, pos} where \textit{anchor} and \textit{pos} are of the same cluster, and \textit{neg} is of a different cluster, and the loss is defined as \footnote{A natural disadvantage of the triplets loss is that it cannot learn from sets with a single cluster, or sets with size 2.} $$L_i = \min{\parr{d_{l2}\parr{anch_i, pos_i}-d_{l2}\parr{anch_i, neg_i}+2, 0}}$$
The learning rate is $1e-4$ and SIAM-3 has 455444 learnable parameters. Due to the triplet choice process, training takes place in around 9 hours. \textbf{MLP} is a straight-forward fully-connected network acting on the flattened feature vectors of the input sets. It uses fully-connected layers of widths $\brac{512,256,512,15^2}$, and has 455649 learnable parameters. \textbf{GNN} is a GraphConv network \cite{morris2018weisfeiler} where the underlying graph is selected as the $k$-NN ($k=5$) graph constructed using the $l_2$ distance between the elements' feature vectors. the GraphConv layers have $\brac{350,350,300,20}$ features, and the edge logits between set elements are based on the inner product between the features of each $2$ elements. GNN has 455900 learnable parameters.

\begin{figure}
	\centering
	\setlength\tabcolsep{0.0pt} 
	\begin{tabular}{cc|cc} 
	    \includegraphics[width=0.25\columnwidth,keepaspectratio]{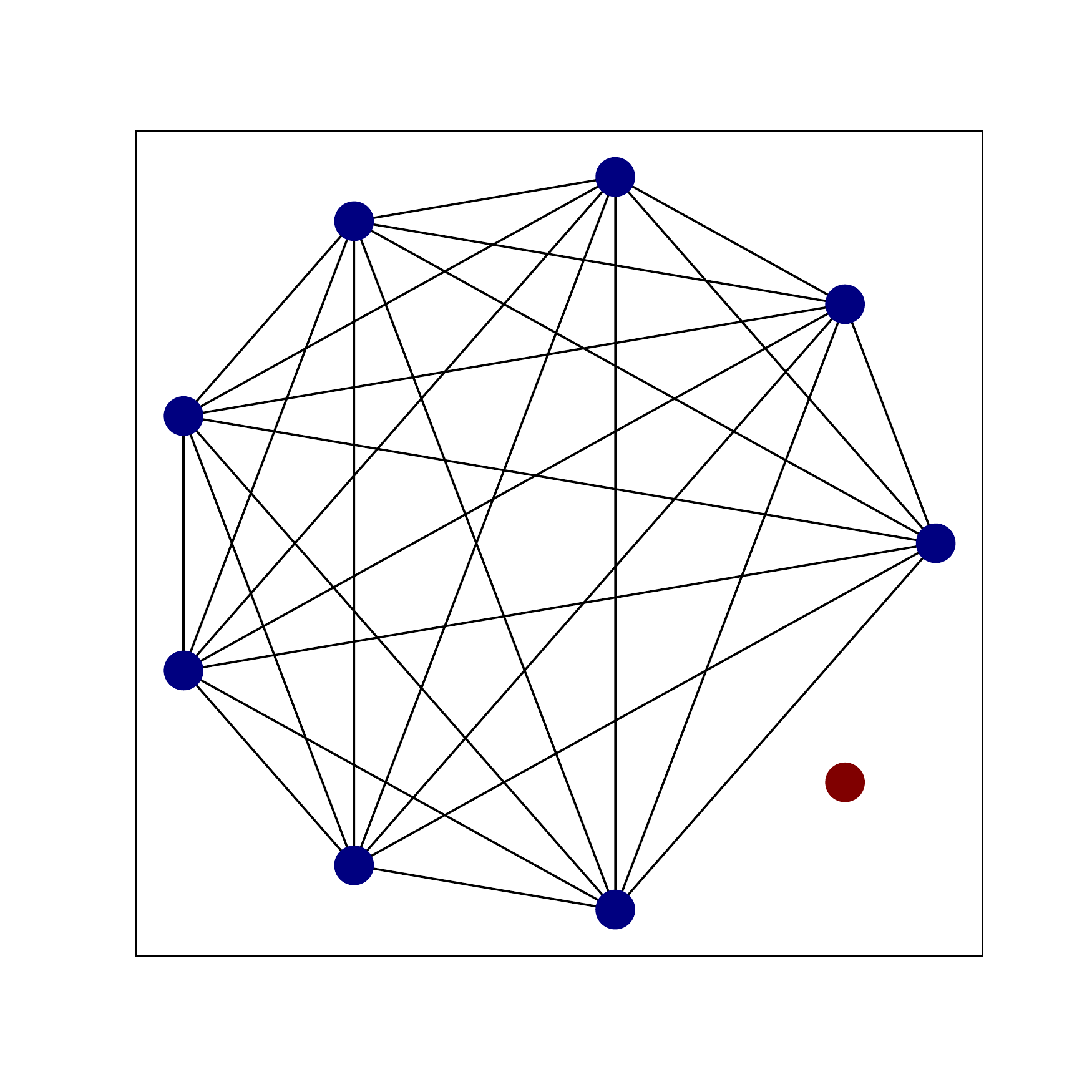} &
		\includegraphics[width=0.25\columnwidth,keepaspectratio]{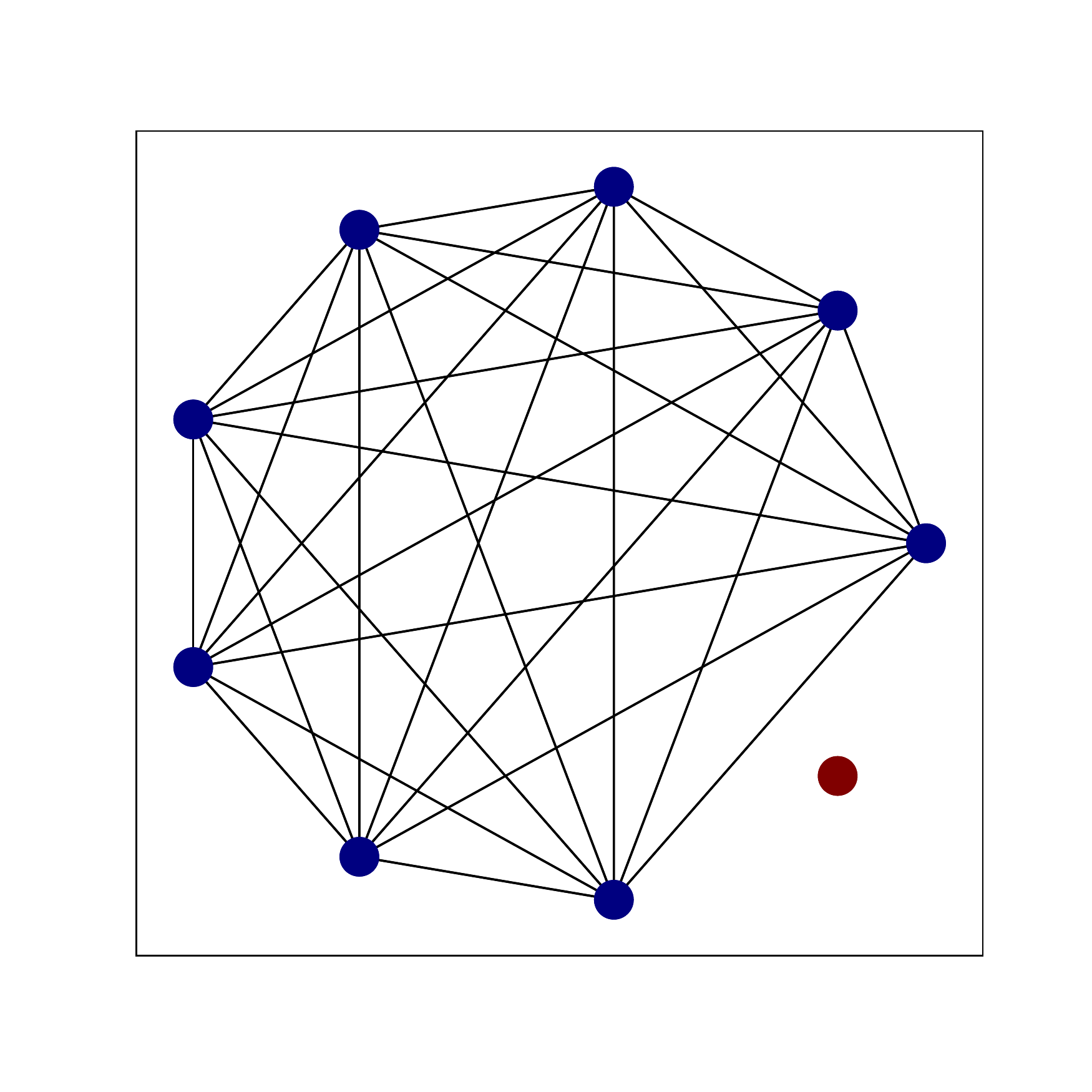} &
		\includegraphics[width=0.25\columnwidth,keepaspectratio]{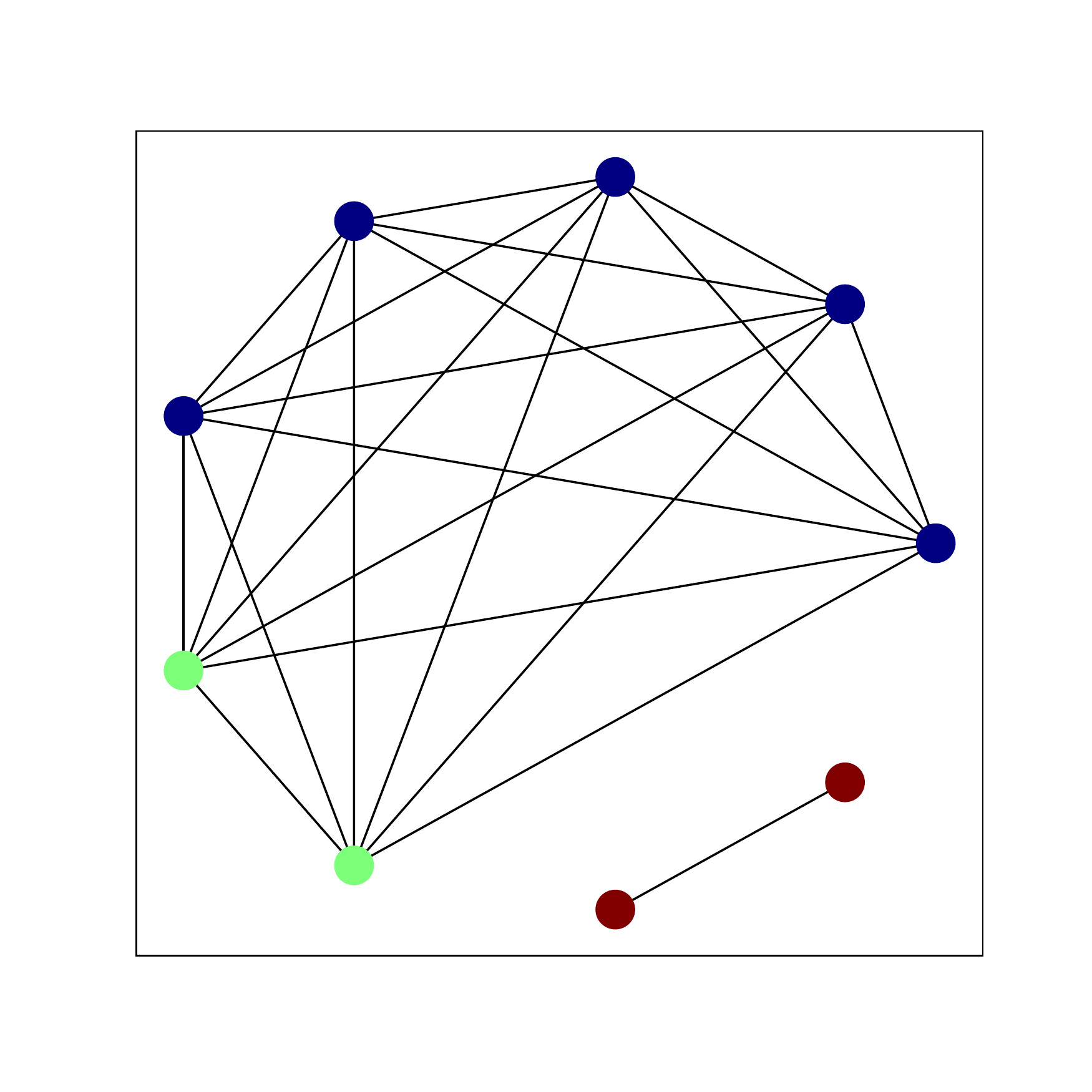} &
		\includegraphics[width=0.25\columnwidth,keepaspectratio]{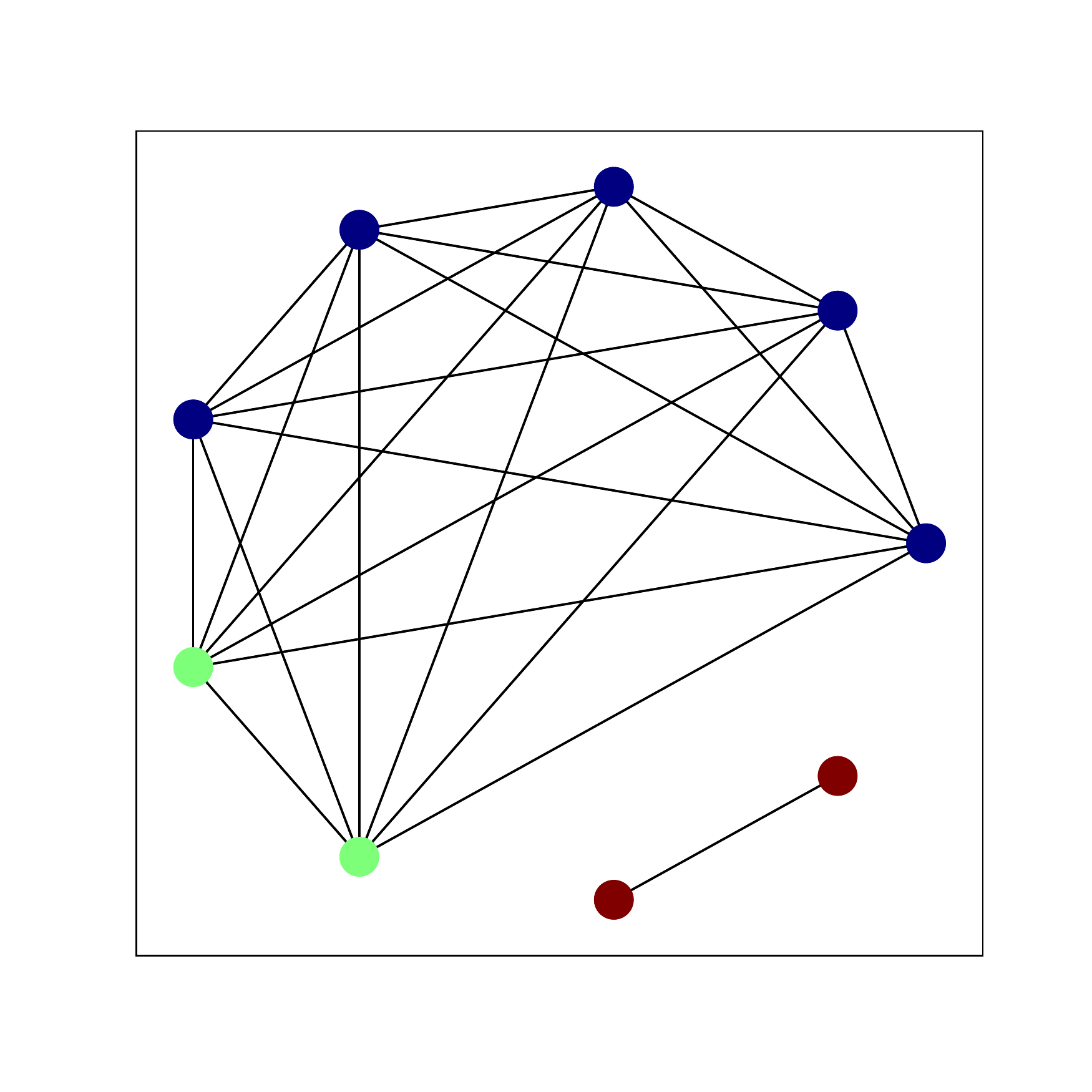} \\
		
		\includegraphics[width=0.25\columnwidth,keepaspectratio]{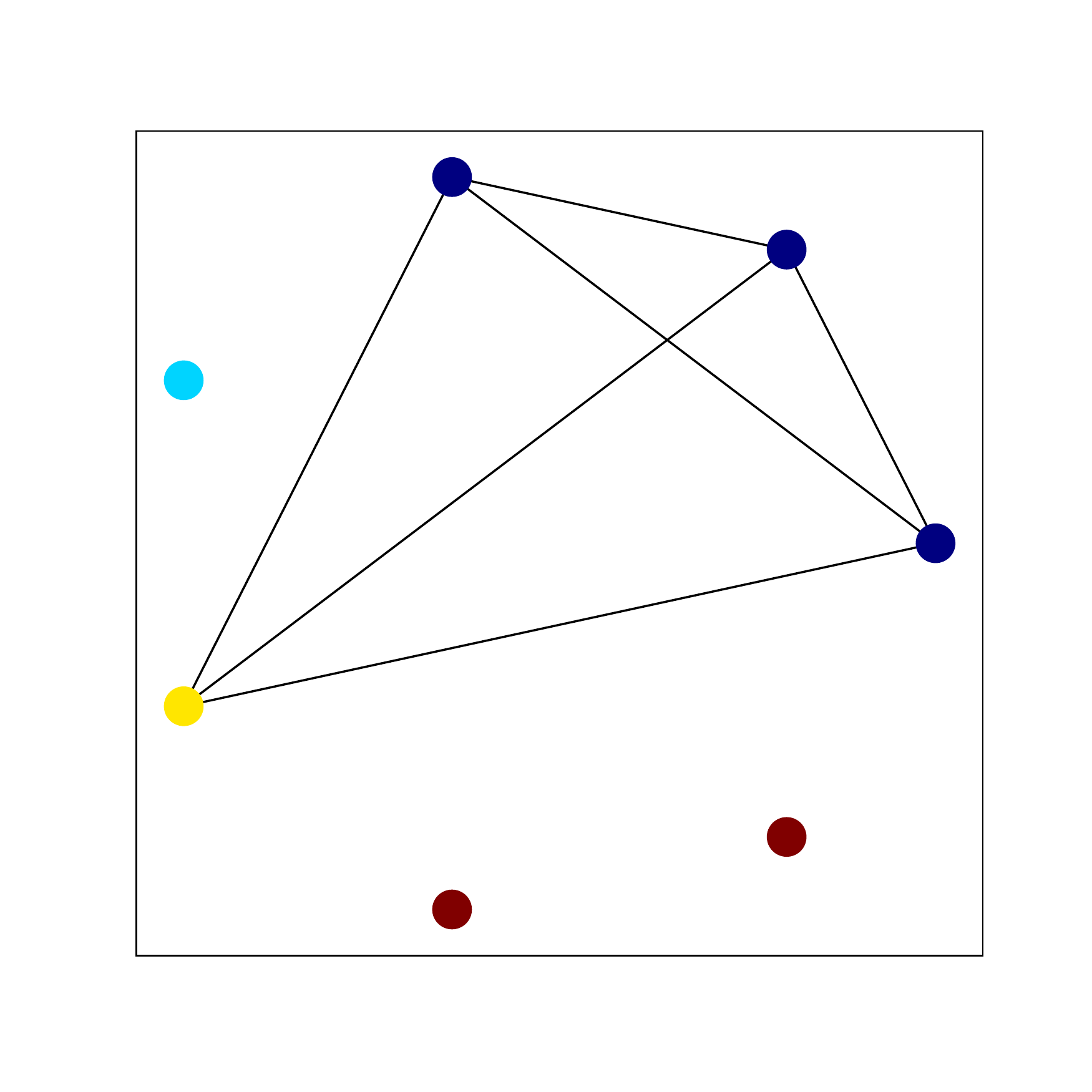} &
		\includegraphics[width=0.25\columnwidth,keepaspectratio]{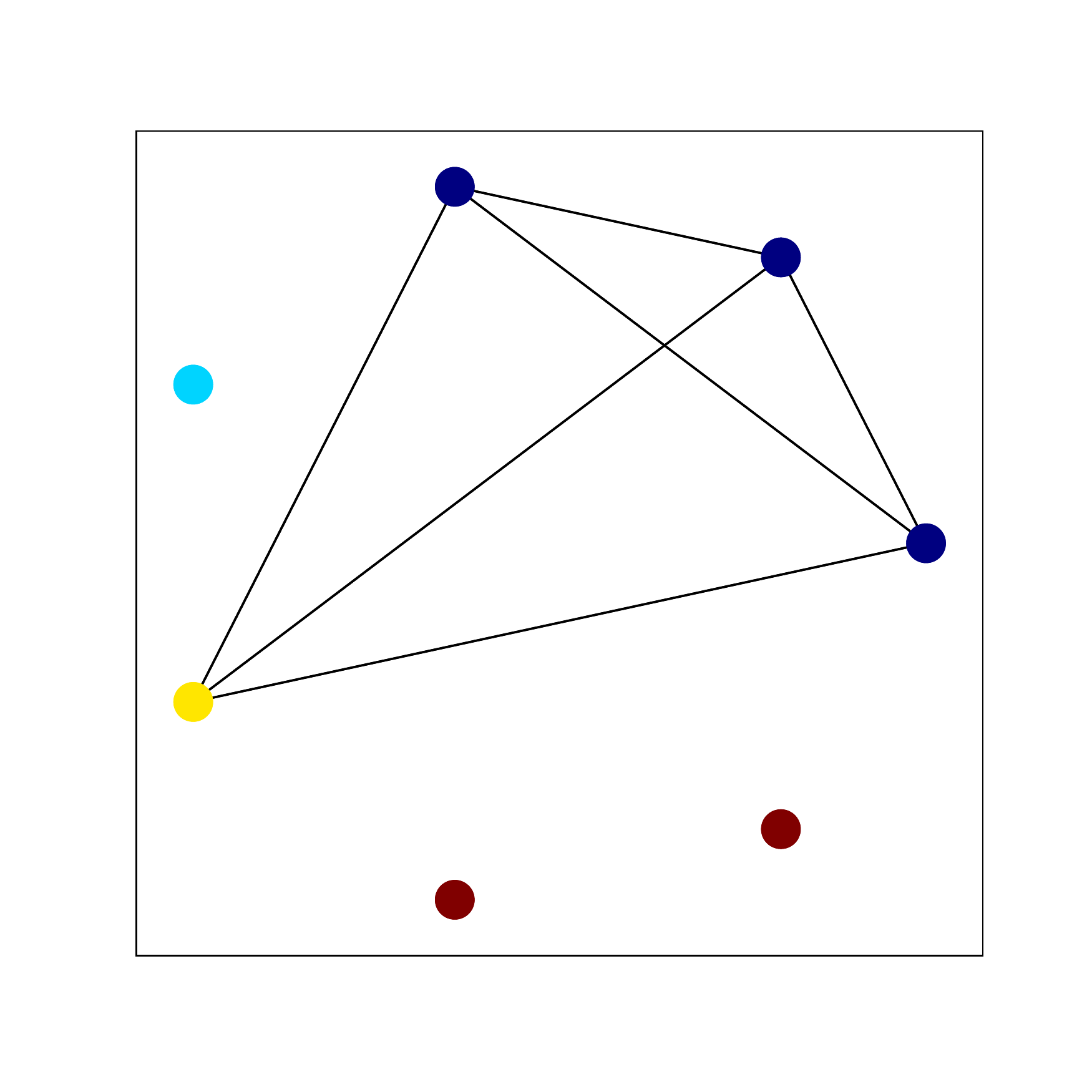} &
		\includegraphics[width=0.25\columnwidth,keepaspectratio]{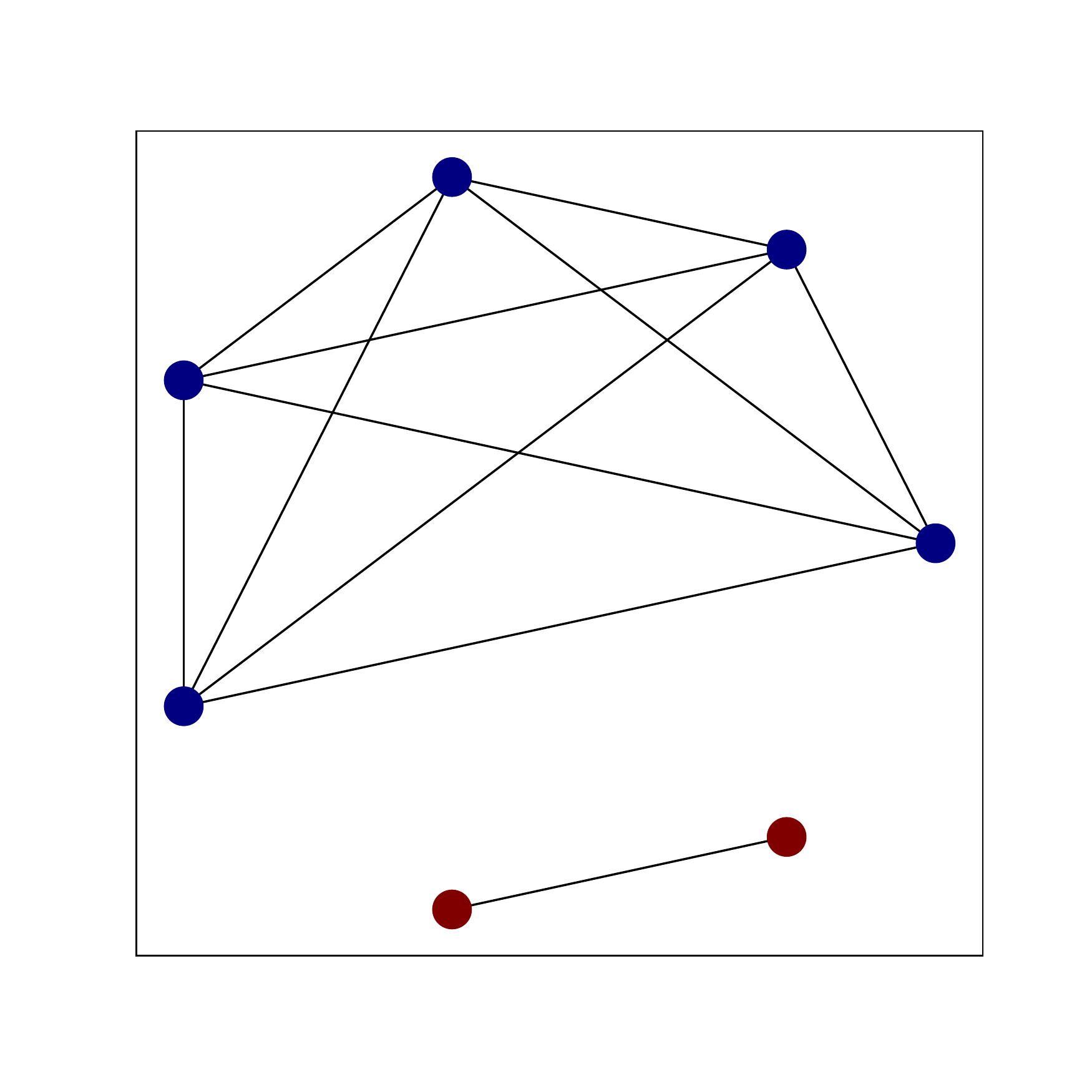} &
		\includegraphics[width=0.25\columnwidth,keepaspectratio]{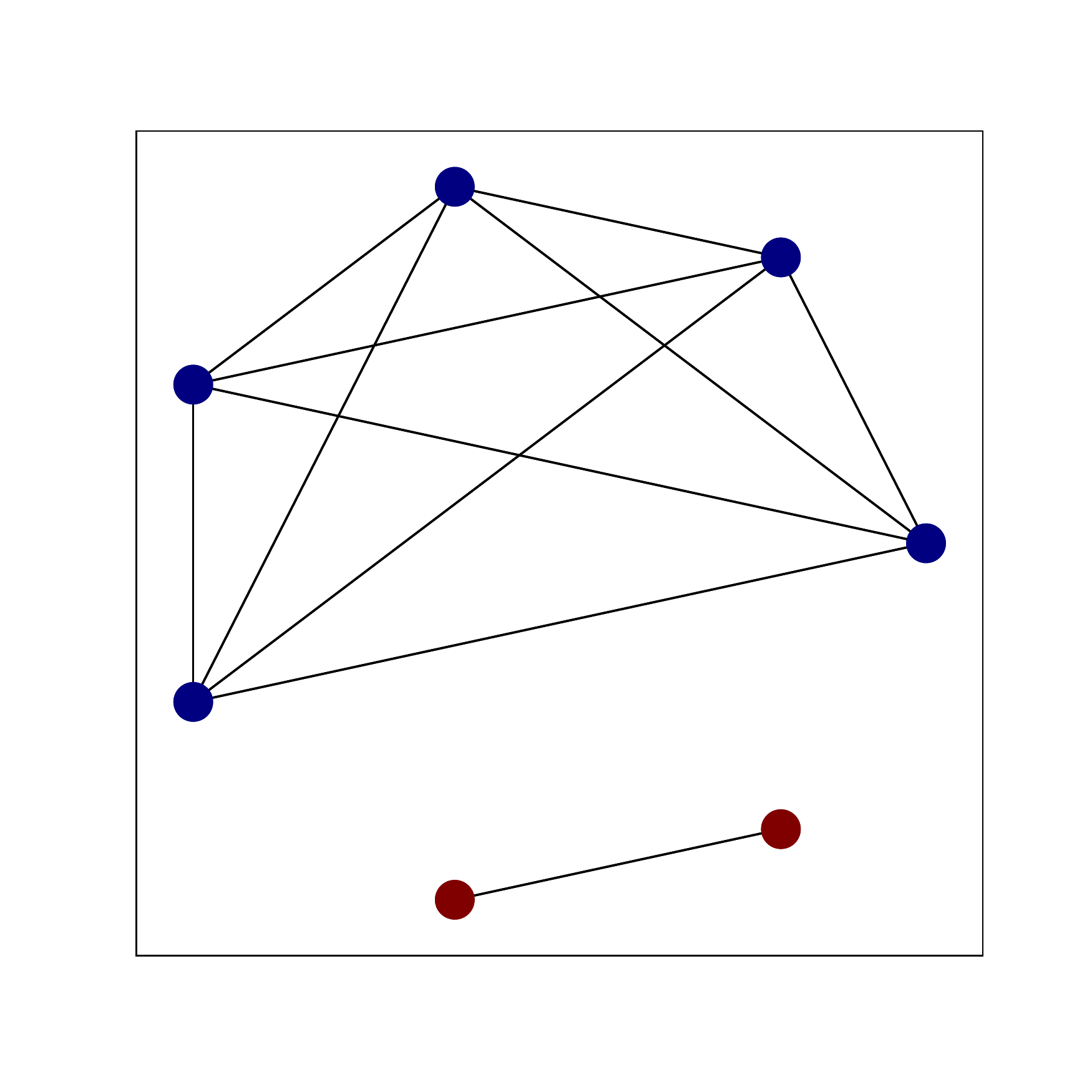} \\
		
		\includegraphics[width=0.25\columnwidth,keepaspectratio]{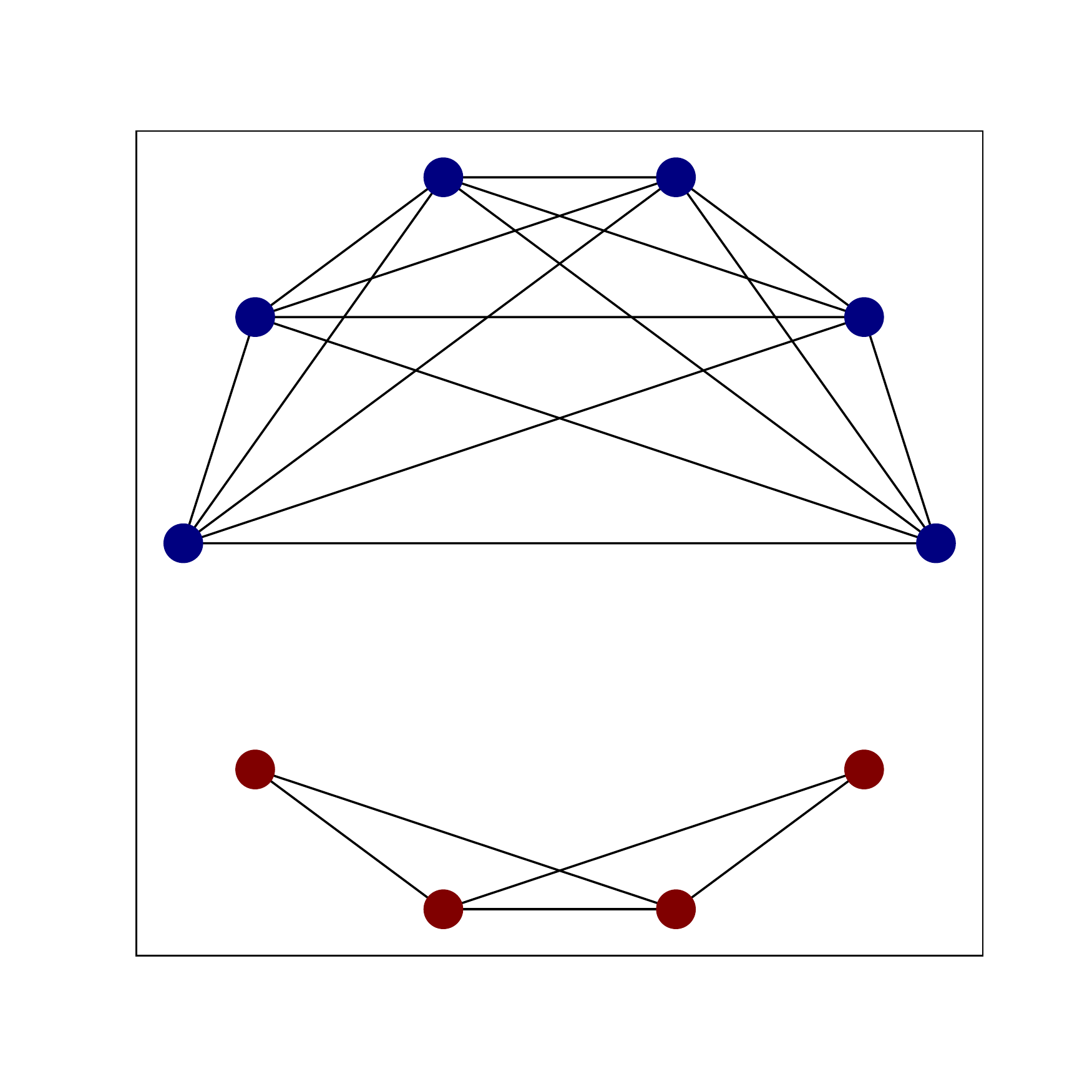} &
		\includegraphics[width=0.25\columnwidth,keepaspectratio]{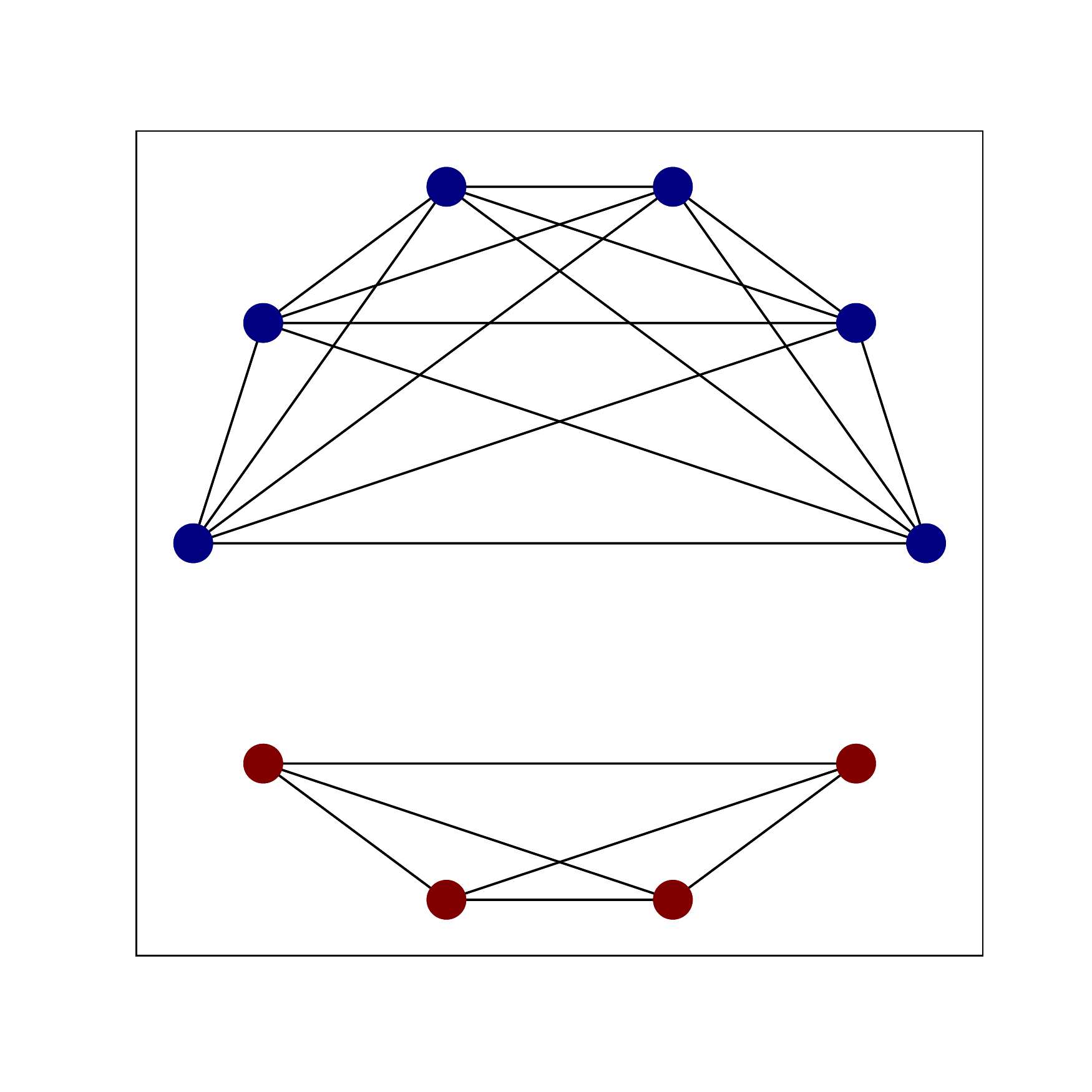} &
		\includegraphics[width=0.25\columnwidth,keepaspectratio]{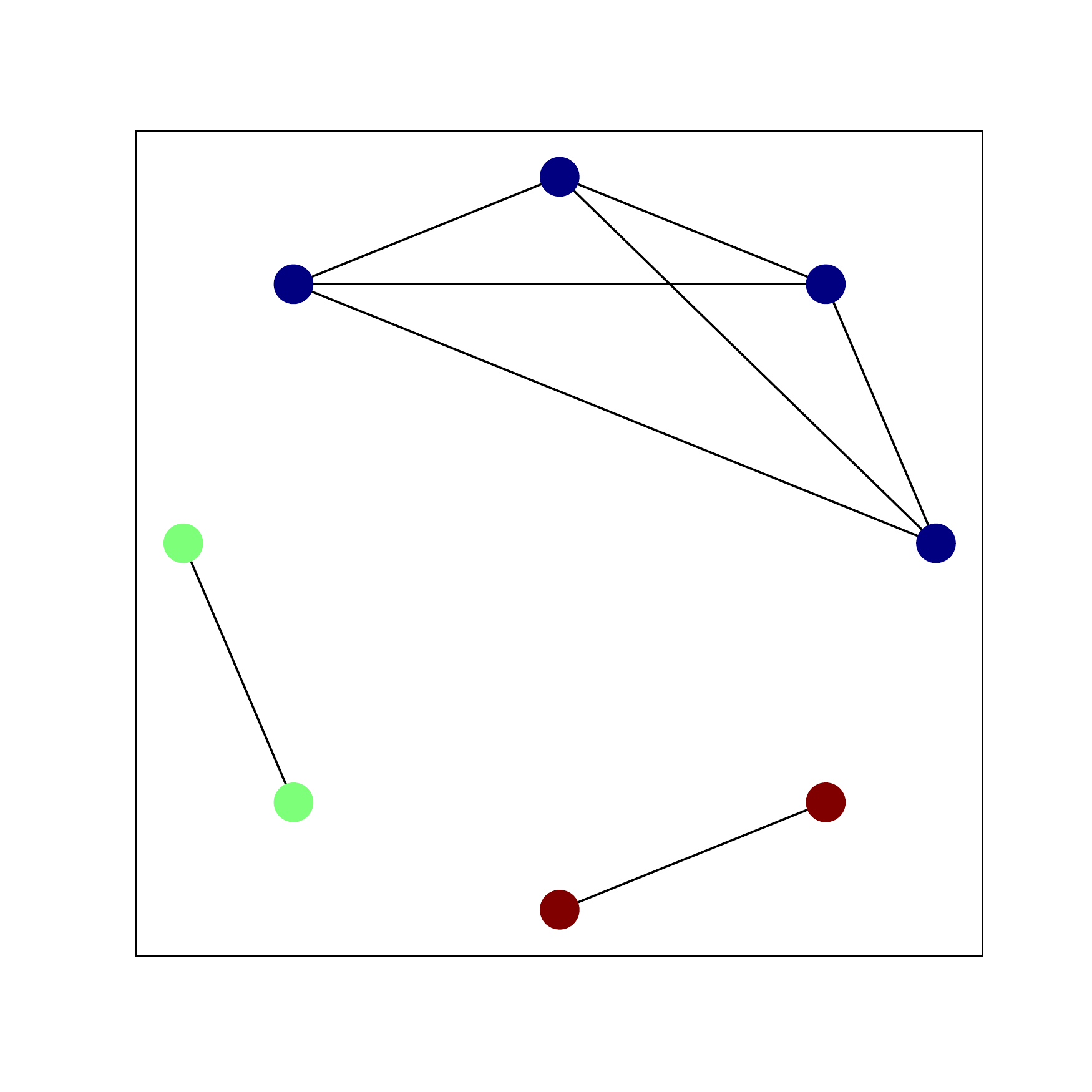} &
		\includegraphics[width=0.25\columnwidth,keepaspectratio]{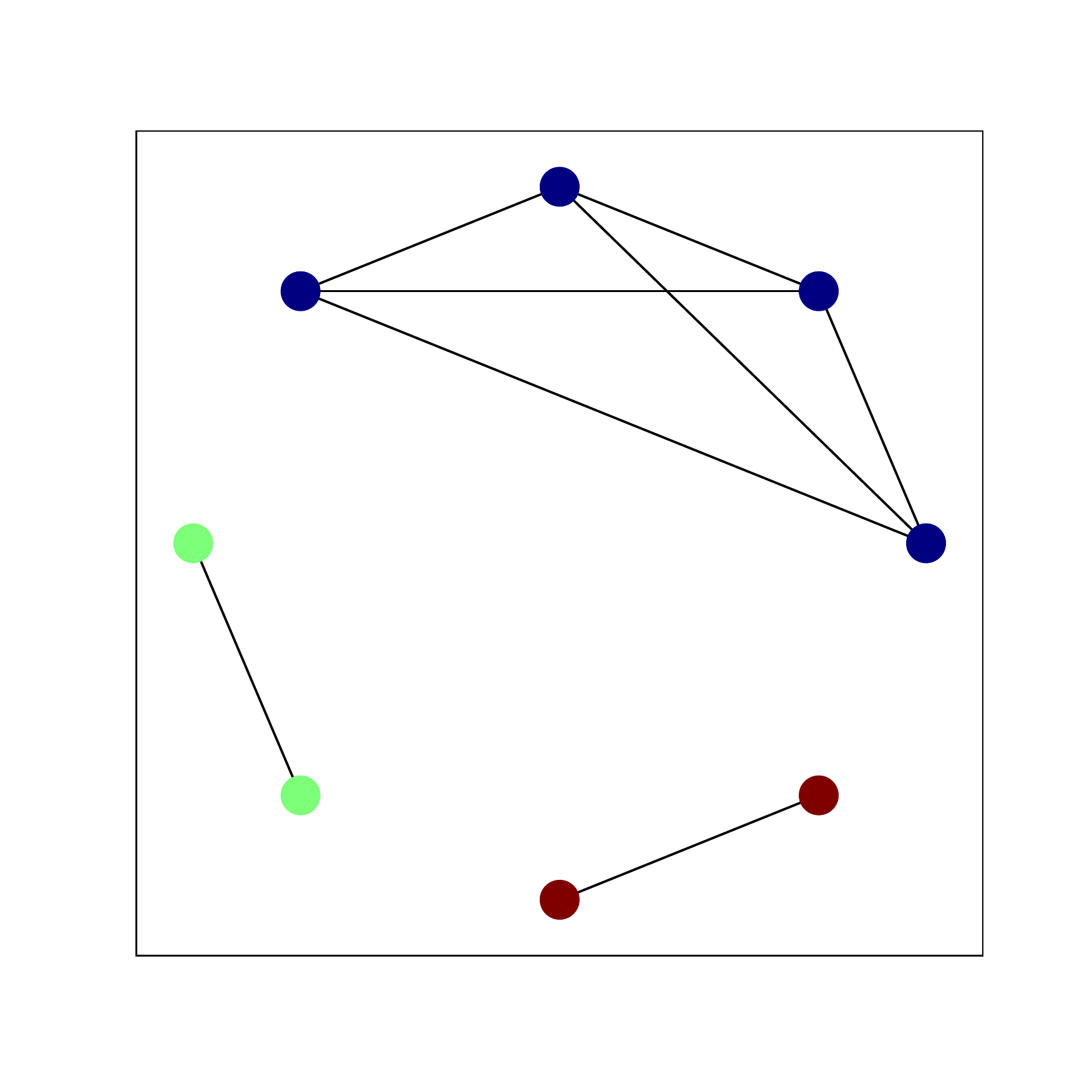} \\
		
		\includegraphics[width=0.25\columnwidth,keepaspectratio]{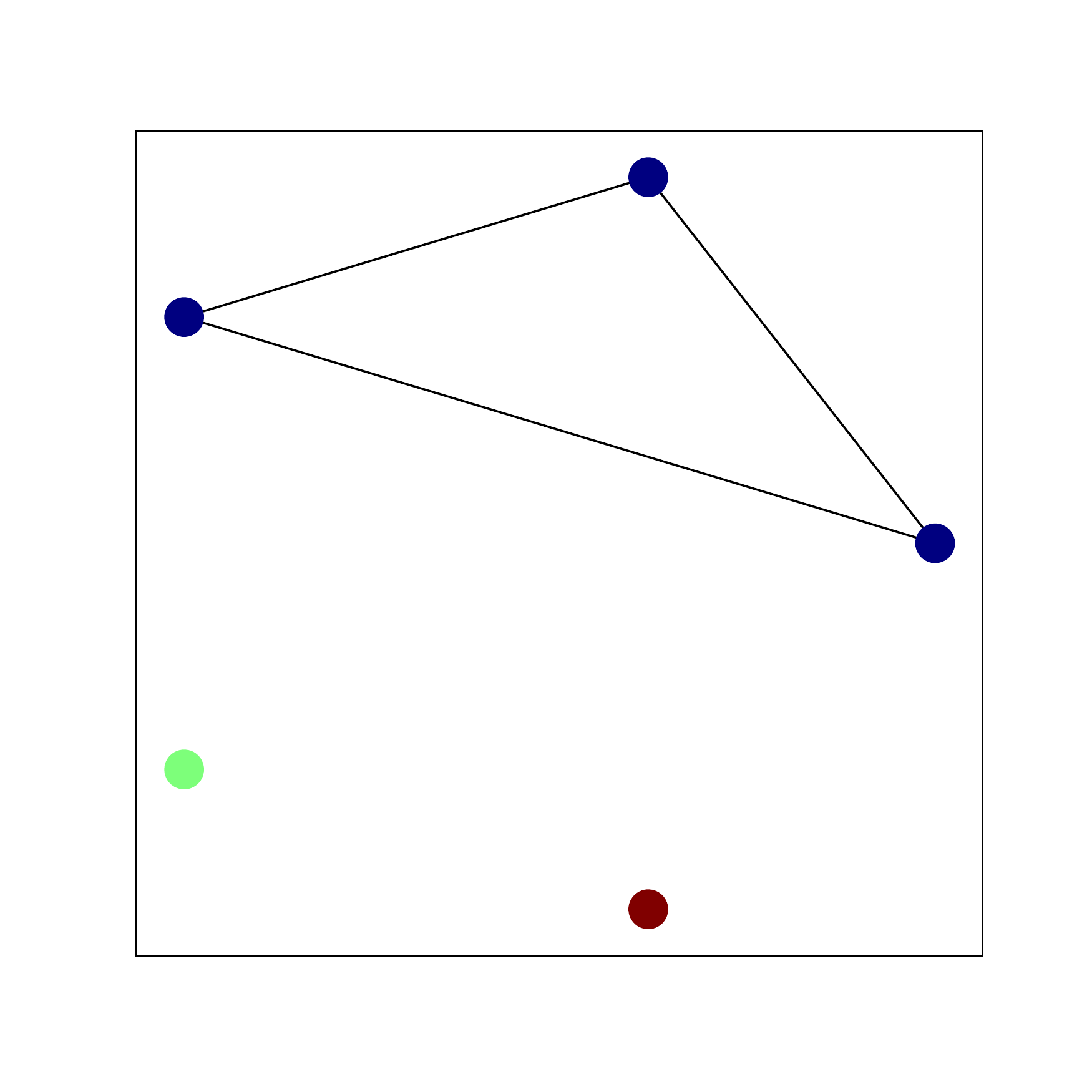} &
		\includegraphics[width=0.25\columnwidth,keepaspectratio]{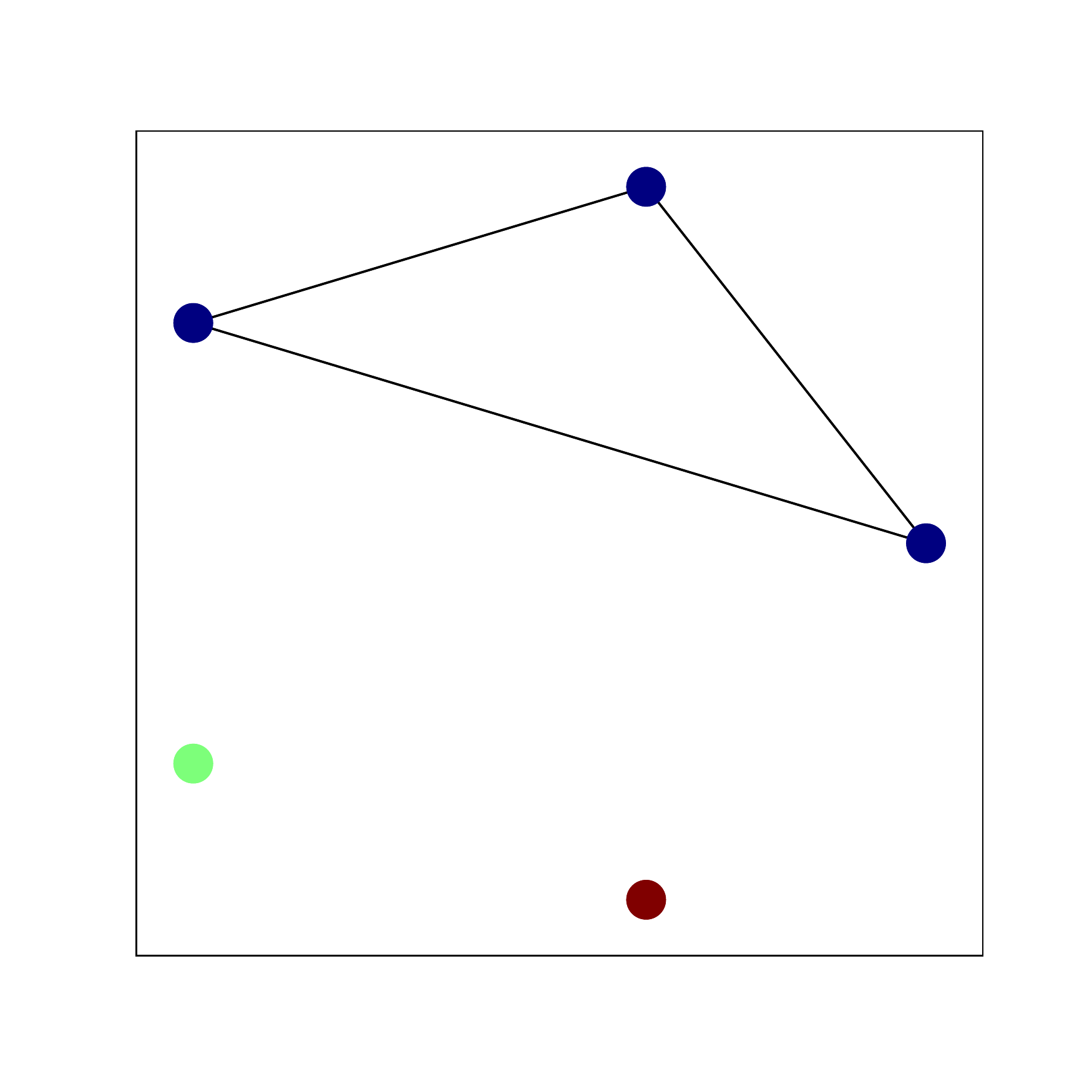} &

		\includegraphics[width=0.25\columnwidth,keepaspectratio]{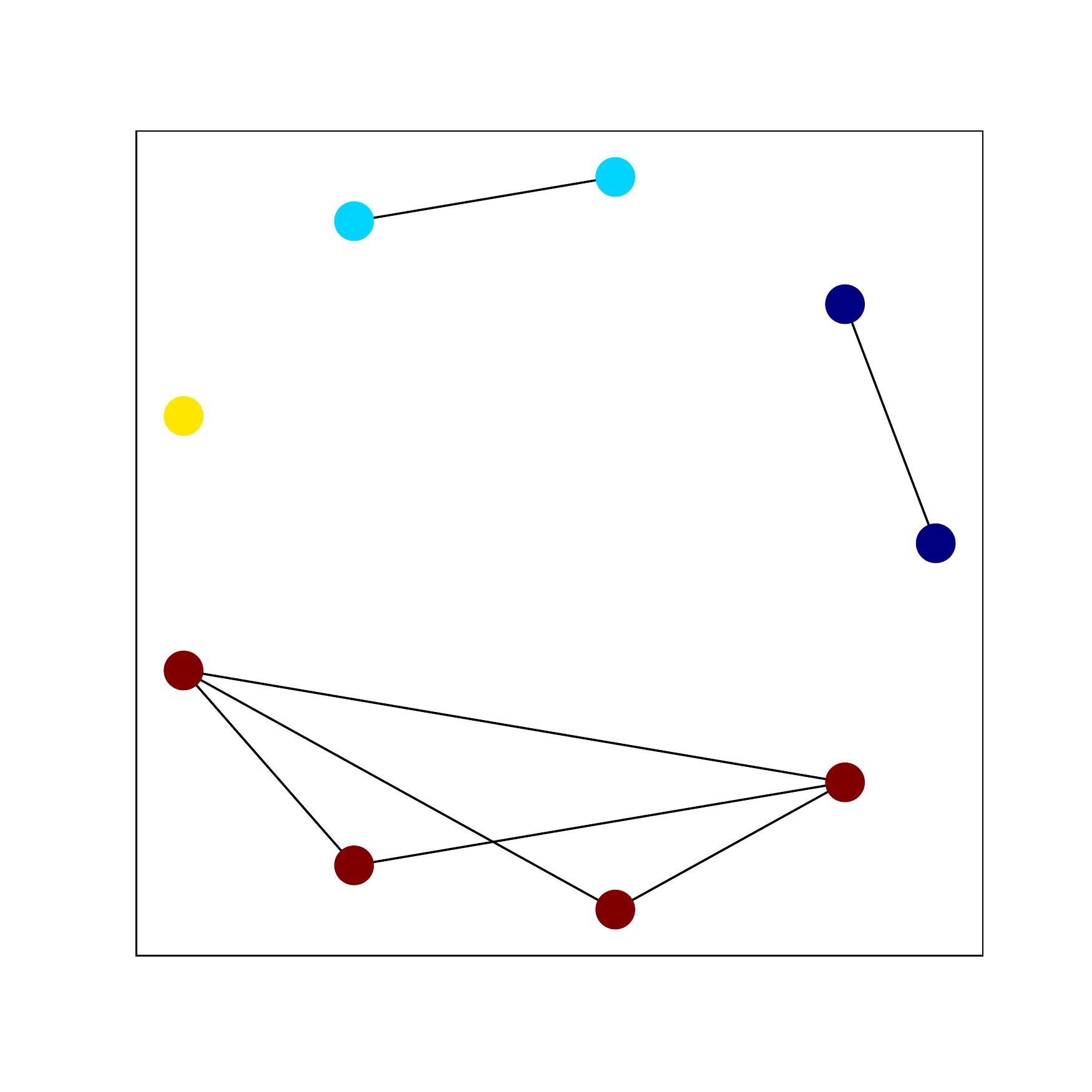} &
		\includegraphics[width=0.25\columnwidth,keepaspectratio]{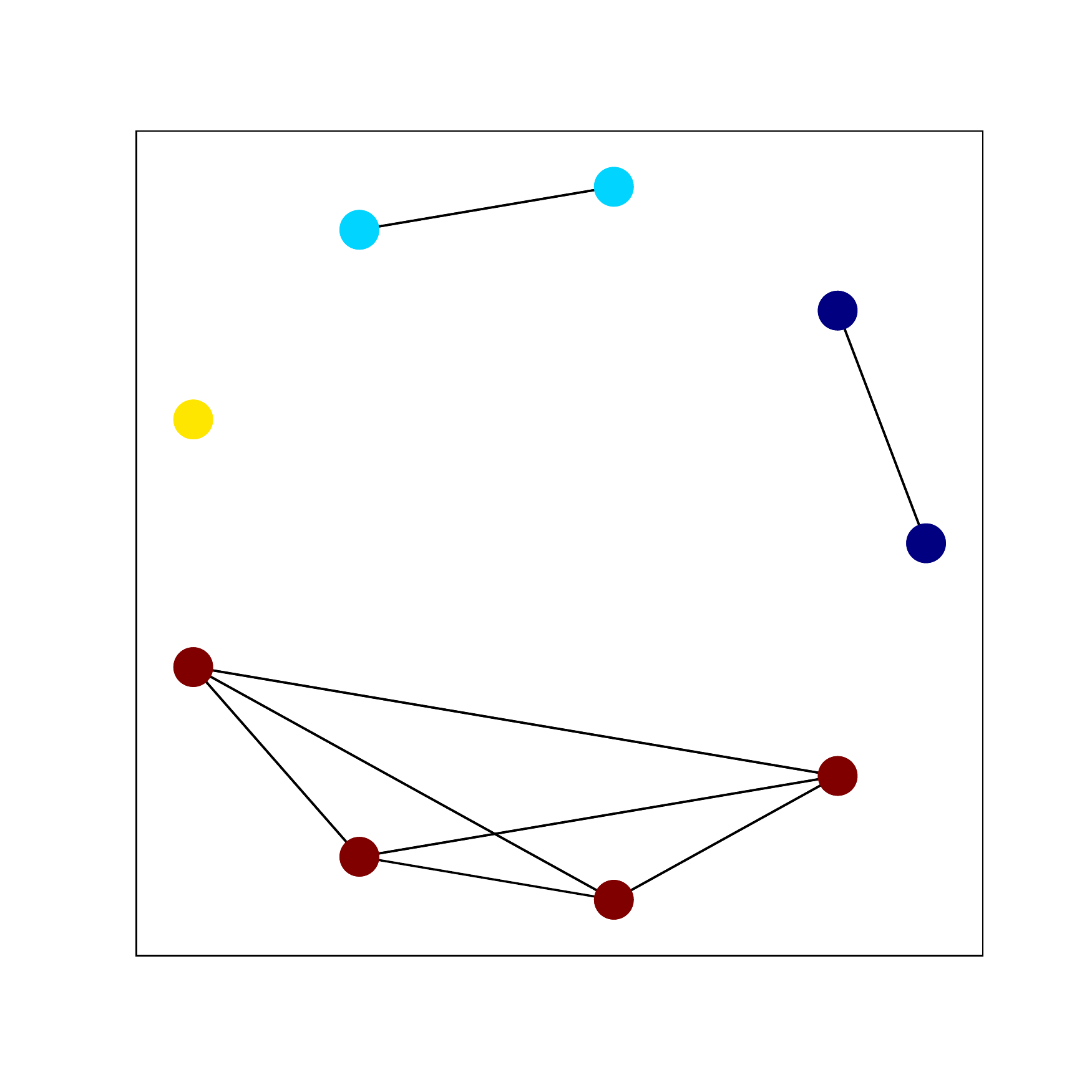} \\

		\includegraphics[width=0.25\columnwidth,keepaspectratio]{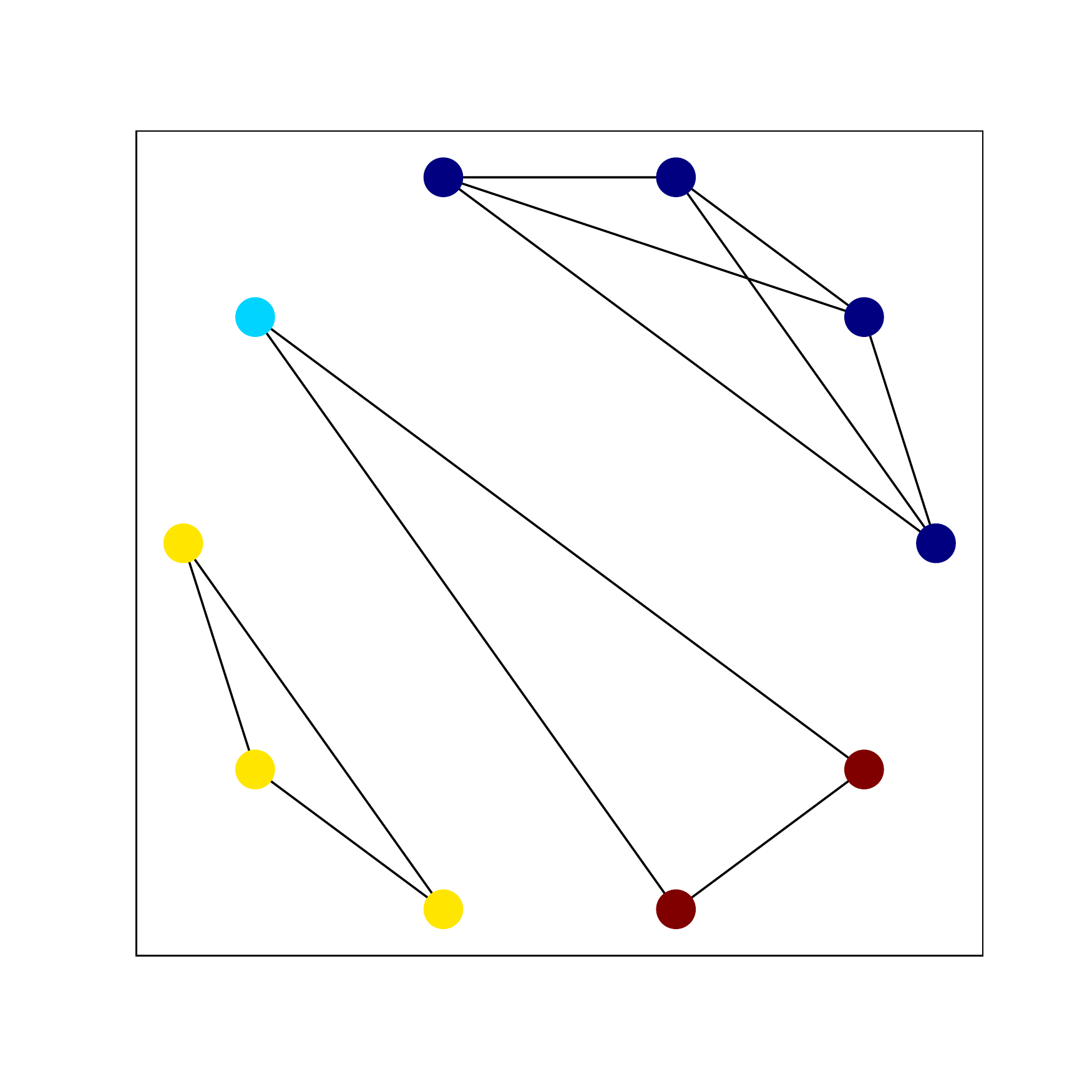} &
		\includegraphics[width=0.25\columnwidth,keepaspectratio]{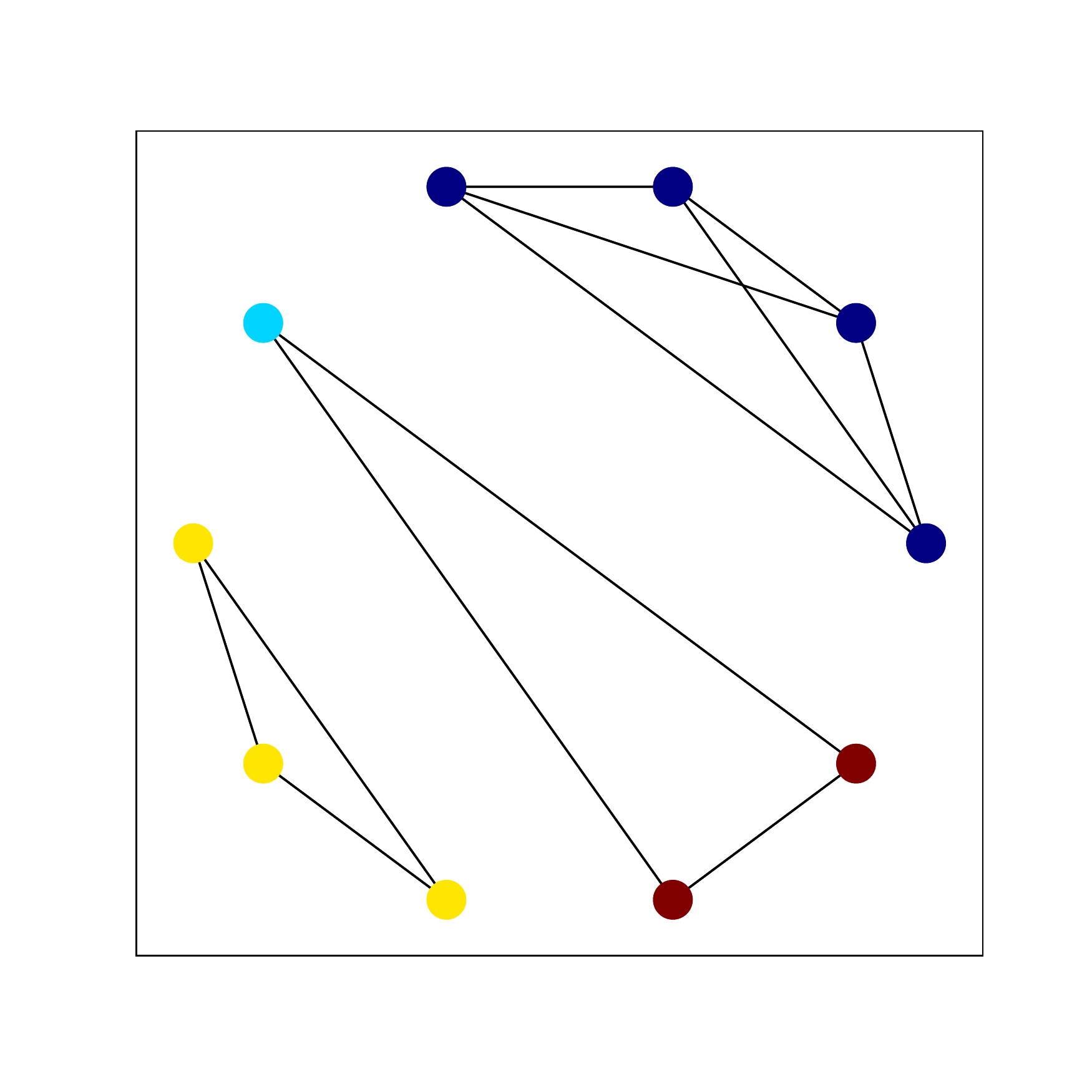} &
		\includegraphics[width=0.25\columnwidth,keepaspectratio]{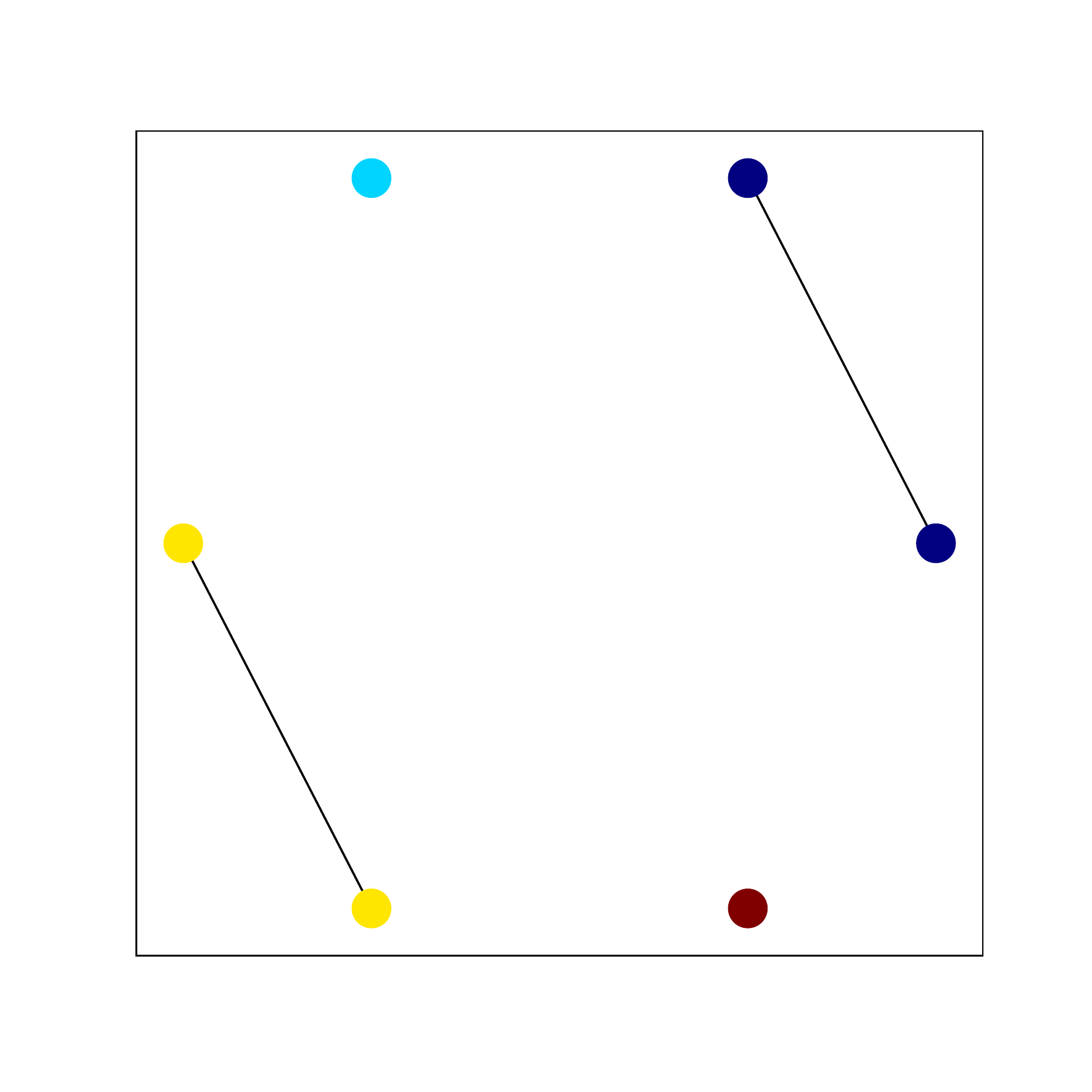} &
		\includegraphics[width=0.25\columnwidth,keepaspectratio]{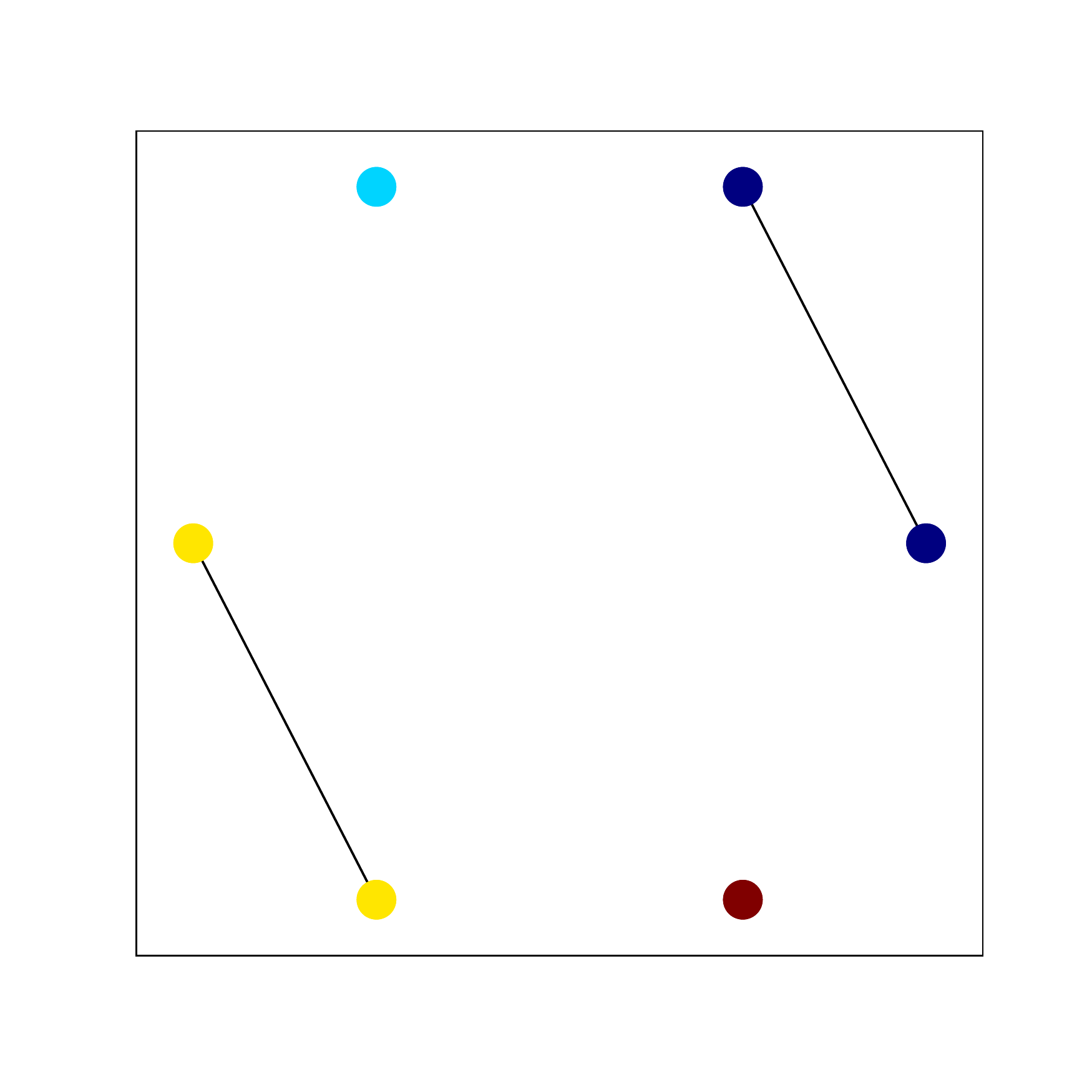} \\
		
		\includegraphics[width=0.25\columnwidth,keepaspectratio]{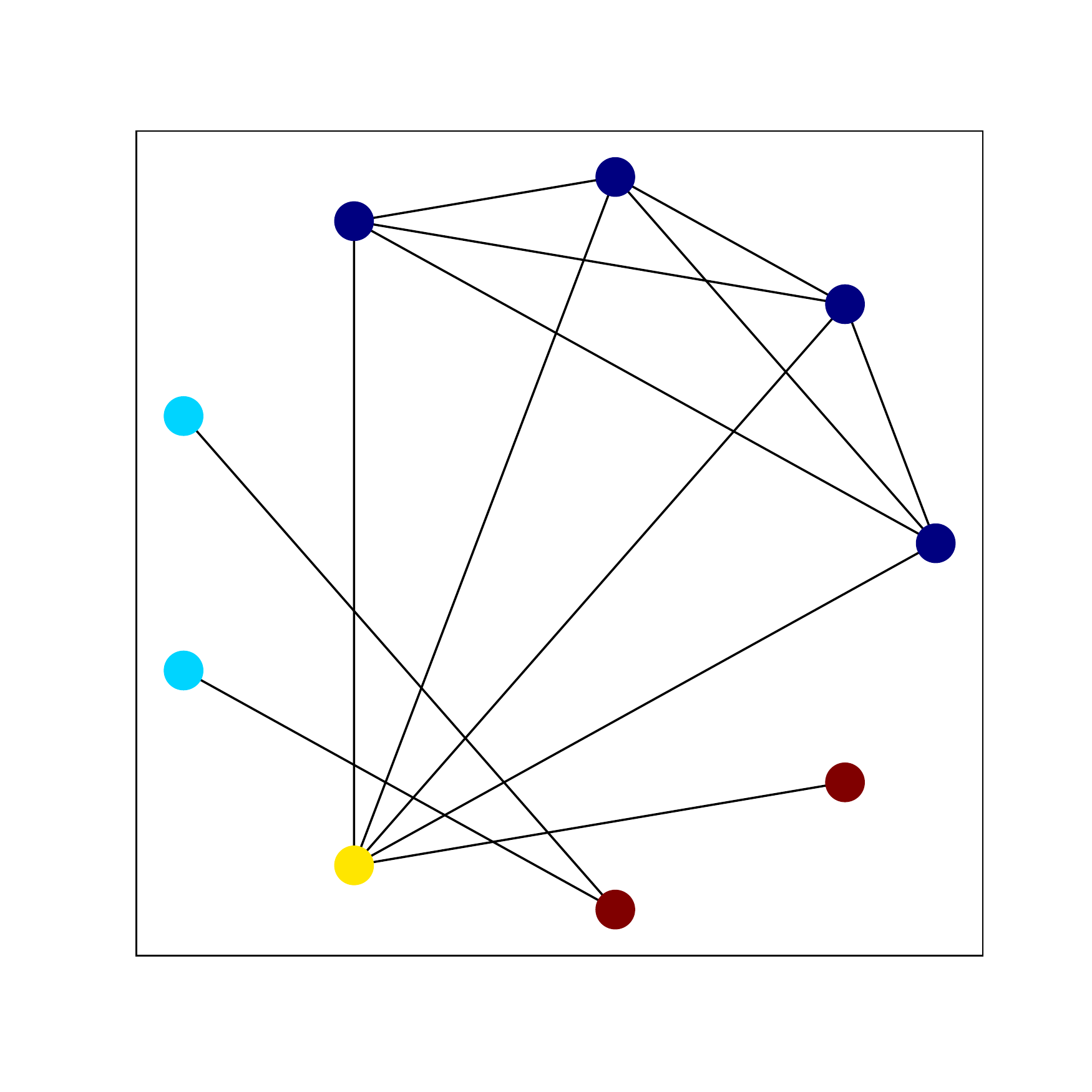} &
		\includegraphics[width=0.25\columnwidth,keepaspectratio]{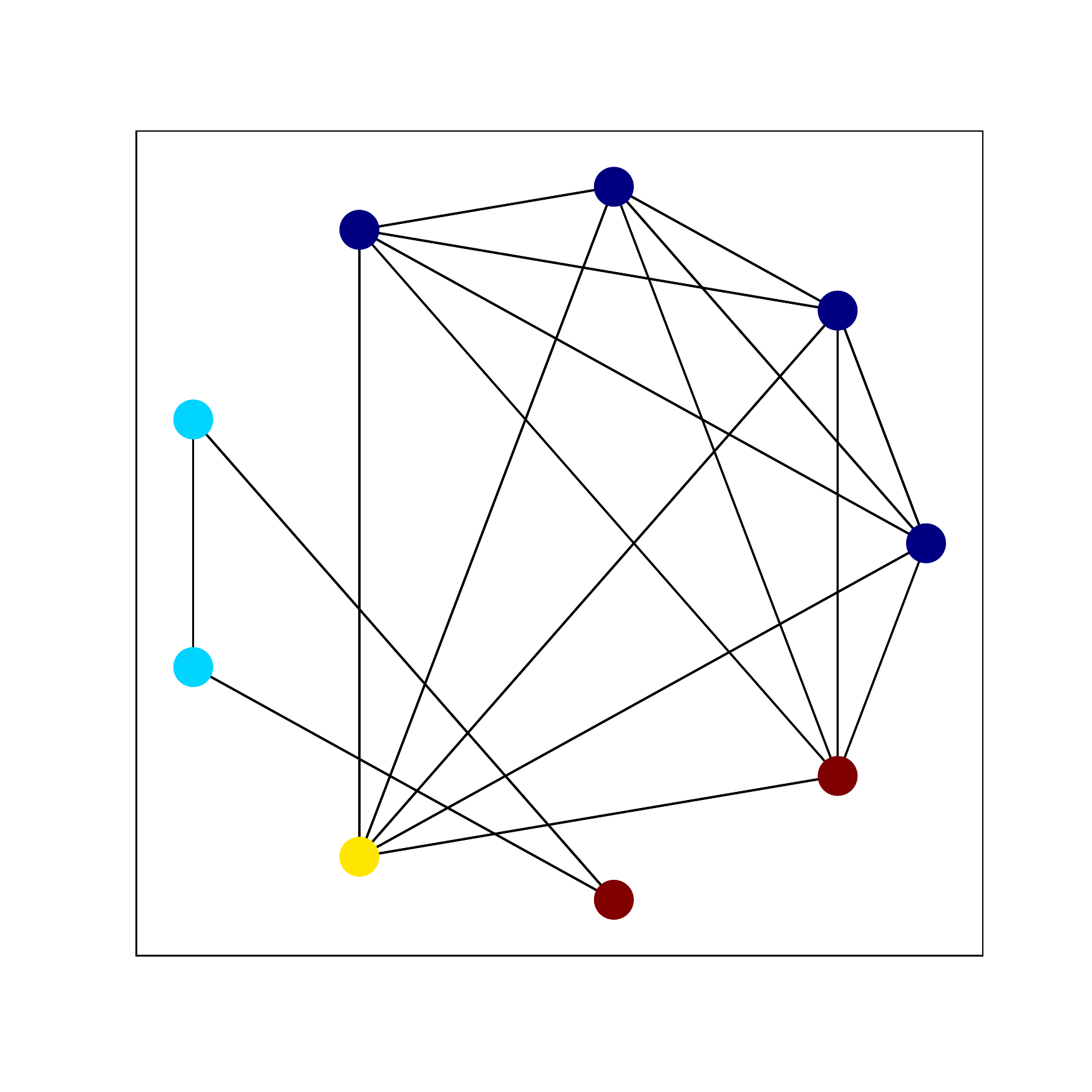} &
		\includegraphics[width=0.25\columnwidth,keepaspectratio]{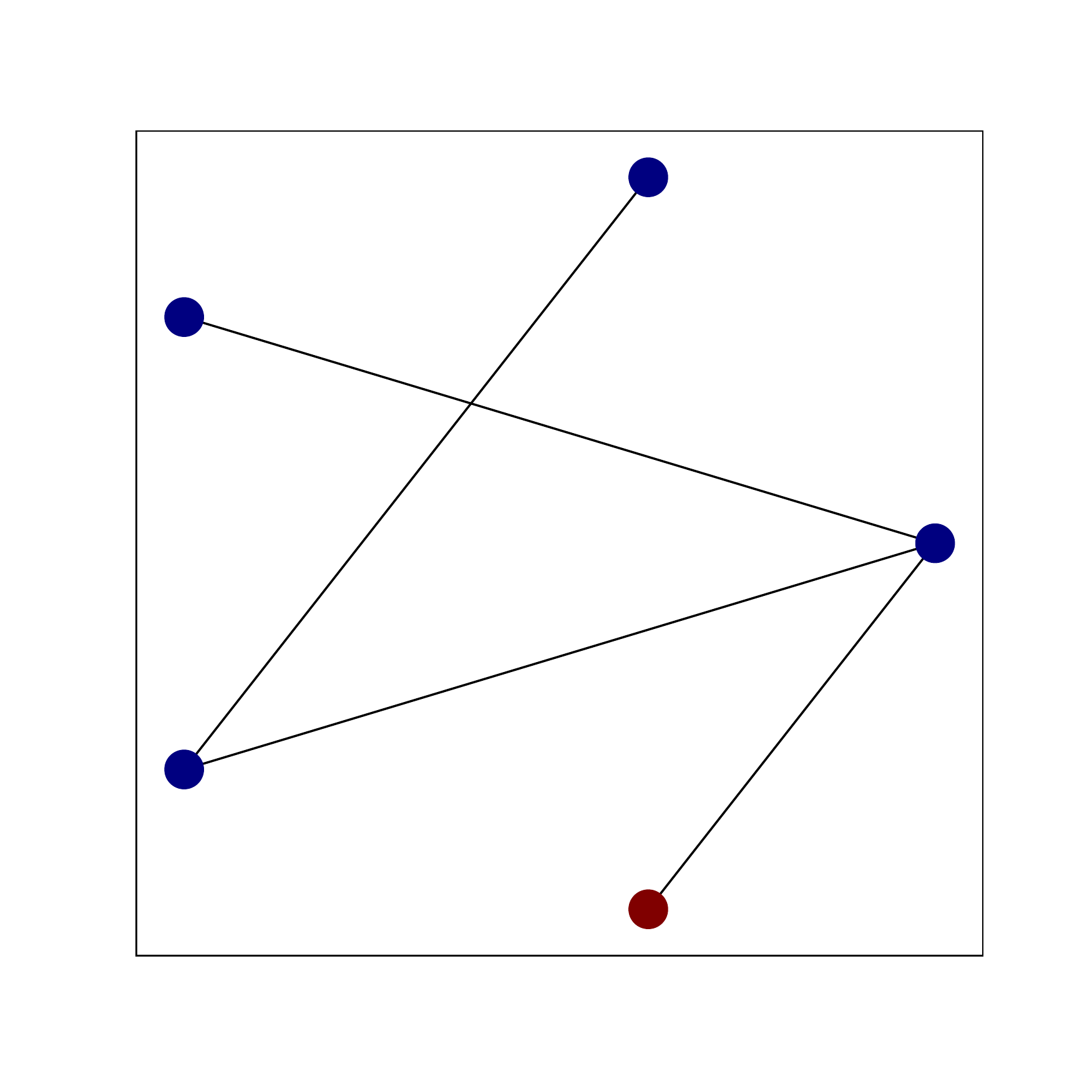} &
		\includegraphics[width=0.25\columnwidth,keepaspectratio]{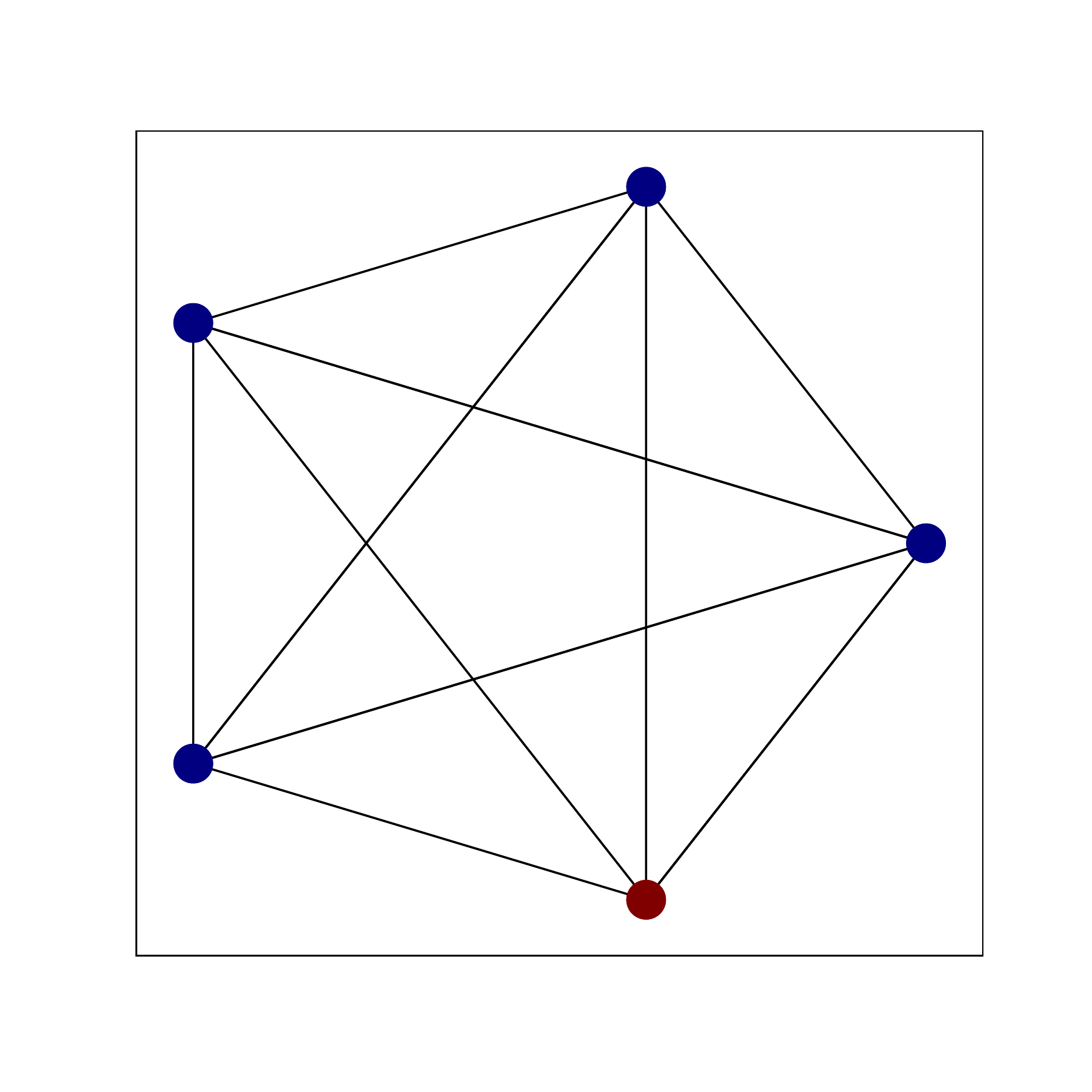} \\
		
	\end{tabular}\vspace{-10pt}
	\caption{Jets qualitative results. For each pair, the left side is before completing edges to a connected-component graph. The color of the vertices refer to the GT cluster. The edges are predicted by the model. }
	\label{fig:infered_jets}
\end{figure}

The dataset is made of training, validation and test set with 543544, 181181 and 181182 instances accordingly. Each of the sets contains all three flavors: bottom, charm and light jets roughly in the same amount, while the flavor of each instance is not part of the input.
We repeat the following evaluation 11 times: (1) training over the dataset, early-stopping when the F1 score over the validation set does not improve for 20 epochs. (2) Predicting the clusters of the test set. (3) Separate the 3 flavors and calculate the metrics for each flavor. Eventually, we have 11 scores for each combination of metrics, flavor and model, and we report the mean$\pm$std. Note that the AVR is evaluated only once since it is not a stochastic algorithm.
We also conducted an ablation study for this experiment, the results for all particle types can be found in Table \ref{tab:ablation_full}.
Examples of inferred graphs can be seen in Figure \ref{fig:infered_jets}. Since edges are predicted, there's a need to project the inferred graph to a connected-components graph. The results show that in most cases, the inferred graphs are already predicted that way.

\begin{table}
	
	\small
	\begin{center}
		\begin{tabular}{l|cccccc|ccc}
			\hline\hline
         & Method & $\vpsi$ \#layers & $\vphi$ & $\vphi$ \#layers & $d_1$ & Attention & F1 & RI & ARI \\
\hline
b jets & S2G & 2 & DeepSets & 5 & 2 & V & 0.649 & 0.736 & 0.493 \\
 & S2G & 2 & DeepSets & 5 & 10 & V & 0.642 & 0.739 & 0.488 \\
 & S2G+ & 2 & DeepSets & 5 & 5 & V & 0.658 & 0.745 & 0.510 \\
 & S2G & 2 & Siamese & 5 & 5 & V & 0.605 & 0.671 & 0.408 \\
 & S2G+ & 2 & DeepSets & 4 & 5 & V & 0.649 & 0.733 & 0.493 \\
 & S2G+ & 2 & DeepSets & 5 & 2 & V & 0.642 & 0.732 & 0.484 \\
 & S2G+ & 2 & DeepSets & 6 & 5 & V & 0.654 & 0.739 & 0.502 \\
 & S2G & 2 & DeepSets & 5 & 5 & X & 0.640 & 0.726 & 0.478 \\
 & S2G & 2 & DeepSets & 4 & 5 & V & 0.649 & 0.741 & 0.498 \\
 & S2G & 2 & PointNetSeg & 5 & 5 & V & 0.630 & 0.720 & 0.462 \\
 & S2G & 2 & DeepSets & 5 & 5 & V & 0.646 & 0.739 & 0.495 \\
 & QUAD & 1 & DeepSets & 5 & 5 & V & 0.637 & 0.730 & 0.470 \\
 & S2G+ & 2 & DeepSets & 5 & 10 & V & 0.655 & 0.749 & 0.510 \\
 & S2G & 2 & DeepSets & 6 & 5 & V & \textbf{0.660} & \textbf{0.753} & \textbf{0.516} \\
 & S2G+ & 2 & Siamese & 5 & 5 & V & 0.438 & 0.303 & 0.026 \\
 & S2G+ & 2 & DeepSets & 5 & 5 & X & 0.643 & 0.729 & 0.482 \\
 & S2G & 1 & DeepSets & 5 & 5 & V & 0.565 & 0.710 & 0.395 \\
 & S2G+ & 2 & PointNetSeg & 5 & 5 & V & 0.619 & 0.717 & 0.451 \\
 & S2G+ & 1 & DeepSets & 5 & 5 & V & 0.577 & 0.717 & 0.414 \\
\hline
c jets & S2G & 2 & DeepSets & 5 & 2 & V & 0.749 & 0.727 & 0.458 \\
 & S2G & 2 & DeepSets & 5 & 10 & V & 0.747 & 0.729 & 0.459 \\
 & S2G+ & 2 & DeepSets & 5 & 5 & V & 0.753 & 0.732 & 0.467 \\
 & S2G & 2 & Siamese & 5 & 5 & V & 0.728 & 0.693 & 0.404 \\
 & S2G+ & 2 & DeepSets & 4 & 5 & V & 0.748 & 0.726 & 0.456 \\
 & S2G+ & 2 & DeepSets & 5 & 2 & V & 0.749 & 0.726 & 0.457 \\
 & S2G+ & 2 & DeepSets & 6 & 5 & V & 0.750 & 0.729 & 0.462 \\
 & S2G & 2 & DeepSets & 5 & 5 & X & 0.743 & 0.720 & 0.444 \\
 & S2G & 2 & DeepSets & 4 & 5 & V & 0.749 & 0.728 & 0.460 \\
 & S2G & 2 & PointNetSeg & 5 & 5 & V & 0.741 & 0.720 & 0.443 \\
 & S2G & 2 & DeepSets & 5 & 5 & V & 0.750 & 0.730 & 0.463 \\
 & QUAD & 1 & DeepSets & 5 & 5 & V & 0.750 & 0.734 & 0.469 \\
 & S2G+ & 2 & DeepSets & 5 & 10 & V & 0.752 & \textbf{0.735} & \textbf{0.470} \\
 & S2G & 2 & DeepSets & 6 & 5 & V & \textbf{0.754} & 0.734 & 0.470 \\
 & S2G+ & 2 & Siamese & 5 & 5 & V & 0.610 & 0.472 & 0.078 \\
 & S2G+ & 2 & DeepSets & 5 & 5 & X & 0.741 & 0.718 & 0.439 \\
 & S2G & 1 & DeepSets & 5 & 5 & V & 0.699 & 0.694 & 0.383 \\
 & S2G+ & 2 & PointNetSeg & 5 & 5 & V & 0.738 & 0.718 & 0.440 \\
 & S2G+ & 1 & DeepSets & 5 & 5 & V & 0.705 & 0.701 & 0.394 \\
\hline
light jets & S2G & 2 & DeepSets & 5 & 2 & V & 0.973 & 0.971 & 0.933 \\
 & S2G & 2 & DeepSets & 5 & 10 & V & 0.970 & 0.968 & 0.927 \\
 & S2G+ & 2 & DeepSets & 5 & 5 & V & 0.973 & 0.970 & 0.932 \\
 & S2G & 2 & Siamese & 5 & 5 & V & 0.973 & 0.970 & 0.926 \\
 & S2G+ & 2 & DeepSets & 4 & 5 & V & 0.973 & 0.971 & 0.933 \\
 & S2G+ & 2 & DeepSets & 5 & 2 & V & 0.974 & \textbf{0.972} & \textbf{0.935} \\
 & S2G+ & 2 & DeepSets & 6 & 5 & V & 0.972 & 0.970 & 0.931 \\
 & S2G & 2 & DeepSets & 5 & 5 & X & 0.973 & 0.971 & 0.931 \\
 & S2G & 2 & DeepSets & 4 & 5 & V & 0.972 & 0.970 & 0.930 \\
 & S2G & 2 & PointNetSeg & 5 & 5 & V & \textbf{0.974} & 0.971 & 0.933 \\
 & S2G & 2 & DeepSets & 5 & 5 & V & 0.972 & 0.970 & 0.931 \\
 & QUAD & 1 & DeepSets & 5 & 5 & V & 0.972 & 0.970 & 0.929 \\
 & S2G+ & 2 & DeepSets & 5 & 10 & V & 0.970 & 0.968 & 0.928 \\
 & S2G & 2 & DeepSets & 6 & 5 & V & 0.972 & 0.971 & 0.932 \\
 & S2G+ & 2 & Siamese & 5 & 5 & V & 0.910 & 0.867 & 0.675 \\
 & S2G+ & 2 & DeepSets & 5 & 5 & X & 0.973 & 0.971 & 0.933 \\
 & S2G & 1 & DeepSets & 5 & 5 & V & 0.968 & 0.969 & 0.926 \\
 & S2G+ & 2 & PointNetSeg & 5 & 5 & V & 0.973 & 0.972 & 0.934 \\
 & S2G+ & 1 & DeepSets & 5 & 5 & V & 0.966 & 0.967 & 0.923 \\
			\hline\hline
		\end{tabular}
	\end{center}
	\caption{Ablation study for particle partitioning.}
	\label{tab:ablation_full}
\end{table}

\begin{figure}
	\centering
	\setlength\tabcolsep{0.0pt} 
	\begin{tabular}{ccccc} 
	 \scriptsize truth & \scriptsize S2G  & \scriptsize S2G+ & \scriptsize baseline \\
			\includegraphics[width=0.25\columnwidth,keepaspectratio]{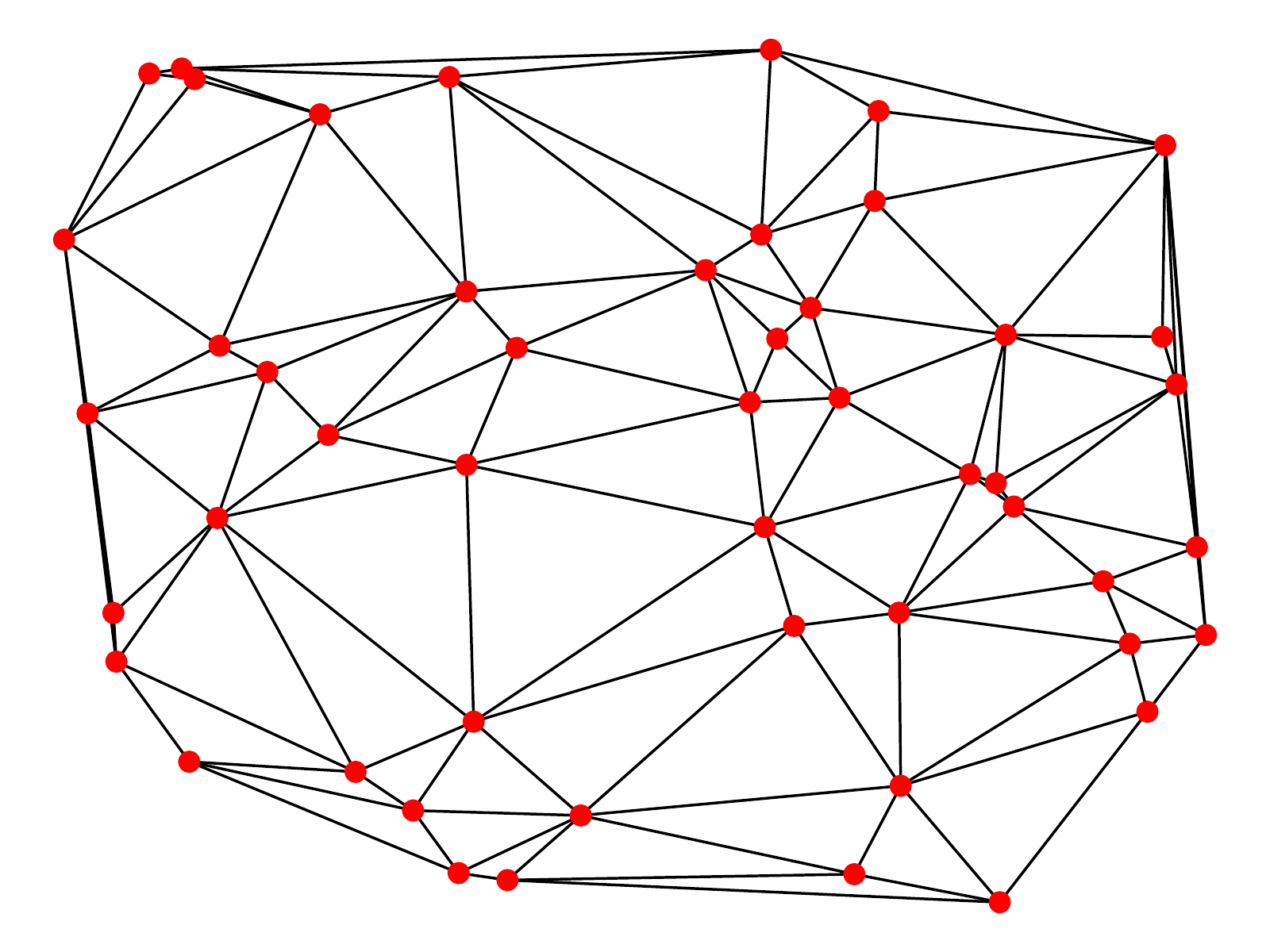} &
		\includegraphics[width=0.25\columnwidth, keepaspectratio]{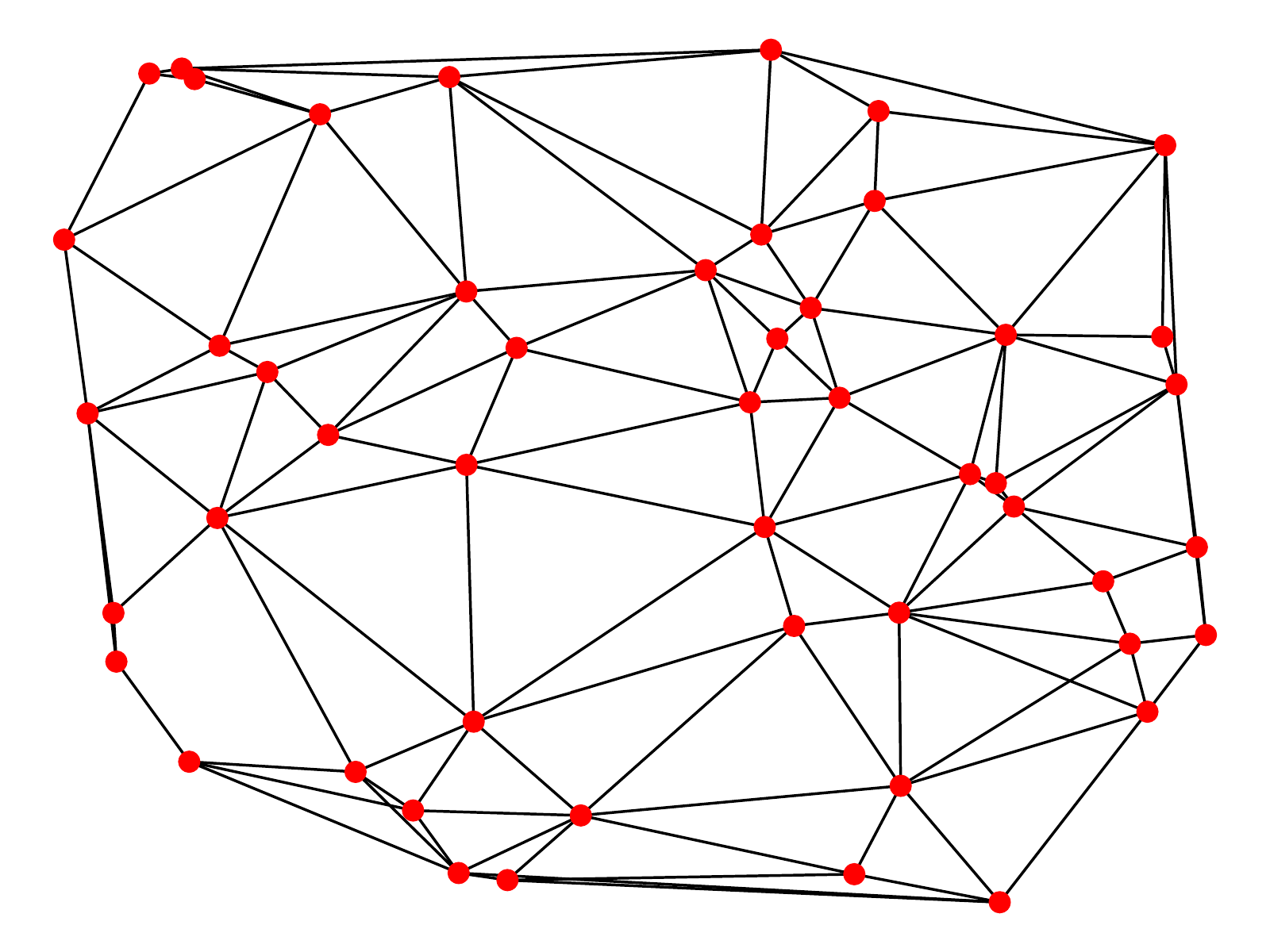} &
		\includegraphics[width=0.25\columnwidth, keepaspectratio]{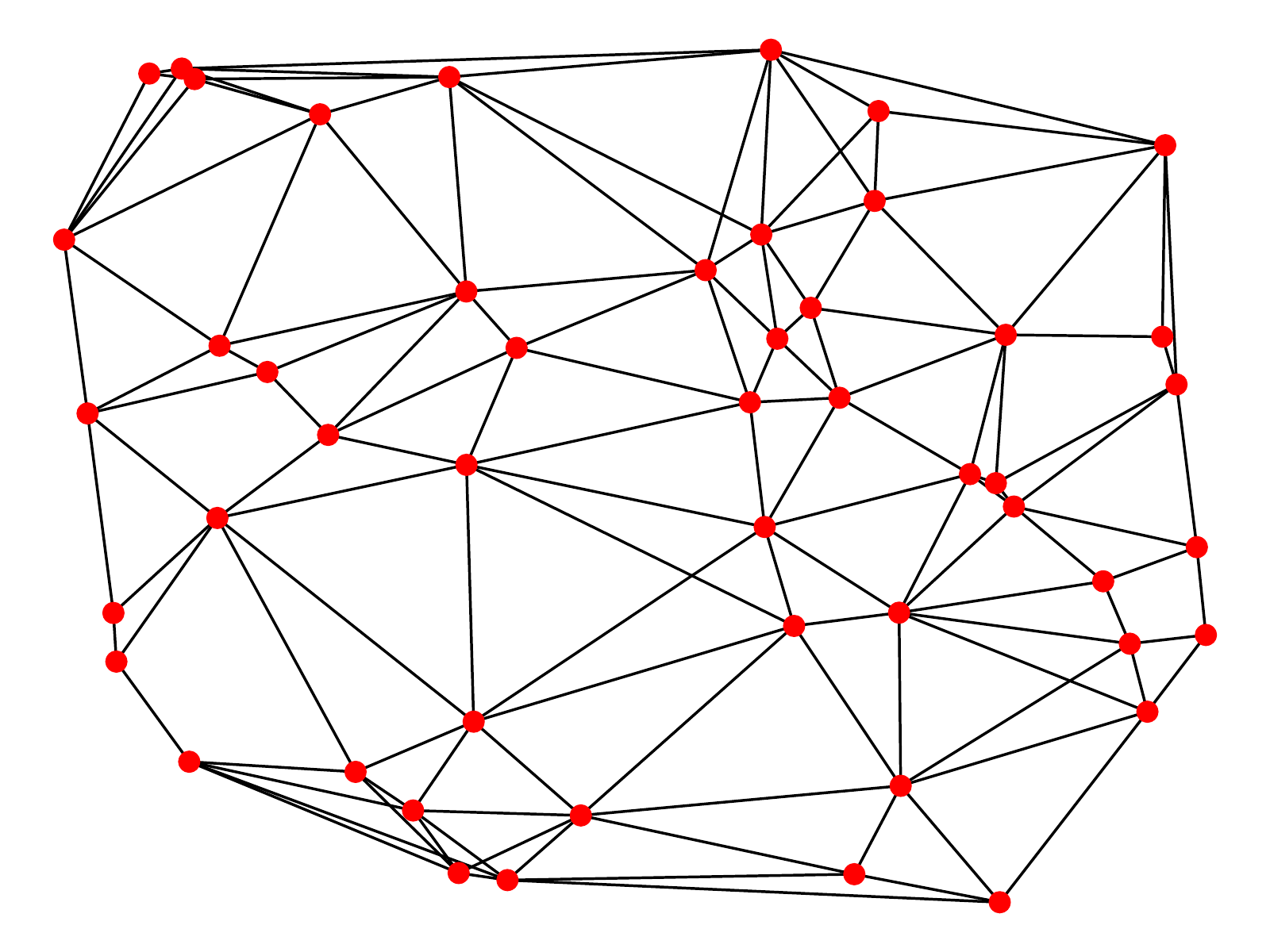}&
		\includegraphics[width=0.25\columnwidth, keepaspectratio]{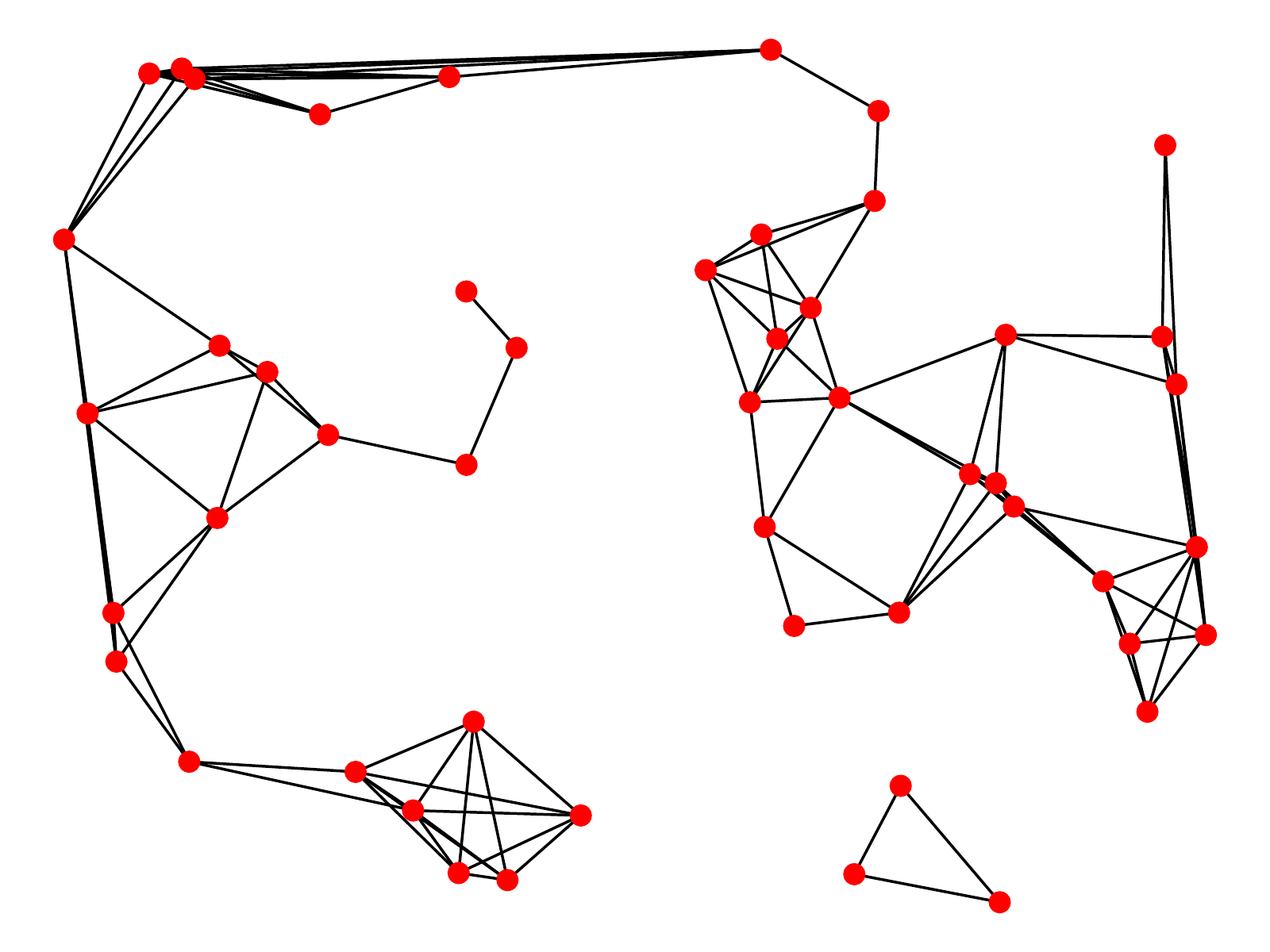}	\\
		\includegraphics[width=0.25\columnwidth,keepaspectratio]{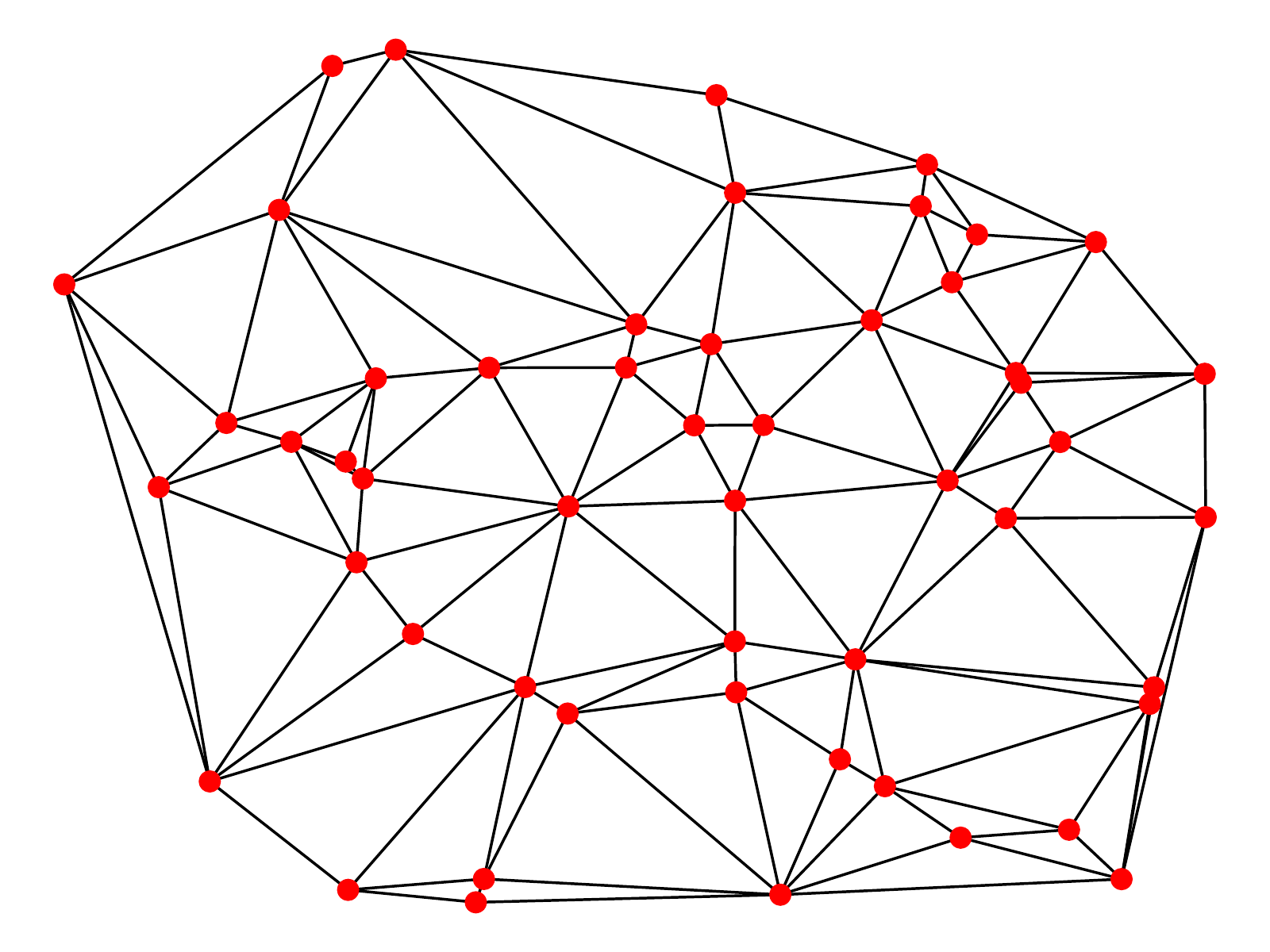} &
		\includegraphics[width=0.25\columnwidth, keepaspectratio]{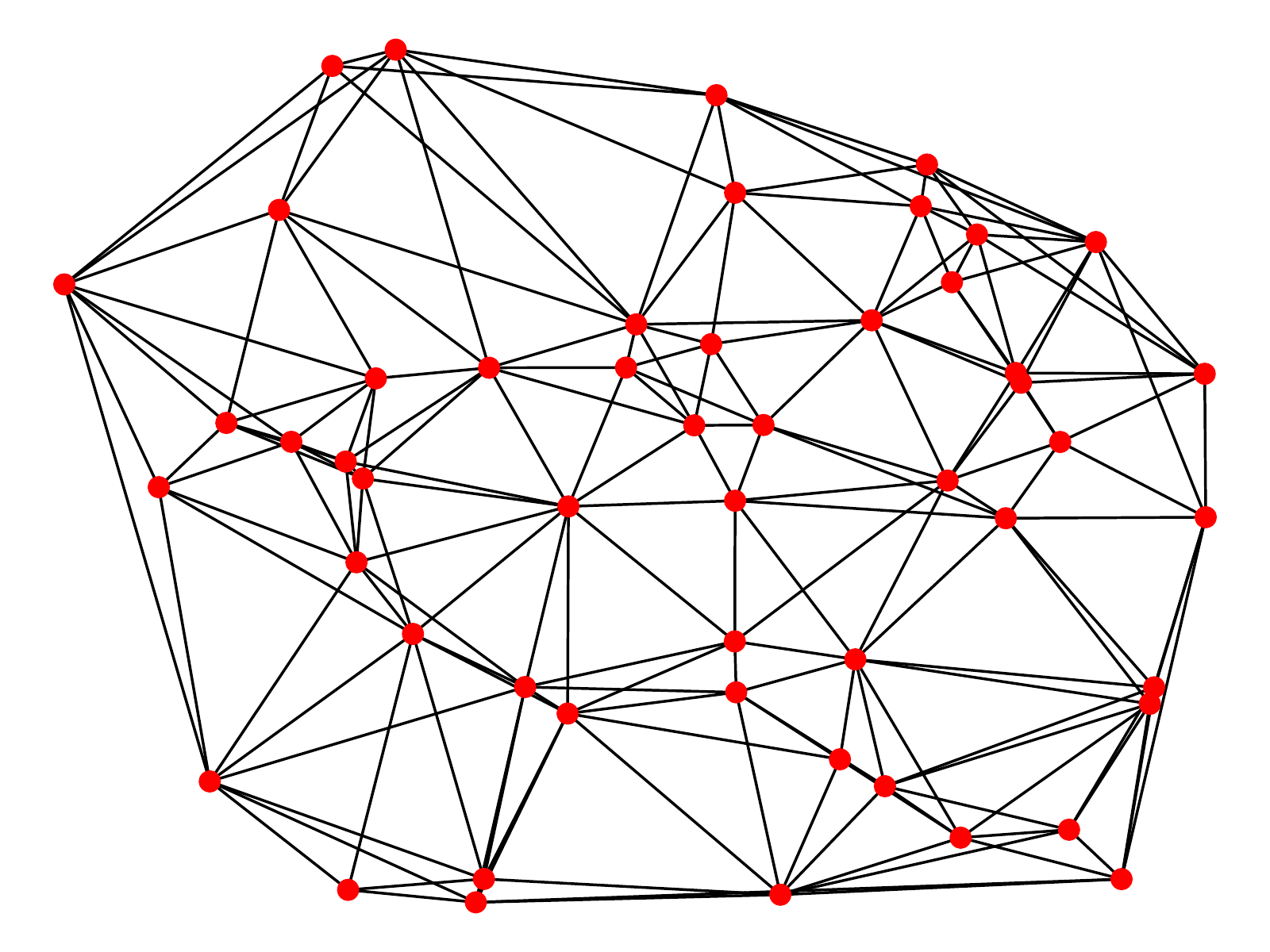} &
		\includegraphics[width=0.25\columnwidth, keepaspectratio]{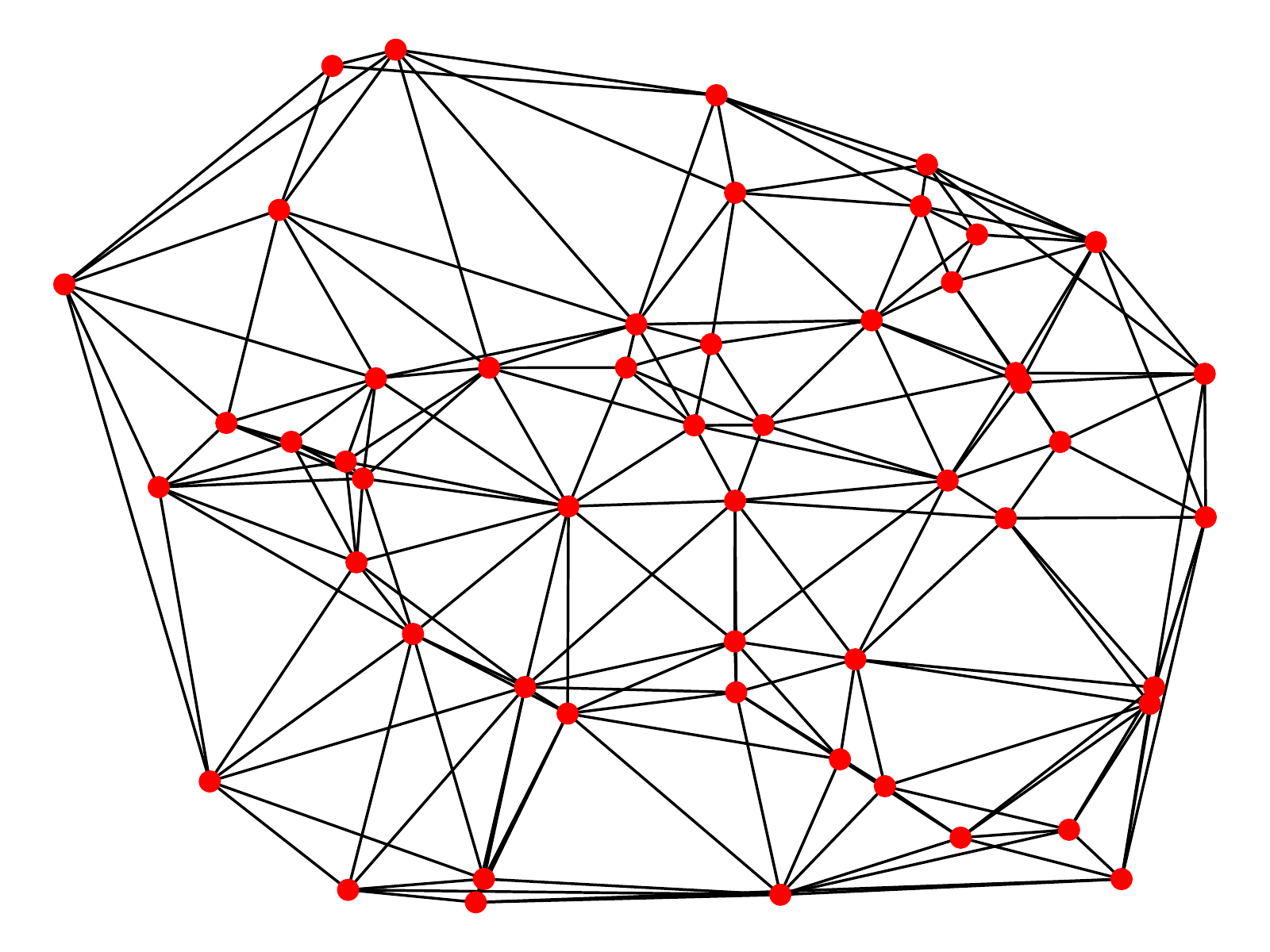}&
		\includegraphics[width=0.25\columnwidth, keepaspectratio]{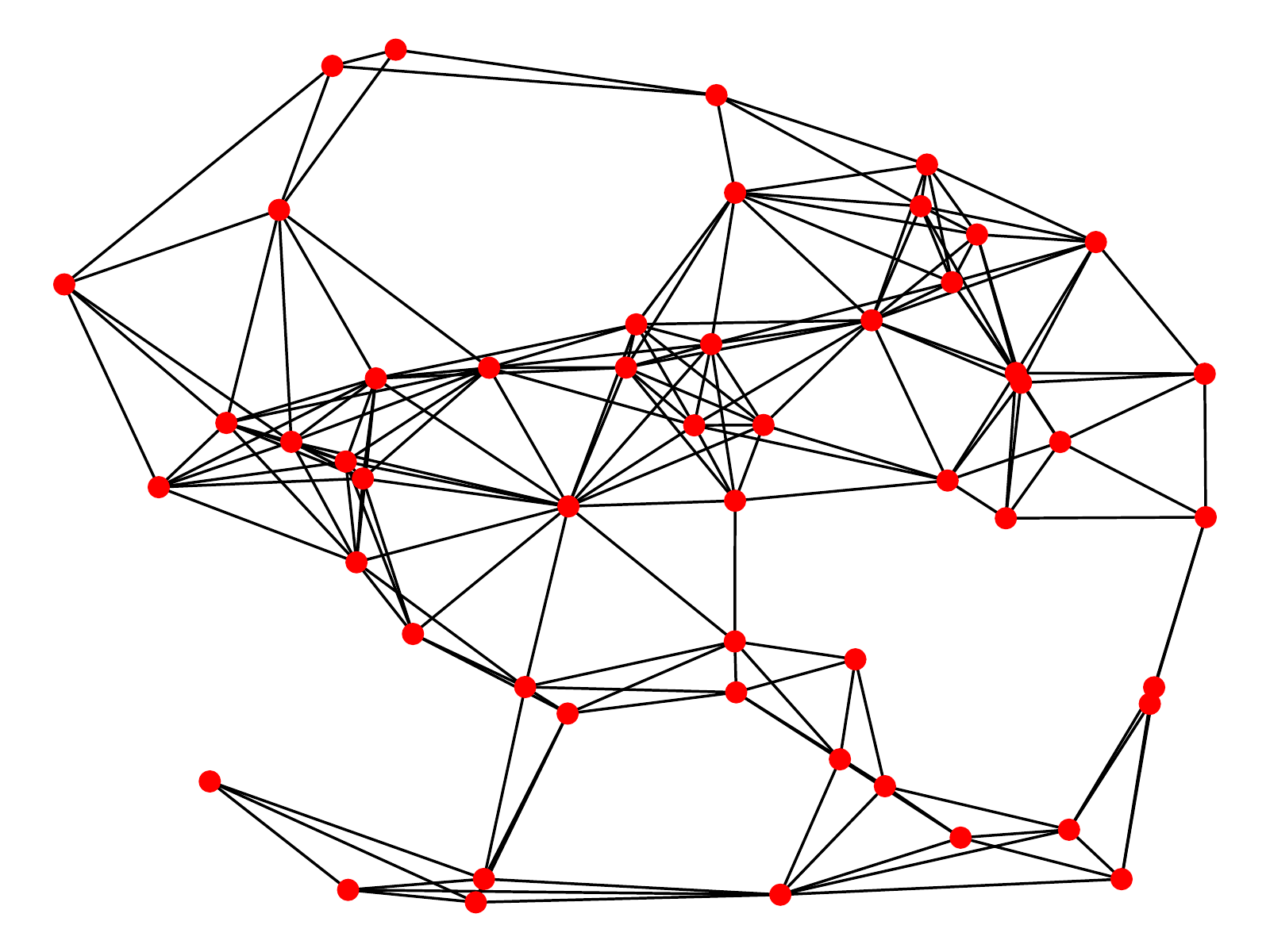}	\\
	\end{tabular}\vspace{-10pt}
	\caption{Results of Delaunay triangulation learning. Top: $n=50$; Bottom: $n \in \{20, \ldots, 80\}$. }
	\label{fig:Delaunay tris}
\end{figure}

\paragraph{Learning Delaunay triangulation.} 
In our model \textbf{S2G}, \textbf{S2G+}, $\vphi$ is an MLP $\brac{500, 500, 500, 1000, 500, 500, 80}$. $\vbeta$ is broadcasting as before for models S2G and S2G+, thus ending with 160 or 400 features per edge, respectively. $\vpsi$ is an MLP $\brac{1000, 1000, 1}$, ending as the edge probability. We use edge-wise binary cross-entropy loss. \textbf{S2G} and \textbf{S2G+} have 5918742 and 6398742 learnable parameters respectively.
The implementation of \textbf{SIAM} baseline is as follows: $\vphi$ is an MLP $\brac{700, 700, 700, 1400, 700, 700, 112}$, $\vbeta$ is broadcasting as in S2G, and $\vpsi$ is MLP $\brac{1400, 1400, 1}$ and edge-wise binary cross-entropy loss. \textbf{SIAM} has 5804037 learnable parameters. \textbf{SIAM-3} uses a Siamese MLP $\brac{500, 1000, 1500, 1250, 1000, 500, 500, 80}$ and has 5922330 learnable parameters. \textbf{GNN$k$ }is as the previous experiment, where $k \in \set{0, 5, 10}$, with 3 GraphConv layers of widths $\brac{1000, 1500, 1000}$, and it has  6007500 learnable parameters.
We searched learning rate from $\set{1e-2, 1e-3, 1e-4}$, using $1e-3$ with Adam optimizer. All models trained for up to 8 hours on a single Tesla V100 GPU.
A qualititive result is shown in Figure~\ref{fig:Delaunay tris}.

\paragraph{Set to 3-edges.}
For \textbf{S2G}, $\vphi$ is implemented using DeepSets $\brac{512,512,512}$. In this task, the model predicts supporting triangles of the convex hull, also referred to as faces, among triplets in the KNN graph. Hence, we do not maintain 3-rd order tensors in the memory. For each node we aggregate all the triangles which lie in its KNN ($k=10$) neighbors. In order to be invariant to the order of the 3 nodes in a face (\ie, the output tensor is symmetric w.r.t.~permutations of the triplets' order), each triplets is viewed as a set and fed to a DeepSets $\brac{64, 64, 64}$, max-pooled, and then to an MLP of widths $\brac{256,128,1}$. The model has 1186689 learnable parameters. As a loss, we used a combination of soft F1 score loss and a face-wise binary cross-entropy loss.
\textbf{SIAM} is identical except that $\vphi$ is implemented by MLP $\brac{1024, 1024, 512}$, and the second DeepSets is replaced by an MLP $\brac{128,128,64}$. It has 1718593 learnable parameters. \textbf{GNN5} is implemented as in the first experiments, with $k=5$, GraphConv layers $\brac{512,512,512,128}$ and the hyper-edge logits are computed as the sum of the product between the corresponding 3 nodes. It has 1184384 learnable paramaters.

For hyper-parameters search, we examined learning rates in $$\set{1e-5, 3e-5, 1e-4, 3e-4, 1e-3, 3e-3},$$ and DeepSets models of width $\set{64,128,256,512}$. We used Adam optimizers with learning rate of $1e-3$. Training took place for up to 36 hours on a single Tesla V100 GPU.

\section{Proofs}
\begin{proof}[Proof (of Theorem \ref{thm:set_to_k_edge_poly}).]
The general proof idea is to consider an arbitrary equivariant set-to-$k$-edge polynomial $\tP^k$ and use its equivariance property to show that it has the form as in \eqref{e:thm_P^k}. This is done by looking at a particular output entry $\tP^k_{\vi^0}$, where say $\vi^0=(1,2,\ldots,k)$. Then the proof considers two subsets of permutations: First, the subgroup of all permutations $\sigma\in S_n$ that fixes the numbers $1,2,\ldots,k$, \ie, $\sigma(\vi^0)=\vi^0$, but permute everything else freely; this subgroup is denoted $\mathrm{stab}(\vi^0)$. Second, permutations of the form $\sigma=(1\ i_1)(2\ i_2)\cdots (k\ i_k)$, where $\vi\in[n]^k$. Each of these permutation subsets reveals a different part in the structure of the equivariant polynomial $\tP^k$ and its relation to invariant polynomials. We provide the full proof next.

It is enough to prove Theorem \ref{thm:set_to_k_edge_poly} for $\dout=1$. Let $\vi^0=(1,2,\ldots,k)$ and consider any permutation $\sigma\in \mathrm{stab}(\vi^0)$. Then from equivariance of $\tP^k$ we have
$$\tP^k_{\vi^0}(\mX)=\tP^k_{\sigma^{-1}(\vi^0)}(\mX)=\tP^k_{\vi^0}(\sigma\cdot \mX),$$ and $\sigma\cdot \mX = (\vx_1,\ldots,\vx_k,\vx_{\sigma^{-1}(k+1)},\ldots,\vx_{\sigma^{-1}(n)})^T$. That is $\tP^k_{\vi^0}$ is invariant to permuting its last $n-k$ elements $\vx_{k+1},\ldots,\vx_n$; we say that $\tP_{\vi^0}$ is $S_{n-k}$ invariant. We next prove that $S_{n-k}$ invariance can be written using a combination of $S_n$ invariant polynomials and tensor products of $\vx_1,\ldots,\vx_k$:
\begin{lemma}\label{lem:recursive_equi_poly}
Let $p(\mX)=p(\vx_1,\ldots,\vx_k,\vx_{k+1},\ldots,\vx_n)$ be $S_{n-k}$ invariant polynomial. That is invariant to permuting the last $n-k$ terms. Then
\begin{equation}\label{e:lem_p_S_n_minus_k}
    p(\mX)=\sum_{\valpha} \vx_1^{\alpha^1}\cdots \vx_k^{\alpha^k}q_{\valpha}(\mX),
\end{equation} where $q_{\valpha}$ are $S_n$ invariant polynomials. 
\end{lemma}
Before we provide the proof of this lemma let us finish the proof of Theorem \ref{thm:set_to_k_edge_poly}. 
So now we know that $\tP^k_{\vi^0}$ has the form \eqref{e:lem_p_S_n_minus_k}. On the other hand let $\vi$ be an arbitrary multi-index and consider the permutation $\sigma=(1\ i_1)(2\ i_2)\cdots (k\ i_k)$. Again by permutation equivariance of $\tP^k$ we have 
\begin{align*}
  \tP^k_{i_1 i_2\cdots i_k}(\mX)&=\tP^k_{\sigma^{-1}(\vi_0)}(\mX)=\tP_{\vi_0}^k(\sigma\cdot \mX) \\ &= \sum_{\valpha} \vx_{i_1}^{\alpha^1}\cdots \vx_{i_k}^{\alpha^k}q_{\valpha}(\mX),
\end{align*}
which is a coordinate-wise form of \eqref{e:thm_P^k} with $\dout=1$. 
\end{proof}

\begin{proof}[Proof (of Lemma \ref{lem:recursive_equi_poly}).]
    First we expand $p$ as 
    \begin{equation}\label{e:lem_p_X}
     p(\mX)=\sum_{\valpha} \vx_1^{\alpha^1}\cdots \vx_k^{\alpha^k} q_{\valpha}(\vx_{k+1},\ldots,\vx_n),
    \end{equation}
    where $p_\valpha$ are $S_{n-k}$ invariant polynomials.  Since $S_{n-k}$ invariant polynomials with variables $\vx_{k+1},\ldots,\vx_n$ are generated by the power-sum multi-symmetric polynomials $$\sum_{i=k+1}^n \vx_i^\alpha = \sum_{i=1}^n \vx_i^\alpha - \sum_{i=1}^k \vx_i^\alpha,$$ with $|\alpha|\leq n-k$, see \eg, \cite{rydh2007minimal}, we have that each $p_\valpha(\vx_{k+1},\ldots,\vx_n)=\sum_{\valpha} \vx_1^{\alpha^1}\cdots \vx_k^{\alpha^k} r_{\valpha}(\mX)$, for some $S_n$ invariant polynomials $r_\valpha$. Plugging this in \eqref{e:lem_p_X} proves the lemma. 
    
    
\end{proof}

\begin{proof}[Proof (Lemma \ref{lem:approx_poly}).]
We can assume $\dout=1$. The general case is proved by finding approximating polynomial to each output feature coordinate. Let $\epsilon>0$. Using Stone-Weierstrass we can find a polynomial $\tQ:K\too\Real^{n^k}$ so that $\max_{\mX\in K}\norm{\tG^k(\mX)-\tQ(\mX)}_\infty< \epsilon$. 
Let $$\tP^k(\mX)  = \frac{1}{n!}\sum_{\sigma\in S_n}\sigma\cdot \tQ(\sigma^{-1} \cdot \mX).$$
Then $\tP^k$ is equivariant and since $\tG^k$ is also equivariant we have
\begin{align*}
  & \norm{\tG^k(\mX)-\tP^k(\mX)}_\infty  \\ &= \frac{1}{n!} \norm{\sum_{\sigma\in S_n}\sigma\cdot \parr{\tG^k(\sigma^{-1} \cdot \mX)- \tQ(\sigma^{-1} \cdot \mX)}}_\infty \\ &<
  \frac{1}{n!}\sum_{\sigma\in S_n}\epsilon = \epsilon.
\end{align*}

\end{proof}

\paragraph{Approximating $\tP^k$ with a network $\tF^k$.} We set a target $\epsilon>0$. Let $U\supset \mH(K)$ be a compact $\epsilon$-neighborhood of $\mH(K)$. $\vp$ is uniformly continuous over $\cup_\vi \vbeta(U)_{\vi,:}$. Choose $\vm$ so to be an $\epsilon/2$-approximation to $\vp$ over $\cup_\vi \vbeta(U)_{\vi,:}$. Let $\delta$ be so that for $\mY,\mY'\in U$, $\norm{\mY-\mY'}_\infty<\delta$ implies $\norm{\vp(\vbeta(\mY))-\vp(\vbeta(\mY'))}_\infty<\epsilon/2$. Now choose $\vphi$ so that it is $\delta_0$-approximation to $\mH$ over $K$ where $\delta_0< \min \set{\delta,\epsilon}$. This can be done since $\vphi$ is a universal set-to-set model as in \cite{segol2020on}. Lastly, putting all the pieces together we get for all $\vi$:
\begin{align*}
  \abs{\vp(\vbeta(\mH(\mX))_{\vi,:})-\vm(\vbeta(\vphi(\mX))_{\vi,:})}  \leq \\  \abs{\vp(\vbeta(\mH(\mX))_{\vi,:})-\vp(\vbeta(\vphi(\mX))_{\vi,:})} +  \\ 
  \abs{\vp(\vbeta(\vphi(\mX))_{\vi,:})-\vm(\vbeta(\vphi(\mX))_{\vi,:})} < \epsilon.
\end{align*}


\begin{proof}[Proof (Proposition \ref{prop:not_universal}).]
Consider the case $k=2$ and the constant function set-to-graph function $\tG(\mX) = \mI$, where $\mI$ is the identity $n\times n$ matrix; that is $\tG$ learns the constant value $1$ for 1-edges (nodes), and $0$ for 2-edges. Since $\vphi$ is equivariant we have that $\vphi(\one)=\one \otimes \va = \one \va^T$, for some vector $\va\in\Real^{d_1}$. Therefore $\vbeta(\vphi(\one))_{i_1,i_2,:}=\brac{\va,\va}$ and $\vm(\vbeta(\vphi(\one)))=\vm(\brac{\va,\va})=b\in\Real$. We get that  $\tF^2(\one)_{i_1,i_2,:}=b$ and $\norm{\mI - \tF^2(\one)}_\infty\geq 1/2$. \end{proof}

\section{Physics background.} 

The Large Hadron Collider (LHC) is the world's highest energy particle collider, located at the CERN laboratory in Geneva, and is used to study the fundamental particles of nature and their interactions. The LHC collides proton beams at high energy, and these collisions produce many new particles, which may be unstable or lead to subsequent particle production. For instance, the production of quarks (fundamental particles that make up protons, neutrons, and other hadrons) will lead to the production of many hadrons and eventually be manifest as a spray of particles called a \textit{jet}. The collisions take place in a vacuum, but the collision point is surrounded by large detectors that measure the outgoing particles that are stable enough to reach the detector several centimeters away.  In order to probe the properties of particles that are unstable, we need to infer which ``flavor'' of quark was the progenitor particle that led to a jet. This classification is performed in many stages, and we focus on a particular aspect of it known as vertex reconstruction, which we describe below.  

The location of the initial proton-proton collision is referred to as the primary vertex. Several particles emanating from the primary vertex are stable, will reach the detector, and will be part of the observed jet. Other particles will be unstable and travel a few millimeters before decaying in what is referred to as a secondary vertex. The unstable particles are not observed; however, the trajectories of the stable charged particles will be measured by detectors located around the collision point. Due to the presence of magnetic fields in the detector, the trajectories of the charged particles are helical. The helical trajectories are called \textit{tracks}, and are summarized by 6 numbers, called the perigee parameters, along with a covariance matrix quantifying the uncertainty of their measurement.

Vertex reconstruction can be composed into two parts, vertex finding and vertex fitting. Vertex finding refers to partitioning the tracks  into vertices, and vertex fitting is computing the most likely origin point in space for a collection of tracks. In the standard vertex reconstruction algorithms, these two parts are often intertwined and done together. In this application we perform the partitioning without performing the actual geometrical fit. From the physics point of view, once we improve the partitioning, the identification of the jets flavor is naturally improved.
Vertex reconstruction propagates to a number of down-stream data analysis tasks, such as particle identification (a classification problem). Therefore, improvements to the vertex reconstruction has significant impact on the sensitivity of collider experiments. 

\begin{figure}
	\centering
	\includegraphics[width=0.5\linewidth]{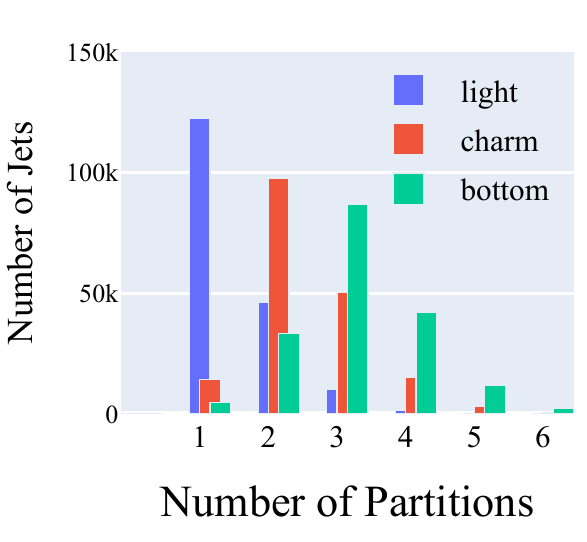} 
	\caption{Distribution of the number of partitions in each type of set.}
	\label{fig:numberofpartitions}
\end{figure}

\paragraph{Dataset.} 

Algorithms for particle physics are typically designed with high-fidelity simulators, which can provide labeled training data. These algorithms are then applied to and calibrated with real data collected by the LHC experiments. Our simulated samples are created with a standard simulation package called {\sc pythia}~\cite{Sj_strand_2015} and the detector is simulated with  {\sc delphes} ~\cite{de_Favereau_2014}. We use this software to generate synthetic datasets for three types (called "flavors") of jets. The generated sets are small, ranging from 2 to 14 elements each. 
The three different jet types are labeled bottom-jets, charm-jets, and light-jets (B/C/L). The important distinction between the flavors is the typical number of partitions in each set. Figure~\ref{fig:numberofpartitions} shows the distribution of the number of partitions (vertices) in each flavor: bottom jets typically have multiple partitions; charm jets also have multiple partitions, but fewer than bottom jets; and light jets typically have only one partition.

\paragraph{AVR algorithm.} We compare the model performance to a non-learning algorithm,  (AVR), implemented in RAVE~\cite{Waltenberger:2011zz}. The basic concept of AVR is to perform a least squares fit of the vertex position given the track trajectories and their errors, remove less compatible tracks from the fit, and refit them to secondary vertices.

\end{document}